%% file: 2LM-M_LLR_polymer_v23.tex
\documentclass[12pt]{article}
\newcommand{\filename}{2LM-M\_LLR\_polymer\_v23}
\usepackage{authblk}
\usepackage{amsfonts}
\usepackage{amssymb}
\usepackage{amsmath}
\usepackage{amsthm}
\usepackage{framed}
\usepackage{graphicx}

\usepackage{enumitem}
\usepackage{geometry}
\usepackage{url}
\usepackage{booktabs}

\usepackage{multirow}

\usepackage{mdframed}
\usepackage{mathtools}
\usepackage{algorithm}  
\usepackage{algorithmic}

\input symbols.tex

\begin{document} 

\begin{center}
   {\Large\bf 
   A Method for Inferring Polymers Based on 
   Linear Regression and Integer Programming}
\end{center}

\input authors.tex

\input Abstract_2LMM_LLR_polymer.tex

\input Introduction_2LMM_LLR_polymer_arXiv_b.tex

\input Preliminary_2LMM_polymer.tex

\input Two-layered_Model_polymer.tex

\input Result_2LMM_LLR_Phase1_polymer.tex

\input Result_2LMM_LLR_Phase2_polymer.tex

\input  Conclusion_2LMM_LLR_polymer.tex

\input Reference_2LMM_LLR_polymer.tex
\clearpage
 \appendix
\centerline{\bf\LARGE Appendix}

\input Linear_Regression.tex

\input Descriptor_M_lfac_polymer.tex

\clearpage
 \input Specification_2LMM_polymer.tex

\clearpage

\input Test_instances_2LMM_polymer.tex

\clearpage

\input Constraints_MILP_2LMM_base_polymer.tex

\input Constraints_MILP_2LMH_ac_cs_ec_polymer.tex

\input Constraints_MILP_2LM_normalization.tex

\end{document}

%% file: symbols.tex
\newcommand{\cnt}{\mathrm{cnt}}  

\newcommand{\lnk}{\mathrm{lnk}}  
\newcommand{\Elnk}{E^\mathrm{lnk}}    
\newcommand{\nlnk}{\mathrm{n}^\mathrm{lnk}}

\newcommand{\lf}{\mathrm{lf}}

\newcommand{\eledeg}{\mathrm{eledeg}}  
\newcommand{\eledegC}{\mathrm{eledeg}_\mathrm{C}}   
\newcommand{\eledegT}{\mathrm{eledeg}_\mathrm{T}}   
\newcommand{\eledegF}{\mathrm{eledeg}_\mathrm{F}}   
\newcommand{\eledegX}{\mathrm{eledeg}_\mathrm{X}}   

\newcommand{\vion}{\mathrm{v}_\mathrm{ion}}  

\newcommand{\ttH}{{\tt H}}  
\newcommand{\ttC}{{\tt C}}  
\newcommand{\ttO}{{\tt O}}  
\newcommand{\ttN}{{\tt N}}  
\newcommand{\ttP}{{\tt P}}  
\newcommand{\ttF}{{\tt F}}  
\newcommand{\ttCl}{{\tt Cl}}  
\newcommand{\ttS}{{\tt S}}  
\newcommand{\ttSi}{{\tt Si}}

\newcommand{\oH}{\overline{{\tt H}}}  

\newcommand{\Z}{\mathbb{Z}}  
  
\newcommand{\C}{\mathbb{C}}  
\newcommand{\Co}{\mathbb{C}}  
\newcommand{\fr}{\mathrm{fr}}    
   
\newcommand{\anC}{\langle \mathbb{C} \rangle}  
\newcommand{\anpsi}{\langle \psi \rangle}  

\newcommand{\VH}{V_{\tt H}}

\newcommand{\R}{\mathbb{R}} 
\newcommand{\RK}{\mathbb{R}^K} 
\newcommand{\RKw}{\mathbb{R}^{K+1}}

\newcommand{\deghyd}{\deg^\mathrm{hyd}}

\newcommand{\FrC}{\mathcal{F}^\mathrm{C}} 
\newcommand{\FrT}{\mathcal{F}^\mathrm{T}} 
\newcommand{\FrF}{\mathcal{F}^\mathrm{F}} 
\newcommand{\FrX}{\mathcal{F}^\mathrm{X}}  

\newcommand{\dcp}{\mathrm{dcp}}

\newcommand{\Vleaf}{V_\mathrm{leaf}} 
\newcommand{\Eleaf}{E_\mathrm{leaf}} 

\newcommand{\sint}{\sigma_\mathrm{int}} 
\newcommand{\sce}{\sigma_\mathrm{ce}}

\newcommand{\Ez}{E_{(0/1)}}
\newcommand{\Ew}{E_{(\geq 1)}}
\newcommand{\Et}{E_{(\geq 2)}}
\newcommand{\Eew}{E_{(=1)}}

\newcommand{\Ilnk}{I_\mathrm{lnk}} 
\newcommand{\Iz}{I_{(0/1)}}
\newcommand{\Iw}{I_{(\geq 1)}}
\newcommand{\It}{I_{(\geq 2)}}
\newcommand{\Iew}{I_{(=1)}}

\newcommand{\Gac}{\Gamma_\mathrm{ac}} 
\newcommand{\Gacs}{\Gamma_\mathrm{ac,<}} 
\newcommand{\Gace}{\Gamma_\mathrm{ac,=}} 
\newcommand{\Gacl}{\Gamma_\mathrm{ac,>}}

\newcommand{\tGacC}{\widetilde{\Gamma}_\mathrm{ac}^\mathrm{C}}  
\newcommand{\tGacT}{\widetilde{\Gamma}_\mathrm{ac}^\mathrm{T}} 
\newcommand{\tGacF}{\widetilde{\Gamma}_\mathrm{ac}^\mathrm{F}}  
\newcommand{\tGacCT}{\widetilde{\Gamma}_\mathrm{ac}^\mathrm{CT}}  
\newcommand{\tGacTC}{\widetilde{\Gamma}_\mathrm{ac}^\mathrm{TC}}  
\newcommand{\tGacCF}{\widetilde{\Gamma}_\mathrm{ac}^\mathrm{CF}}  
\newcommand{\tGacTF}{\widetilde{\Gamma}_\mathrm{ac}^\mathrm{TF}}

\newcommand{\tLdgX}{\widetilde{\Lambda}_\mathrm{dg}^\mathrm{X}}   
\newcommand{\tLdgC}{\widetilde{\Lambda}_\mathrm{dg}^\mathrm{C}}   
\newcommand{\tLdgT}{\widetilde{\Lambda}_\mathrm{dg}^\mathrm{T}}   
\newcommand{\tLdgF}{\widetilde{\Lambda}_\mathrm{dg}^\mathrm{F}}

\newcommand{\tGecC}{\widetilde{\Gamma}_\mathrm{ec}^\mathrm{C}}  
\newcommand{\tGecT}{\widetilde{\Gamma}_\mathrm{ec}^\mathrm{T}} 
\newcommand{\tGecF}{\widetilde{\Gamma}_\mathrm{ec}^\mathrm{F}}  
\newcommand{\tGecCT}{\widetilde{\Gamma}_\mathrm{ec}^\mathrm{CT}}  
\newcommand{\tGecTC}{\widetilde{\Gamma}_\mathrm{ec}^\mathrm{TC}}  
\newcommand{\tGecCF}{\widetilde{\Gamma}_\mathrm{ec}^\mathrm{CF}}  
\newcommand{\tGecTF}{\widetilde{\Gamma}_\mathrm{ec}^\mathrm{TF}}

\newcommand{\typ}{\mathrm{t}}

\newcommand{\w}{w}
\newcommand{\x}{x}

\newcommand{\ta}{{\tt a}}
\newcommand{\tb}{{\tt b}}
\newcommand{\Ldg}{\Lambda_{\mathrm{dg}}}

\newcommand{\fc}{\mathrm{fc}} 
\newcommand{\betar}{\beta_\mathrm{r}}

\newcommand{\val}{\mathrm{val}}

\newcommand{\inte}{\mathrm{int}}

\newcommand{\F}{\mathcal{F}}

\newcommand{\T}{\mathcal{T}}

\newcommand{\nint}{\mathrm{n}^\mathrm{int}}

\newcommand{\h}{\mathrm{ht}}

\newcommand{\cs}{\mathrm{cs}}
\newcommand{\ch}{\mathrm{ch}}

\newcommand{\dg}{\mathrm{dg}}

\newcommand{\na}{\mathrm{na}}
 
\newcommand{\naX}{\mathrm{na}_\mathrm{X}}

\newcommand{\naC}{\mathrm{na}_\mathrm{C}}
\newcommand{\naT}{\mathrm{na}_\mathrm{T}}
\newcommand{\naF}{\mathrm{na}_\mathrm{F}}

\newcommand{\ecX}{\mathrm{ec}_\mathrm{X}}

\newcommand{\ecC}{\mathrm{ec}_\mathrm{C}}
\newcommand{\ecT}{\mathrm{ec}_\mathrm{T}}
\newcommand{\ecF}{\mathrm{ec}_\mathrm{F}}
\newcommand{\ecCT}{\mathrm{ec}_\mathrm{CT}}
\newcommand{\ecTC}{\mathrm{ec}_\mathrm{TC}}
\newcommand{\ecTF}{\mathrm{ec}_\mathrm{TF}}
\newcommand{\ecCF}{\mathrm{ec}_\mathrm{CF}}
 
\newcommand{\acX}{\mathrm{ac}_\mathrm{X}}

\newcommand{\acC}{\mathrm{ac}_\mathrm{C}}
\newcommand{\acT}{\mathrm{ac}_\mathrm{T}}
\newcommand{\acF}{\mathrm{ac}_\mathrm{F}}
\newcommand{\acCT}{\mathrm{ac}_\mathrm{CT}}
\newcommand{\acTC}{\mathrm{ac}_\mathrm{TC}}
\newcommand{\acTF}{\mathrm{ac}_\mathrm{TF}}
\newcommand{\acCF}{\mathrm{ac}_\mathrm{CF}}

\newcommand{\bdX}{\mathrm{bd}_\mathrm{X}}

\newcommand{\bdC}{\mathrm{bd}_\mathrm{C}}
\newcommand{\bdT}{\mathrm{bd}_\mathrm{T}}
\newcommand{\bdF}{\mathrm{bd}_\mathrm{F}}
\newcommand{\bdCT}{\mathrm{bd}_\mathrm{CT}}
\newcommand{\bdTC}{\mathrm{bd}_\mathrm{TC}}
\newcommand{\bdTF}{\mathrm{bd}_\mathrm{TF}}
\newcommand{\bdCF}{\mathrm{bd}_\mathrm{CF}}

\newcommand{\ns}{\mathrm{ns}}

\newcommand{\ec}{\mathrm{ec}}
\newcommand{\ac}{\mathrm{ac}}

\newcommand{\bl}{\mathrm{bl}}

\newcommand{\bd}{\mathrm{bd}}

\newcommand{\UB}{\mathrm{UB}}
\newcommand{\LB}{\mathrm{LB}}

\newcommand{\ex}{\mathrm{ex}}

\newcommand{\GC}{G_\mathrm{C}}

\newcommand{\mC}{m_\mathrm{C}}

\newcommand{\hC}{h^\mathrm{C}}
\newcommand{\hT}{h^\mathrm{T}} 
 
\newcommand{\hX}{h^\mathrm{X}}

\newcommand{\VF}{V_\mathrm{F}}
\newcommand{\VT}{V_\mathrm{T}}
\newcommand{\VC}{V_\mathrm{C}} 
 
\newcommand{\VX}{V_\mathrm{X}}

\newcommand{\ET}{E_\mathrm{T}}
\newcommand{\EC}{E_\mathrm{C}}

\newcommand{\EF}{E_\mathrm{F}}
\newcommand{\ECT}{E_\mathrm{CT}}
\newcommand{\ETC}{E_\mathrm{TC}}
\newcommand{\ETF}{E_\mathrm{TF}}

\newcommand{\ECF}{E_\mathrm{CF}}

\newcommand{\EX}{E_\mathrm{X}}

\newcommand{\vT}{{v^\mathrm{T}}}
\newcommand{\vC}{{v^\mathrm{C}}} 
  
\newcommand{\vX}{{v^\mathrm{X}}}

\newcommand{\vF}{{v^\mathrm{F}}}

\newcommand{\eF}{{e^\mathrm{F}}}
\newcommand{\eT}{{e^\mathrm{T}}}
\newcommand{\eC}{{e^\mathrm{C}}} 
 
\newcommand{\eX}{{e^\mathrm{X}}}

\newcommand{\eCF}{{e^\mathrm{CF}}}

\newcommand{\eCT}{{e^\mathrm{CT}}}
\newcommand{\eTC}{{e^\mathrm{TC}}}
\newcommand{\eTF}{{e^\mathrm{TF}}}

\newcommand{\tT}{{t_\mathrm{T}}}
\newcommand{\tC}{{t_\mathrm{C}}} 
\newcommand{\tF}{{t_\mathrm{F}}} 
 
\newcommand{\tX}{{t_\mathrm{X}}}

\newcommand{\IC}{{I_\mathrm{C}}}

\newcommand{\degCint}{{\deg_\mathrm{C}^\mathrm{int}}}
\newcommand{\degTint}{{\deg_\mathrm{T}^\mathrm{int}}}
\newcommand{\degFint}{{\deg_\mathrm{F}^\mathrm{int}}}
\newcommand{\degXint}{{\deg_\mathrm{X}^\mathrm{int}}}

\newcommand{\hyddeg}{\mathrm{hyddeg}}

\newcommand{\hyddegX}{\mathrm{hyddeg}^\mathrm{X}}

\newcommand{\degCex}{{\deg_\mathrm{C}^\mathrm{ex}}}
\newcommand{\degTex}{{\deg_\mathrm{T}^\mathrm{ex}}}
\newcommand{\degFex}{{\deg_\mathrm{F}^\mathrm{ex}}}
\newcommand{\degXex}{{\deg_\mathrm{X}^\mathrm{ex}}}
\newcommand{\degF}{{\deg^\mathrm{F}}}
\newcommand{\degT}{{\deg^\mathrm{T}}}
\newcommand{\degC}{{\deg^\mathrm{C}}} 
 
\newcommand{\degX}{{\deg^\mathrm{X}}}

\newcommand{\degCT}{\deg_\mathrm{CT}}
\newcommand{\degTC}{\deg_\mathrm{TC}}

\newcommand{\degCTT}{\deg^\mathrm{CT}_\mathrm{T}}
\newcommand{\degTCT}{\deg^\mathrm{TC}_\mathrm{T}}
\newcommand{\degCFF}{\deg^\mathrm{CF}_\mathrm{F}}
\newcommand{\degTFF}{\deg^\mathrm{TF}_\mathrm{F}}

\newcommand{\tldgC}{{\widetilde{\deg}_\mathrm{C}} }

\newcommand{\cF}{{c_\mathrm{F}}}

\newcommand{\kC}{{k_\mathrm{C}}}

\newcommand{\chiF}{{\chi^\mathrm{F}}} 
\newcommand{\dclrF}{\delta_{\chi}^\mathrm{F}}
\newcommand{\clrF}{\mathrm{clr}^{\mathrm{F}}}

\newcommand{\chiT}{{\chi^\mathrm{T}}}
\newcommand{\dclrT}{\delta_{\chi}^\mathrm{T}}
\newcommand{\clrT}{\mathrm{clr}^{\mathrm{T}}}

\newcommand{\tail}{\mathrm{tail}} 
\newcommand{\hd}{\mathrm{head}}

\newcommand{\dlfrF}{\delta_\mathrm{fr}^\mathrm{F}}

\newcommand{\dlfrC}{\delta_\mathrm{fr}^\mathrm{C}} 
 
\newcommand{\dlfrX}{\delta_\mathrm{fr}^\mathrm{X}}
 
\newcommand{\ddgF}{\delta_\mathrm{dg}^\mathrm{F}}
\newcommand{\ddgT}{\delta_\mathrm{dg}^\mathrm{T}}
\newcommand{\ddgC}{\delta_\mathrm{dg}^\mathrm{C}} 
 
\newcommand{\ddgX}{\delta_\mathrm{dg}^\mathrm{X}}

\newcommand{\ddgFint}{\delta_\mathrm{dg,F}^\mathrm{int}}
\newcommand{\ddgTint}{\delta_\mathrm{dg,T}^\mathrm{int}}
\newcommand{\ddgCint}{\delta_\mathrm{dg,C}^\mathrm{int}}  
\newcommand{\ddgXint}{\delta_\mathrm{dg,X}^\mathrm{int}}

\newcommand{\bF}{\beta^\mathrm{F}}
\newcommand{\bT}{\beta^\mathrm{T}}
\newcommand{\bC}{\beta^\mathrm{C}} 
\newcommand{\bX}{\beta^\mathrm{X}}
  
\newcommand{\bCT}{\beta^\mathrm{CT}}
\newcommand{\bTC}{\beta^\mathrm{TC}} 
\newcommand{\bTF}{\beta^\mathrm{TF}} 
\newcommand{\bCF}{\beta^\mathrm{CF}} 
\newcommand{\bXF}{\beta^\mathrm{XF}} 
\newcommand{\bsF}{\beta^{*\mathrm{F}}}

\newcommand{\bFex}{\beta^\mathrm{F}_\mathrm{ex}} 
\newcommand{\bTex}{\beta^\mathrm{T}_\mathrm{ex}} 
\newcommand{\bCex}{\beta^\mathrm{C}_\mathrm{ex}} 
\newcommand{\bXex}{\beta^\mathrm{X}_\mathrm{ex}}

\newcommand{\delbF}{\delta_{\beta}^\mathrm{F}}
\newcommand{\delbT}{\delta_{\beta}^\mathrm{T}}
\newcommand{\delbC}{\delta_{\beta}^\mathrm{C}}

\newcommand{\delbCT}{\delta_{\beta}^\mathrm{CT}}
\newcommand{\delbTC}{\delta_{\beta}^\mathrm{TC}}

\newcommand{\delbsF}{\delta_{\beta}^{*\mathrm{F}}} 
\newcommand{\delbX}{\delta_{\beta}^\mathrm{X}}

\newcommand{\aF}{{\alpha}^\mathrm{F}}
\newcommand{\aT}{{\alpha}^\mathrm{T}}
\newcommand{\aC}{{\alpha}^\mathrm{C}}  
 
\newcommand{\aX}{{\alpha}^\mathrm{X}}
\newcommand{\aCT}{{\alpha}^\mathrm{CT}}
\newcommand{\aTC}{{\alpha}^\mathrm{TC}}

\newcommand{\aCF}{{\alpha}^\mathrm{CF}}  
\newcommand{\aTF}{{\alpha}^\mathrm{TF}}

\newcommand{\delaC}{\delta_\mathrm{\alpha}^{\mathrm{C}}}
\newcommand{\delaT}{\delta_\mathrm{\alpha}^{\mathrm{T}}}
\newcommand{\delaF}{\delta_\mathrm{\alpha}^{\mathrm{F}}}
\newcommand{\delaX}{\delta_\mathrm{\alpha}^{\mathrm{X}}}

\newcommand{\dlnsF}{\delta_{\mathrm{ns}}^\mathrm{F}}
\newcommand{\dlnsT}{\delta_{\mathrm{ns}}^\mathrm{T}}
\newcommand{\dlnsC}{\delta_{\mathrm{ns}}^\mathrm{C}} 
\newcommand{\dlnsX}{\delta_{\mathrm{ns}}^\mathrm{X}}

\newcommand{\dlacTlnk}{\delta_{\mathrm{ac}}^\mathrm{T,lnk}} 

\newcommand{\dlacF}{\delta_{\mathrm{ac}}^\mathrm{F}}
\newcommand{\dlacT}{\delta_{\mathrm{ac}}^\mathrm{T}}
\newcommand{\dlacC}{\delta_{\mathrm{ac}}^\mathrm{C}}

\newcommand{\dlacCT}{\delta_{\mathrm{ac}}^\mathrm{CT}}
\newcommand{\dlacTC}{\delta_{\mathrm{ac}}^\mathrm{TC}}

\newcommand{\dlacCF}{\delta_{\mathrm{ac}}^\mathrm{CF}} 
\newcommand{\dlacTF}{\delta_{\mathrm{ac}}^\mathrm{TF}} 
\newcommand{\dlacX}{\delta_{\mathrm{ac}}^\mathrm{X}}

\newcommand{\dlecTlnk}{\delta_{\mathrm{ec}}^\mathrm{T,lnk}}
\newcommand{\dlecF}{\delta_{\mathrm{ec}}^\mathrm{F}}
\newcommand{\dlecT}{\delta_{\mathrm{ec}}^\mathrm{T}}
\newcommand{\dlecC}{\delta_{\mathrm{ec}}^\mathrm{C}}

\newcommand{\dlecCT}{\delta_{\mathrm{ec}}^\mathrm{CT}}

\newcommand{\dlecCTC}{\delta_{\mathrm{ec,C}}^\mathrm{CT}}
\newcommand{\dlecTC}{\delta_{\mathrm{ec}}^\mathrm{TC}}
\newcommand{\dlecTCC}{\delta_{\mathrm{ec,C}}^\mathrm{TC}}

\newcommand{\dlecCFC}{\delta_{\mathrm{ec,C}}^\mathrm{CF}}

\newcommand{\dlecTFT}{\delta_{\mathrm{ec,T}}^\mathrm{TF}} 
 
\newcommand{\dlecX}{\delta_{\mathrm{ec}}^\mathrm{X}}

\newcommand{\DlacFp}{\Delta_{\mathrm{ac}}^\mathrm{F+}}
\newcommand{\DlacTp}{\Delta_{\mathrm{ac}}^\mathrm{T+}}
\newcommand{\DlacCp}{\Delta_{\mathrm{ac}}^\mathrm{C+}}

\newcommand{\DlacCTp}{\Delta_{\mathrm{ac}}^\mathrm{CT+}}
\newcommand{\DlacTCp}{\Delta_{\mathrm{ac}}^\mathrm{TC+}}

\newcommand{\DlacCFp}{\Delta_{\mathrm{ac}}^\mathrm{CF+}} 
\newcommand{\DlacTFp}{\Delta_{\mathrm{ac}}^\mathrm{TF+}} 
\newcommand{\DlacXp}{\Delta_{\mathrm{ac}}^\mathrm{X+}}

\newcommand{\DlacFm}{\Delta_{\mathrm{ac}}^\mathrm{F-}}
\newcommand{\DlacTm}{\Delta_{\mathrm{ac}}^\mathrm{T-}}
\newcommand{\DlacCm}{\Delta_{\mathrm{ac}}^\mathrm{C-}}

\newcommand{\DlacCTm}{\Delta_{\mathrm{ac}}^\mathrm{CT-}}
\newcommand{\DlacTCm}{\Delta_{\mathrm{ac}}^\mathrm{TC-}}

\newcommand{\DlacCFm}{\Delta_{\mathrm{ac}}^\mathrm{CF-}} 
\newcommand{\DlacTFm}{\Delta_{\mathrm{ac}}^\mathrm{TF-}} 
\newcommand{\DlacXm}{\Delta_{\mathrm{ac}}^\mathrm{X-}}

\newcommand{\DlecFp}{\Delta_{\mathrm{ec}}^\mathrm{F+}}
\newcommand{\DlecTp}{\Delta_{\mathrm{ec}}^\mathrm{T+}}
\newcommand{\DlecCp}{\Delta_{\mathrm{ec}}^\mathrm{C+}}

\newcommand{\DlecCTp}{\Delta_{\mathrm{ec}}^\mathrm{CT+}}
\newcommand{\DlecTCp}{\Delta_{\mathrm{ec}}^\mathrm{TC+}}

\newcommand{\DlecCFp}{\Delta_{\mathrm{ec}}^\mathrm{CF+}} 
\newcommand{\DlecTFp}{\Delta_{\mathrm{ec}}^\mathrm{TF+}} 
\newcommand{\DlecXp}{\Delta_{\mathrm{ec}}^\mathrm{X+}}

\newcommand{\DlecFm}{\Delta_{\mathrm{ec}}^\mathrm{F-}}
\newcommand{\DlecTm}{\Delta_{\mathrm{ec}}^\mathrm{T-}}
\newcommand{\DlecCm}{\Delta_{\mathrm{ec}}^\mathrm{C-}}

\newcommand{\DlecCTm}{\Delta_{\mathrm{ec}}^\mathrm{CT-}}
\newcommand{\DlecTCm}{\Delta_{\mathrm{ec}}^\mathrm{TC-}}

\newcommand{\DlecCFm}{\Delta_{\mathrm{ec}}^\mathrm{CF-}} 
\newcommand{\DlecTFm}{\Delta_{\mathrm{ec}}^\mathrm{TF-}} 
\newcommand{\DlecXm}{\Delta_{\mathrm{ec}}^\mathrm{X-}}

%% file: authors.tex
\begin{center}
Ryota Ido$^1$, 
Shengjuan Cao$^1$, 
Jianshen Zhu$^1$, 
Naveed Ahmed Azam$^1$, 
Kazuya Haraguchi$^{1}$, 
Liang Zhao$^2$, 
Hiroshi Nagamochi$^1$ 
 and  
 Tatsuya Akutsu$^3$ 
\end{center} 
%
%
{\small 
$^1$Department of Applied Mathematics and Physics, Kyoto University, Kyoto 606-8501, Japan\\
$^2$Graduate School of Advanced Integrated Studies in Human Survavibility
     (Shishu-Kan),    Kyoto University, Kyoto 606-8306, Japan \\
 $^3$Bioinformatics Center,  Institute for Chemical Research, 
  Kyoto University, Uji 611-0011, Japan 
}

%% file: Abstract_2LMM_LLR_polymer.tex
\begin{quote}  
{\bf Abstract}\\  
A novel framework has recently been proposed for designing 
the molecular structure of chemical compounds
with a desired chemical property  
using both artificial neural networks 
 and mixed integer linear programming.  
In this paper, we design a new method
for inferring a polymer based on the framework.  
For this, we introduce  a new way of representing a polymer as a form of monomer
and define new descriptors that feature the structure of polymers. 
We also use linear regression as a building block of constructing 
a prediction function in the framework. 
 The results  of our computational experiments 
 reveal a set of chemical properties on polymers to which 
 a prediction function constructed with linear regression  performs well. 
We also observe that the proposed method can infer polymers 
with up to 50 non-hydrogen atoms in a monomer form.

\noindent 
{\bf Keywords: } Machine Learning, Linear Regression, Integer Programming,
Polymers, 
Cheminformatics, Materials Informatics,
QSAR/QSPR, Molecular Design. 


\end{quote}

%% file: Introduction_2LMM_LLR_polymer_arXiv_b.tex
\section{Introduction}\label{sec:introduction}

\noindent {\bf Background~}
In recent years, molecular design has received a great deal of attention
from various research fields such as chemoinformatics, bioinformatics,
and materials informatics \cite{Tetko20,Xia17,Sanchez18}.
In particular,
extensive studies have been done for molecular design 
using \emph{artificial neural networks} (ANNs).
Various ANN models have been applied in these studies,
which include recurrent neural networks~\cite{Segler18,Yang17}, 
variational autoencoders~\cite{Gomez18}, 
grammar variational autoencoders~\cite{Kusner17},
generative adversarial networks~\cite{DeCao18,Prykhodko19},
and invertible flow models~\cite{Madhawa19,Shi20}.
Many of these studies employ graph convolution techniques~\cite{Kipf16}
to effectively handle molecules represented as chemical graphs.

Molecular design has also been studied for many years in chemoinformatics,
under the name of
\emph{inverse quantitative structure activity relationship}
(inverse QSAR).
The purpose of this framework is to seek for chemical structures having
desired chemical activities under some constraints,
where the task of prediction of chemical activities from their chemical
structures is referred to as QSAR (or, forward QSAR).
In both forward and inverse QSAR,
chemical structures are represented as undirected graphs (chemical graphs).
Then, chemical graphs are transformed into vectors of real or integer numbers,
which are called \emph{descriptors} in chemoinformatics and vectors
correspond to \emph{feature vectors} in machine learning.
One of the typical approaches to inverse QSAR is 
to infer feature vectors from given chemical activities and constraints
and then reconstruct chemical graphs from these feature
vectors~\cite{Miyao16,Ikebata17,Rupakheti15}.
However, the  reconstruction itself is a challenging task because
it is known to be NP-hard (i.e., theoretically intractable)~\cite{AFJS12}.
Such a difficulty is also suggested from a huge size of chemical
graph space.
For example, the number of chemical graphs with up to
30 atoms (vertices) {\tt C}, {\tt N}, {\tt O}, and  {\tt S}
may  exceed~$10^{60}$~\cite{BMG96}. 
Due to this inherent difficulty, most methods for inverse QSAR,
including recent ANN-based ones,
do not guarantee optimal or exact solutions.

The targets of most of the inverse QSAR methods and ANN-based molecular design
methods had been small chemical compounds.
On the other hand, it is known that
macromolecules, especially \emph{polymers}, have also a wide range 
of applications in both medical science and
material science~\cite{Connor17,Miccio20}.
Accordingly, several studies have recently been done on
computational design of polymers~\cite{Kumar19,Wu19}.
However, it was pointed out that very few studies addressed the representation
of polymer structures~\cite{David20}, 
and thus the development of novel and useful
representation methods for polymers remains a challenge.

\begin{figure}[!ht]  \begin{center}
\includegraphics[width=.77\columnwidth]{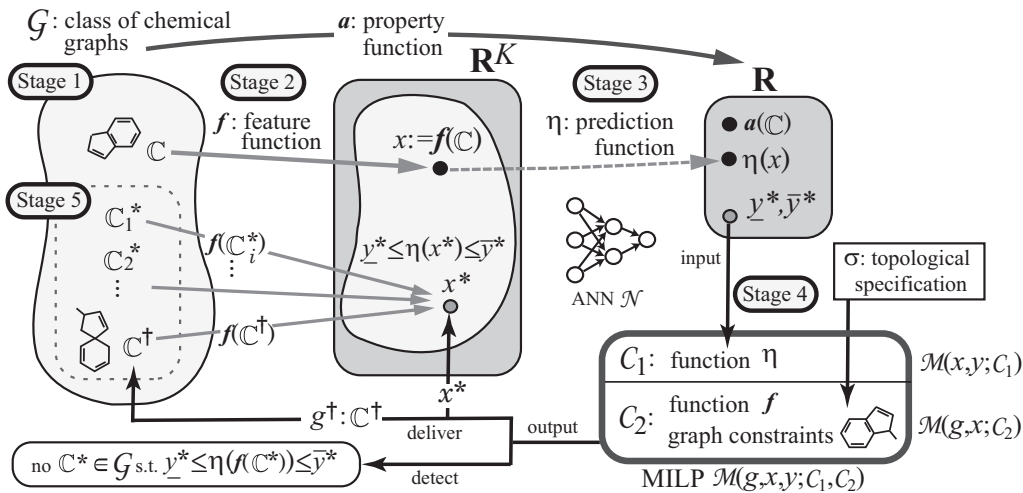}
\end{center} \caption{An illustration of a framework for inferring
a set of chemical graphs $\C^*$.   } 
\label{fig:framework}  \end{figure}    

\smallskip
\noindent {\bf Framework~}
Akutsu and Nagamochi~\cite{AN19} proved that  
the computation process of a given ANN can be simulated
with a mixed integer linear programming (MILP).
Based on this,
a novel  framework for inferring chemical graphs has been developed 
and revised~\cite{ACZSNA20,ZZCSNA20,ZAHZNA21}, 
as illustrated in Figure~\ref{fig:framework}. 
It constructs a prediction function in the first phase and
infers a chemical graph in the second phase. 
The first phase of the framework consists of three stages.
In Stage~1, we choose a chemical property $\pi$ and a class $\mathcal{G}$ 
of graphs, where a property function
$a$ is defined so that $a(\C)$ is the value of $\pi$  for a compound $\C\in \mathcal{G}$,
and collect a data set $D_{\pi}$ of chemical graphs in  $\mathcal{G}$ 
such that $a(\C)$ is available for every $\C \in D_{\pi}$.
In Stage~2, we introduce a feature function $f: \mathcal{G}\to \mathbb{R}^K$ 
for a positive integer $K$.    
In Stage~3, we construct a prediction function $\eta$ 
with an ANN $\mathcal{N}$ that,  
given a   vector  $x\in \mathbb{R}^K$, 
returns a value $y=\eta(x)\in \mathbb{R}$    
so that $\eta(f(\C))$ serves as a predicted value
to the real value $a(\C)$ of $\pi$ for each $\C\in D_\pi$.  
Given two reals $\underline{y}^*$ and $\overline{y}^*$
 as an interval  for a  target chemical value,
the  second phase infers  chemical graphs $\C^*$
with $\underline{y}^*\leq \eta(f(\C^*))\leq \overline{y}^*$ in the next two stages. 
We have obtained a  feature function $f$ and a  prediction function $\eta$
and call an additional constraint on the substructures of target chemical graphs 
a {\em topological specification}. 
In Stage~4, we prepare the following two  MILP formulations: 
\begin{enumerate}[nosep]
\item[-]
 MILP $\mathcal{M}(x,y;\mathcal{C}_1)$
with a set $\mathcal{C}_1$ of linear constraints on variables $x$ and $y$
(and some other auxiliary variables) 
  simulates the  process of computing $y:=\eta(x)$ from a vector $x$; and
\item[-]
 MILP $\mathcal{M}(g,x;\mathcal{C}_2)$
with a set $\mathcal{C}_2$ of linear constraints on  variable  $x$ and
 a variable vector  $g$ that represents a chemical graph $\C$
(and some other auxiliary variables)  
  simulates the  process of computing $x:=f(\C)$ from a chemical graph $\C$
and chooses a chemical graph $\C$ that satisfies the given topological specification
$\sigma$. 
\end{enumerate} 
Given an interval with $\underline{y}^*,\overline{y}^* \in \mathbb{R}$,
 we solve the combined
MILP $\mathcal{M}(g,x,y;\mathcal{C}_1,\mathcal{C}_2)$
to find  a feature vector $x^*\in \mathbb{R}^K$
 and a chemical graph $\C^{\dagger}$  with the specification
$\sigma$ such that $f(\C^\dagger)=x^*$ and  
$\underline{y}^*\leq \eta(x^*) \leq \overline{y}^*$
(where if the MILP instance is infeasible then this suggests that there 
does  not exist such a desired chemical graph).
In Stage~5,  we generate other  chemical graphs $\C^*$
such that $\underline{y}^*\leq \eta(f(\C^*))\leq \overline{y}^*$
 based on the output chemical graph $\C^\dagger$.

MILP formulations required in Stage~4 have been designed   
for chemical compounds with 
cycle index 0 (i.e., acyclic) \cite{ZZCSNA20,AZSSSZNA20},
cycle index 1~\cite{IAWSNA20} and 
cycle index 2~\cite{ZCSNA20}, where no sophisticated topological  specification
was available yet. 
Azam~et~al.~\cite{AZSSSZNA20} 
 introduced a restricted class of acyclic graphs 
 that is  characterized  by an integer ${\rho}$, called
 a ``branch-parameter''
such that the restricted class still covers  most of the acyclic
chemical compounds in the database. 
 Akutsu and Nagamochi~\cite{AN20} 
extended the idea to define a restricted class of  cyclic graphs,
called ``${\rho}$-lean cyclic graphs'' 
and introduced a set of flexible rules for  describing a topological specification.
Recently, Tanaka et~al.~\cite{TZAHZNA21} 
(resp.,  Zhu~et~al.~\cite{ZAHZNA21}) used a decision tree 
 (resp.,  linear regression)  
to construct a prediction function $\eta$ in Stage~3 in the framework
and derived an MILP $\mathcal{M}(x,y;\mathcal{C}_1)$ 
that simulates the computation process of  a decision tree
 (resp.,  linear regression).

\smallskip
\noindent {\bf Two-layered Model~}
 Shi et~al.~\cite{SZAHZNA21} 
  proposed a method, called a {\em two-layered model}  for representing
the feature of a chemical graph in order to deal with an arbitrary graph 
in the framework. 
In the two-layered model, a chemical graph  $\C$
with a parameter ${\rho}\geq 1$ is regarded as two parts:
the exterior and the interior of the hydrogen-suppressed chemical graph  
$\anC$ obtained from  $\C$ by removing hydrogen.
The exterior  consists of  maximal acyclic induced subgraphs with height
at most ${\rho}$ in $\anC$ and
the interior is the connected subgraph of $\anC$ 
obtained by ignoring the exterior.  
 Shi et~al.~\cite{SZAHZNA21}  defined 
 a feature vector $f(\C)$ of a chemical graph  $\C$
to be a combination of the frequency of adjacent atom pairs in the interior  and
the frequency of chemical acyclic graphs among the set of  chemical rooted trees $T_u$
rooted at interior-vertices $u$. 
Tanaka~et~al.~\cite{TZAHZNA21} constructed a prediction function  
with a decision tree by using the feature vector by  Shi et~al.~\cite{SZAHZNA21}. 
Recently, Zhu~et~al.~\cite{ZAHZNA21} 
extended the model to treat   chemical elements of multiple valence
and chemical compounds with cations and anions. 

\smallskip
\noindent {\bf Contribution~} 
In order to extend our MILP-based framework for designing novel polymers,
we modify the method due to  Zhu~et~al.~\cite{ZAHZNA21}.
For this, we introduce a new way of representing a polymer as a form of monomer
and define new descriptors that feature the structure of polymers. 
We modify the MILP formulation proposed by  Zhu~et~al.~\cite{ZAHZNA21}
due to the change  of feature function $f$ 
 (the detail of the MILP $\mathcal{M}(g,x;\mathcal{C}_2)$ can be found in  
 Appendix~\ref{sec:full_milp}).  
To generate target chemical graphs $\C^*$ in Stage~5,
we also use and modify the dynamic programming algorithm 
due to Zhu~et~al.~\cite{ZAHZNA21}.

We implemented the framework based on the refined two-layered model 
and a prediction function by linear regression. 
A polymer was inferred by using the framework for the first time in this paper,
where Tanaka~et~al.~\cite{TZAHZNA21} studied constructing a prediction function
with a decision tree for some polymer properties but 
have not argued topological specification of polymers and 
inference of a polymer. 
 The results  of our computational experiments 
 reveal a set of chemical properties on polymers to which 
 a prediction function constructed  with linear regression on 
  our feature function  performs well. 
We also observe that  the proposed method can infer a polymer 
with  up to 50 non-hydrogen atoms in a monomer form.

The paper is organized as follows.  
Section~\ref{sec:preliminary} introduces some notions on graphs,
 a modeling of chemical compounds and 
 define a new monomer representation of polymers.  
Section~\ref{sec:2LM} describes the two-layered model for polymers.
Section~\ref{sec:experiment} reports the results on some computational 
experiments conducted for eight chemical properties on polymers such as 
glass transition
and 
experimental amorphous density.
Section~\ref{sec:conclude} makes some concluding remarks.   

Some technical details are given in Appendices:   
Appendix~\ref{sec:linear_regression} for the idea of linear regression
and an MILP $\mathcal{M}(x,y;\mathcal{C}_1)$ formulated by Zhu~et~al.~\cite{ZAHZNA21}
that simulates a process of computing a prediction function constructed by linear regression; 
Appendix~\ref{sec:descriptor} for  all descriptors in our feature function on polymers; 
 Appendix~\ref{sec:specification}  for a full description of 
a topological specification;
Appendix~\ref{sec:test_instances}   for the detail of  test instances  
used in our computational experiment for Stages~4 and 5; and 
Appendix~\ref{sec:full_milp}
  for the details of our MILP formulation  $\mathcal{M}(g,x;\mathcal{C}_2)$.
Note that the modification of the dynamic programming algorithm
is not given in  Appendices
because it is slight and straightforward.
   
%

%% file: Preliminary_2LMM_polymer.tex
\section{Preliminary}\label{sec:preliminary}

This section  introduces some notions and terminologies on graphs,
  modeling of chemical compounds and our choice of descriptors. 
 
Let $\mathbb{R}$, $\mathbb{Z}$  and $\mathbb{Z}_+$ 
denote the sets of reals, integers and non-negative integers, respectively.
For two integers $a$ and $b$, let $[a,b]$ denote the set of 
integers $i$ with $a\leq i\leq b$.

\bigskip\noindent
{\bf  Graph} 
Given a  graph $G$, let $V(G)$ and $E(G)$ denote the sets
of vertices and edges, respectively.     
For a subset $V'\subseteq V(G)$ (resp., $E'\subseteq E(G))$ of
a graph $G$, 
let $G-V'$ (resp., $G-E'$) denote the graph obtained from $G$
by removing the vertices in $V'$ (resp.,  the edges in $E'$),
where we remove all edges incident to a vertex in $V'$ in $G-V'$. 
An edge subset $E'\subseteq E(G)$ in a connected graph $G$ is called
{\em separating} (resp., {\em non-separating})
if $G-E'$  becomes disconnected (resp., $G-E'$ remains connected). 
The {\em rank}  $\mathrm{r}(G)$ of a graph $G$  is defined to be 
the minimum $|F|$ of an edge subset $F\subseteq E(G)$
such that $G-F$ contains no cycle, 
where  $\mathrm{r}(G)=|E(G)|-|V(G)|+1$ for a connected graph $G$.    
Observe that   $\mathrm{r}(G-E')=\mathrm{r}(G)-|E'|$ holds
for any non-separating edge subset $E'\subseteq E(G)$. 
An edge $e\in E(G)$ in a connected graph $G$
  is called a {\em bridge} if $\{e\}$ is separating.
For a connected cyclic graph $G$, an edge $e$ is called a {\em core-edge} if
it is in a cycle of $G$ or is a bridge $e=u_1u_2$ such that
each of the connected graphs $G_i$, $i=1,2$ of $G-e$ contains a cycle. 
A vertex incident to a core-edge is called a {\em core-vertex} of $G$. 
A path with two end-vertices $u$ and $v$ is called a {\em $u,v$-path}. 
A set $F$ of edges in $G$ is called a {\em circular set} if
$G$ contains a cycle $C$ that contains all edges in $F$ and
for every edge $e\in F$, $F\setminus \{e\}$ is the set of all bridges $e'\in F$
in the graph $G-e$.  
 
 We define a {\em rooted} graph to be
 a graph with a  designated vertex, called a {\em root}. 
 %
%
 For a graph $G$ possibly with a root, 
 a {\em leaf-vertex} is defined to be a non-root vertex  with degree 1.
 We  call  the edge $uv$ incident to a leaf vertex $v$ a {\em leaf-edge},
 and denote by $\Vleaf(G)$ and $\Eleaf(G)$
  the sets of leaf-vertices and leaf-edges  in $G$, respectively.
 For a graph  or a rooted graph $G$,
 we define graphs $G_i, i\in \mathbb{Z}_+$ obtained from $G$
 by removing the set of leaf-vertices $i$ times so that
\[ G_0:=G; ~~ G_{i+1}:=G_i - \Vleaf(G_i), \]
where we call a vertex $v$ a {\em tree vertex} if $v\in \Vleaf(G_i)$
for some $i\geq 0$. 
Define the {\em height} $\h(v)$ of each tree vertex $v\in \Vleaf(G_i)$
to be $i$; and 
$\h(v)$ of each non-tree vertex $v$ adjacent to a tree vertex 
to be $\h(u)+1$ for the maximum $\h(u)$ of a tree vertex $u$ adjacent to $v$,
where we do not define height of any non-tree vertex not adjacent to any tree vertex. 
We call a vertex $v$ with $\h(v)=k$ a {\em leaf $k$-branch}.
The {\em height} $\h(T)$ of a rooted tree $T$ is defined
to be the maximum of $\h(v)$ of a vertex $v\in V(T)$. 
For an integer $k\geq 0$, we call a  rooted tree $T$
 {\em $k$-lean} if $T$ has at most one leaf $k$-branch.
For an unrooted cyclic graph $G$, we regard that
the set of non-core-edges in $G$ induces
a collection $\mathcal{T}$ of trees each of which is rooted at a core-vertex,
where we call $G$  {\em $k$-lean} if each of the rooted trees in $\mathcal{T}$ 
is $k$-lean. 
 
\subsection{Modeling of Chemical Compounds}\label{sec:chemical_model}

We review a modeling of chemical compounds (monomers) 
and introduce a new way of representing a polymer as a form of monomer.  
  
To represent a chemical compound, 
we introduce a set  of   chemical elements such as 
  {\tt H} (hydrogen),   
 {\tt C} (carbon), {\tt O} (oxygen), {\tt N} (nitrogen)  and so on.
 To distinguish a chemical element $\ta$ with multiple valences such as {\tt S} (sulfur),
 we denote a chemical element $\ta$ with a valence $i$ by $\ta_{(i)}$,
 where we do not use such a suffix $(i)$ 
 for a chemical element $\ta$ with a unique valence. 
Let $\Lambda$ be a set of chemical elements $\ta_{(i)}$.
For example,  $\Lambda=\{\ttH,  \ttC, \ttO, \ttN, \ttP, \ttS_{(2)}, \ttS_{(4)}, \ttS_{(6)}\}$. 
Let $\val: \Lambda\to [1,6]$ be a valence function.
For example, $\val(\ttH)=1$, $\val(\ttC)=4$, $\val(\ttO)=2$, $\val(\ttP)=5$,
$\val(\ttS_{(2)})=2$, $\val(\ttS_{(4)})=4$ and $\val(\ttS_{(6)})=6$.
 For each  chemical element $\ta\in \Lambda$, 
let $\mathrm{mass}(\ta)$  denote the mass of~$\ta$.

A chemical compound  is represented by a {\em chemical graph} defined to be
a tuple $\C=(H,\alpha,\beta)$  of
  a simple, connected undirected graph $H$ and  
    functions   $\alpha:V(H)\to \Lambda$  and  $\beta: E(H)\to [1,3]$.
The set of atoms and the set of bonds in the compound 
are represented by the vertex set $V(H)$ and the edge set $E(H)$, respectively.
The chemical element assigned to a vertex $v\in V(H)$
is represented by $\alpha(v)$ and 
 the bond-multiplicity  between two adjacent vertices  $u,v\in V(H)$
is represented by $\beta(e)$ of the edge $e=uv\in E(H)$.
We say that two tuples $(H_i,\alpha_i,\beta_i), i=1,2$ are
{\em isomorphic} if they admit an isomorphism $\phi$,
i.e.,  a bijection $\phi: V(H_1)\to V(H_2)$
such that
 $uv\in E(H_1), \alpha_1(u)=\ta, \alpha_1(v)=\tb, \beta_1(uv)=m$
 $\leftrightarrow$  
 $\phi(u)\phi(v) \in E(H_2), \alpha_2(\phi(u))=\ta, 
 \alpha_2(\phi(v))=\tb, \beta_2(\phi(u)\phi(v))=m$. 
 When $H_i$ is rooted at a vertex $r_i,  i=1,2$,
 $(H_i,\alpha_i,\beta_i), i=1,2$ are
{\em rooted-isomorphic} (r-isomorphic) if 
they admit an isomorphism $\phi$ such that $\phi(r_1)=r_2$.

 For a notational convenience, we  use
 a function $\beta_\C: V(H)\to [0,12]$ for a chemical graph $\C=(H,\alpha,\beta)$,
  such that $\beta_\C(u)$ means the sum of bond-multiplicities
 of edges incident to a vertex $u$; i.e., 
\[ \beta_\C(u) \triangleq \sum_{uv\in E(H) }\beta(uv) 
\mbox{ for each vertex $u\in V(H)$.}\]
For each vertex $u\in V(H)$, 
 define the {\em electron-degree} $\eledeg_\C(u)$  to be 
\[  \eledeg_\C(u) \triangleq  \beta_\C(u) - \val(\alpha(u)). \]
For each  vertex $u\in V(H)$, let $\deg_\C(v)$ denote 
the number of vertices adjacent to $u$ in $\C$. 
  
  For a chemical   graph  $\C=(H,\alpha,\beta)$, 
  let  $V_{\ta}(\C)$, $\ta\in \Lambda$
   denote the set of  vertices $v\in V(H)$ such that $\alpha(v)=\ta$ in $\C$
  and define the {\em hydrogen-suppressed chemical graph} $\anC$ 
to be  the graph obtained from $H$ by
  removing all the vertices $v\in \VH(\C)$.
  

\begin{figure}[h!] \begin{center}
\includegraphics[width=.96\columnwidth]{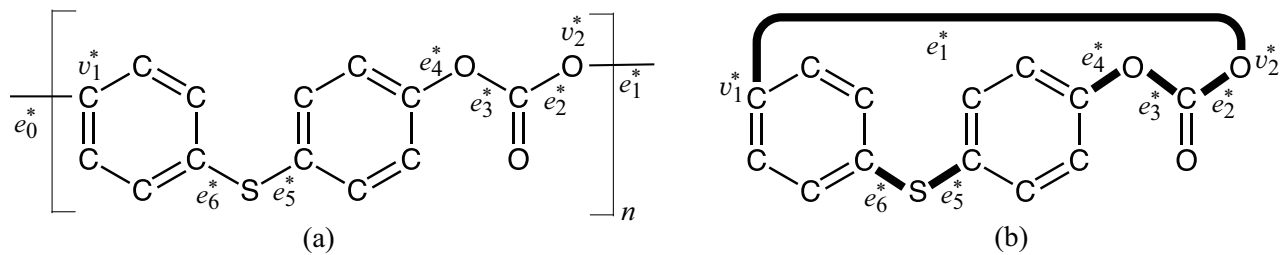}
\end{center}
\caption{(a) A repeating unit of polymer: thioBis(4-phenyl)carbonate,
where $e^*_0$ and $e^*_1$ are the connecting-edges 
and $v^*_1$ and $v^*_2$ are the connecting-vertices;
(b) A monomer form of the polymer in (a), where 
the link-edges are depicted with thick lines 
and $v^*_1$ and $v^*_2$ are the connecting-vertices.}
\label{fig:polymer_example}  \end{figure} 

\bigskip
\noindent
{\bf Polymers~}
In this paper, we treat a polymer that is a linear concatenation 
of a single repeating unit with  two connecting-edges of $e^*_0$ and $e^*_1$
such that two adjacent units in the concatenation
are joined with the connecting-edges. 
We call   the two vertices incident 
to the  two connecting-edges the {\em connecting-vertices}.  
 Figure~\ref{fig:polymer_example}(a) illustrates 
 an example of a repeating unit of such a polymer,
 where $v^*_1$ and $v^*_2$ are the connecting-vertices. 
 
Tanaka~et~al.~\cite{TZAHZNA21} proposed a modeling
of a polymer as a monomer with no connecting-edges
by introducing an artificial chemical element ${\tt a}^*$
to which the original two connecting-edges 
of a repeating unit become newly incident.
When the number of repeating units in a polymer is extremely large, 
other edges in the repeating unit may have a similar role with 
the connecting-edges.  
For example,  edge $e^*_2$ of the repeating unit 
in Figure~\ref{fig:polymer_example}(a) 
can serve as the connecting-edges of a different repeating unit
by splitting  $e^*_2$ into two edges  and merging $e^*_0$ and $e^*_1$
  into a single edge.

To take this into consideration,  this paper introduces a new way of
representing a polymer as a monomer form. 
We call an edge $e$ in a repeating unit of a polymer 
a  {\em link-edge} if it is passed by every path 
between the connecting-edges $e^*_0$ and $e^*_1$.
For example, the link-edges  in the  repeating unit in  Figure~\ref{fig:polymer_example}(a) 
 are given by $e^*_2,e^*_3,\ldots,e^*_6$. 
To represent a polymer as a monomer, 
we regard the two connecting-edges $e^*_0$ and $e^*_1$ as a single edge $e^*_1$,
as illustrated in Figure~\ref{fig:polymer_example}(b). 
We call the resulting chemical graph the {\em monomer representation},
where we also call the edge $e^*_1$ a  {\em link-edge} in the representation.
We still call the vertices incident to $e^*_1$ the {\em connecting-vertices}
and distinguish  them from other vertices because
a polymer that is synthesized from a specified repeating unit actually
may end with  the  connecting-vertices.  
(A polymer of a cyclic sequence of a repeating unit that has no particular ends
can be modeled as our monomer representation with no connecting-vertices.)
In what follows, a  polymer  is represented by the   monomer representation $\C$,
and the set of link-edges in $\C$ is denoted by $\Elnk(\C)$.
Note that the set $\Elnk(\C)$ is a circular set in $\C$.


%% file: Two-layered_Model_polymer.tex
\section{Two-layered Model}\label{sec:2LM}

This section reviews the two-layered model 
proposed  by Zhu~et~al.~\cite{ZAHZNA21}  
and makes a necessary modification
 so as to apply it to the case of polymers.
 
 Let  $\C=(H,\alpha,\beta)$ be a chemical graph
 and  ${\rho}\geq 1$ be an integer, which we call a {\em branch-parameter}.
 
  A {\em two-layered model} of $\C$ is a partition of
 the hydrogen-suppressed chemical graph $\anC$ into
 an ``interior'' and an ``exterior'' in the following way. 
 We call a vertex $v\in V(\anC)$
   (resp., an edge $e\in E(\anC))$ of   $\C$
   an {\em exterior-vertex} (resp.,    {\em exterior-edge}) if
    $\h(v)< {\rho}$ (resp., $e$ is incident to an  exterior-vertex)
and denote the sets of exterior-vertices and exterior-edges 
by $V^\ex(\C)$ and $E^\ex(\C)$, respectively, 
and denote  $V^\inte(\C)=V(\anC)\setminus  V^\ex(\C)$ and 
$E^\inte(\C)=E(\anC)\setminus E^\ex(\C)$, respectively.
We call a vertex in $V^\inte(\C)$ (resp.,   an edge in $E^\inte(\C)$) 
   an {\em interior-vertex} (resp.,    {\em interior-edge}). 
 The set  $E^\ex(\C)$ of  exterior-edges forms 
a collection of connected graphs each of which is
regarded as a rooted tree $T$ rooted at 
a vertex $v\in V(T)$ with the maximum $\h(v)$. 
Let $\mathcal{T}^\ex(\anC)$ denote 
the set of these chemical rooted trees in $\anC$. 
The {\em interior} of $\C$ is defined to be the subgraph
 $(V^\inte(\C),E^\inte(\C))$ of $\anC$. 

\begin{figure}[h!] \begin{center}
\includegraphics[width=.90\columnwidth]{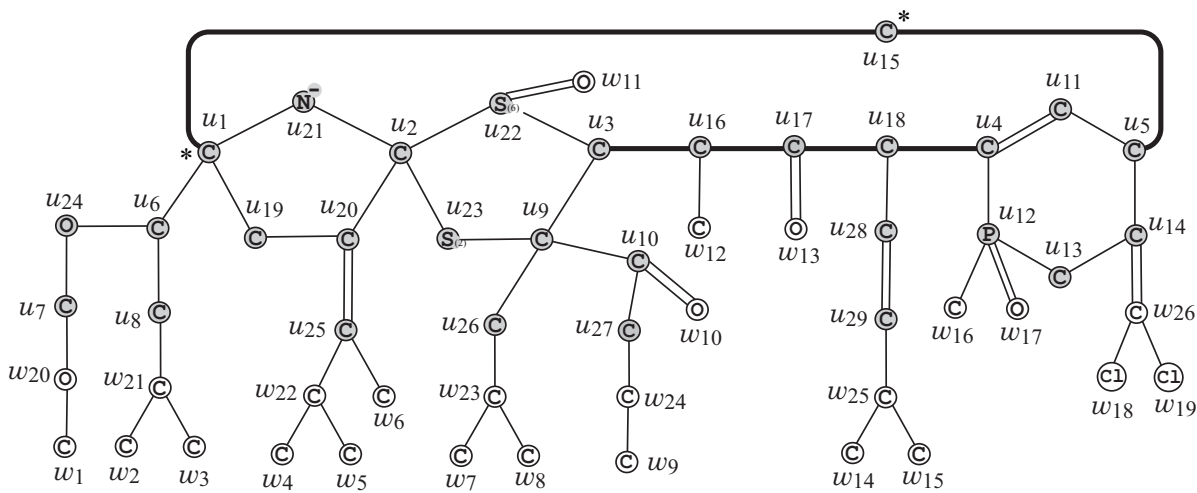}
\end{center} \caption{An illustration 
of  the hydrogen-suppressed  monomer representation $\anC$  
obtained from a polymer $\C$ by removing all the 
 hydrogens, 
where the link-edges are depicted with thick lines and 
$V^\ex(\C)=\{w_i \mid i\in [1,26]\}$ and
$V^\inte(\C)=\{u_i \mid i\in [1,29]\}$    for  ${\rho}=2$
 and the connecting-vertices are marked with asterisks.  
 }
\label{fig:example_polymer} \end{figure}

Differently from standard monomers, we distinguish the link-edges in the monomer form
of a polymer from other edges in order to feature the topological structure of the polymer. 
Figure~\ref{fig:example_polymer}
 illustrates an example of a hydrogen-suppressed polymer $\anC$
 with $\Elnk(\C)=\{u_1u_{15},$ $u_5u_{15}, u_3u_{16}, u_{16}u_{17},  u_{17}u_{18}, u_4u_{18}\}$.

For a branch-parameter ${\rho}=2$, 
the interior of  the chemical graph $\anC$ in Figure~\ref{fig:example_polymer} 
is obtained by removing the set of vertices with degree 1 ${\rho}=2$ times; i.e., 
first remove  the set  $V_1=\{w_1,w_2,\ldots,w_{19}\}$ of vertices of degree 1 in $\anC$ 
and then remove  the set
 $V_2=\{w_{20},w_{16},\ldots,w_{26}\}$ of vertices of degree 1 in $\anC-V_1$,
 where the removed vertices become the exterior-vertices of $\anC$. 

\begin{figure}[h!] \begin{center}
\includegraphics[width=.90\columnwidth]{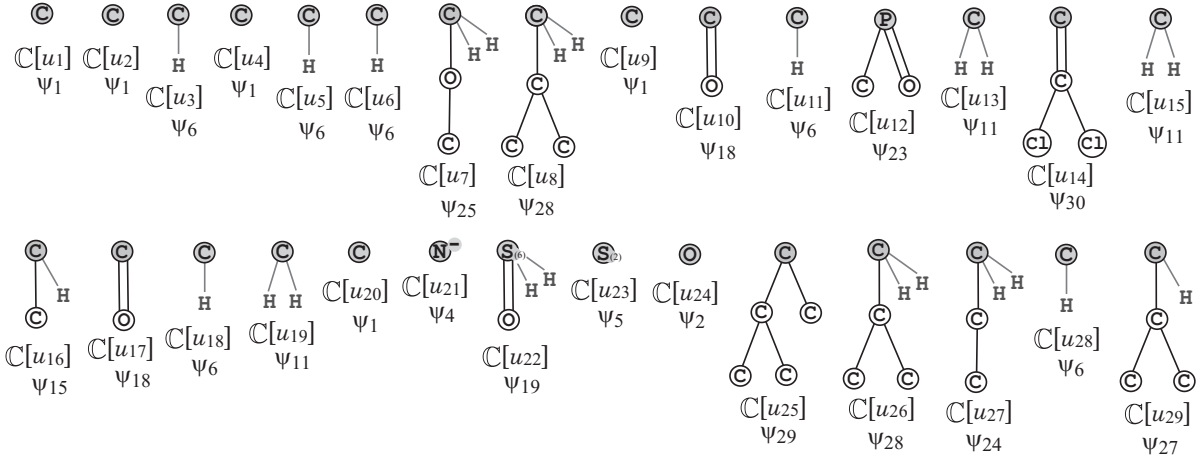}
\end{center} \caption{
The set $\mathcal{T}^\ex(\C)$ of  2-fringe-trees  $\C[u_i], i\in [1,29]$ of 
$\C$ with $\anC$ 
in Figure~\ref{fig:example_polymer}, 
where the root of each tree is depicted with a gray circle and
 the hydrogens attached to non-root vertices are omitted in the figure.  
 }
\label{fig:example_fringe-tree} \end{figure}

For each interior-vertex $u\in V^\inte(\C)$,
let $T_u\in \mathcal{T}^\ex(\anC)$ denote the chemical tree rooted at $u$
(where possibly $T_u$ consists of vertex $u$) and 
define the {\em $\rho$-fringe-tree} $\C[u]$ to be  
the chemical rooted tree obtained from $T_u$ by putting back
 the hydrogens originally attached with $T_u$ in $\C$. 
Let $\mathcal{T}(\C)$ denote the set of $\rho$-fringe-trees 
$\C[u], u \in V^\inte(\C)$. 
Figure~\ref{fig:example_fringe-tree}  illustrates
the set  $\mathcal{T}(\C)=\{\C[u_i]\mid i\in [1,29]\}$ of the 2-fringe-trees 
  of the example $\C$ in Figure~\ref{fig:example_polymer}. 

\smallskip
\noindent {\bf Feature Function~} 
 The feature of an  interior-edge $e=uv\in E^\inte(\C)$ 
 such that $\alpha(u)=\ta$, $\deg_{\anC}(u)=d$, 
 $\alpha(v)=\tb$, $\deg_{\anC}(v)=d'$  and $\beta(e)=m$  is represented by 
 a tuple $(\ta d, \tb d', m)$, which is called the {\em edge-configuration}
   of the edge $e$, where   we call the tuple $(\ta, \tb, m)$ 
 the {\em adjacency-configuration} of the edge $e$.
 
For an integer $K$, a feature vector $f(\C)$ of a chemical graph $\C$
is defined by a {\em feature function} $f$ that consists of $K$ descriptors. 
We call  $\RK$ {\em  the feature space}.

 Tanaka et~al.~\cite{TZAHZNA21}   defined  a feature vector $f(\C)\in \RK$  
to be a combination of the frequency 
of edge-configurations of   the interior-edges  and
the frequency of chemical rooted trees among the set 
of  chemical rooted trees $\C[u]$
over all interior-vertices $u$. 
Zhu~et~al.~\cite{ZAHZNA21} additionally included 
two descriptors that feature  the leaf-edges and the rank of a chemical graph. 
In this paper, we further introduce new descriptors that features
the link-edges in the monomer representation of polymers
(see  Appendix~\ref{sec:descriptor} 
for  all descriptors in our feature function on polymers).
 Note that introduction of new descriptors requires us to modify 
 the subsystem of simulating the computation process of a feature function $f$
 in an MILP $\mathcal{M}(x,y;\mathcal{C}_1)$.
 We use the same MILP formulation used by 
Zhu~et~al.~\cite{ZAHZNA21} for $\mathcal{M}(x,y;\mathcal{C}_1)$
by making a necessary modification
(see  Appendix~\ref{sec:full_milp}
  for the details of our MILP formulation  $\mathcal{M}(g,x;\mathcal{C}_2)$).

\smallskip
\noindent {\bf Topological Specification~} 
  Tanaka et~al.~\cite{TZAHZNA21}  also introduced
a set of rules for describing a topological specification
in the following way:      
\begin{enumerate}[nosep]
\item[(i)]
a {\em seed graph} $\GC$ as an  abstract form of  a target chemical graph $\C$;
\item[(ii)]
 a set $\mathcal{F}$ of chemical rooted trees  as candidates
 for a tree  $\C[u]$ rooted at each interior-vertex $u$ in $\C$; 
and 
\item[(iii)]
lower and upper bounds on the number of components 
 in a target chemical graph such as  chemical elements, 
double/triple bonds and the interior-vertices in $\C$. 
\end{enumerate} 

\begin{figure}[h!] \begin{center}
\includegraphics[width=.98\columnwidth]{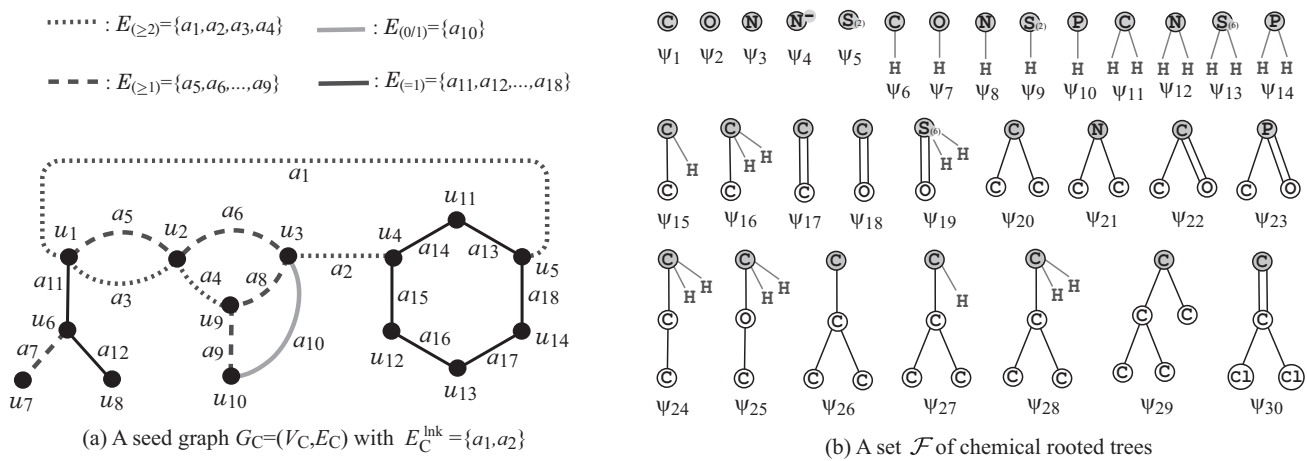}
\end{center} \caption{
(a) A seed graph $\GC$ with $\EC^\lnk=\{a_1,a_2\}$, 
where the vertices in $\VC$ are depicted with gray circles,
the edges in $\Et$ are depicted with dotted lines,
the edges in $\Ew$ are depicted with dashed lines,
the edges in $\Ez$ are depicted with gray bold lines and  
the edges in $\Eew$ are depicted with black solid lines;
(b) A set $\mathcal{F}=\{\psi_1,\psi_2,\ldots,\psi_{30}\}\subseteq
\mathcal{F}(D_\pi)$ of 30 chemical rooted trees
$\psi_i, i\in [1,30]$, where the root of each tree is depicted with a gray circle, 
where  the hydrogens attached to non-root vertices are omitted in the figure.    }
\label{fig:specification_example_1} \end{figure}  

\begin{figure}[h!] \begin{center}
\includegraphics[width=.90\columnwidth]{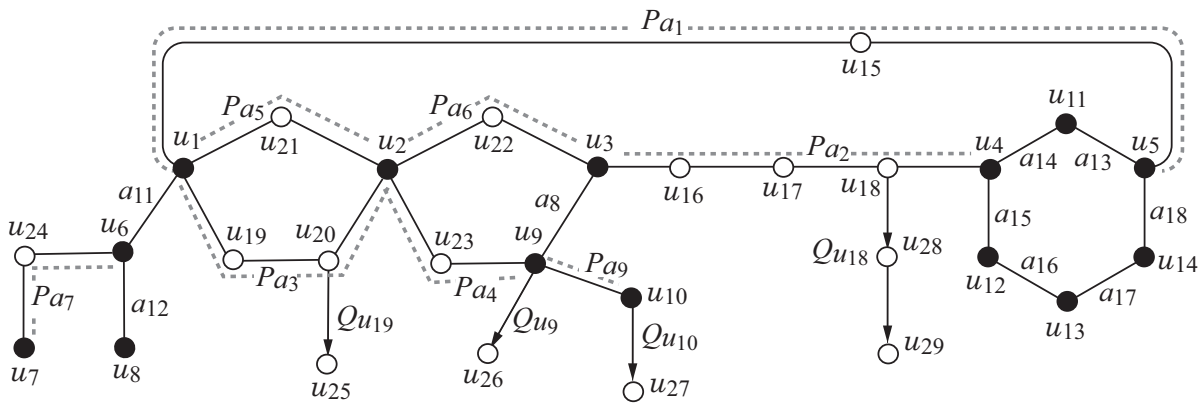}
\end{center} \caption{
A graph  obtained from the seed graph $\GC$  
in Figure~\ref{fig:specification_example_1}(a), where each path $Q_u$ rooted at a vertex $u$ is
depicted with arrows and the vertices newly introduced from $\GC$ are depicted with white circles.   
 }
\label{fig:test_subgraph_polymer} \end{figure}

Figure~\ref{fig:specification_example_1}(a) and (b)
 illustrate  examples of  a  seed graph  $\GC$ and 
 a set $\mathcal{F}$ of chemical rooted trees, respectively. 
 Given a seed graph $\GC$, 
 the interior of   a target chemical graph $\C$ is constructed
 from $\GC$ by replacing some edges $a=uv$ 
 with paths $P_a$ between the end-vertices
 $u$ and $v$ and by attaching new paths $Q_v$ to some vertices $v$.  
%
For example, the chemical graph $\anC$ 
in Figure~\ref{fig:example_polymer} is constructed
from the seed  graph  $\GC$ in Figure~\ref{fig:specification_example_1}(a)
as follows.
\begin{enumerate}[nosep,  leftmargin=*]
\item[-]
First replace  nine edges
 $a_1=u_1 u_{5},  a_2=u_3 u_{4},  a_3=u_1 u_{2}, a_4=u_{2}u_{9}$,
 $a_5=u_1 u_{2}, a_6=u_{2}u_{3}, a_7=u_6 u_{7}, a_8=u_{3}u_{9}$
and $a_9=u_{9}u_{10}$ in  $\GC$ 
 with new paths  
$P_{a_1}=(u_1,u_{15},u_{5})$, 
$P_{a_2}=(u_{3},u_{16},u_{17},u_{18},u_{4})$,
$P_{a_3}=(u_{1},u_{19},u_{20},u_{2})$, 
 $P_{a_4}=(u_{2},u_{23},u_{9})$,
 $P_{a_5}=(u_{1},u_{21},u_{2})$, 
$P_{a_6}=(u_2,u_{22},u_{3})$, 
$P_{a_7}=(u_{6},u_{24},u_{7})$,
$P_{a_8}=(u_{3},  u_{9})=a_8$ and 
$P_{a_9}=(u_{9},u_{10})$,  
 respectively
 to obtain a subgraph $G_1$ of $\anC$. 
\item[-]
Next attach to this graph  $G_1$ three new paths 
$Q_{u_9}=(u_9,u_{26})$, 
$Q_{u_{10}}=(u_{10},u_{27})$, 
$Q_{u_{18}}=(u_{18},u_{28},u_{29})$ and 
$Q_{u_{20}}=(u_{20},u_{25})$
to obtain  
the interior of  $\anC$, as illustrated in  
Figure~\ref{fig:test_subgraph_polymer}. 
\item[-]
Finally  attach to the interior   29 trees selected from the set $\mathcal{F}$ 
and assign chemical elements and bond-multiplicities in the interior
to  obtain a chemical graph $\C$  in Figure~\ref{fig:example_polymer}. 
In Figure~\ref{fig:example_fringe-tree},
  $\psi_1\in \mathcal{F}$ is selected for $\C[u_i]$, $i\in\{1,2,4,9,20\}$.
  Similarly 
  $\psi_2$  for  $\C[u_{24}]$,
  $\psi_4$  for $\C[u_{21}]$, 
  $\psi_5$  for $\C[u_{23}]$, 
  $\psi_6$  for $\C[u_i]$, $i\in\{3,5,6,11,18,28\}$, 
   $\psi_{11}$    for $\C[u_i]$, $i\in\{13,15,19\}$,
   $\psi_{15}$    for $\C[u_{16}]$, 
   $\psi_{18}$    for $\C[u_i]$, $i\in\{10,17\}$,
   $\psi_{19}$    for $\C[u_{22}]$,
   $\psi_{23}$    for $\C[u_{12}]$,
   $\psi_{24}$    for $\C[u_{27}]$, 
   $\psi_{25}$   for $\C[u_{7}]$, 
   $\psi_{27}$    for $\C[u_{29}]$, 
   $\psi_{28}$   for $\C[u_i]$, $i\in\{8,26\}$,
   $\psi_{29}$   for $\C[u_{25}]$ 
   and 
    $\psi_{30}$  for $\C[u_{14}]$. 
\end{enumerate} 

\bigskip
Our definition of a topological specification is analogous with the one  by 
 Zhu~et~al.~\cite{ZAHZNA21}
   except for a necessary modification due to our polymer model 
   with link-edges 
(see Appendix~\ref{sec:specification} 
for a full description of topological specification).

%% file: Result_2LMM_LLR_Phase1_polymer.tex
\section{Computational Results}\label{sec:experiment}

We implemented our method of Stages~1 to 5 
for inferring chemical graphs under a given topological specification and
conducted experiments  to evaluate the computational efficiency. 
We executed the experiments on a PC with 
 Processor:  Core i7-9700 (3.0GHz; 4.7 GHz at the maximum) and 
Memory: 16 GB RAM DDR4. 

\medskip \noindent
{\bf Results on Phase~1.  }
We have conducted experiments of linear regression for 
ten chemical properties on polymers  among which
we report the following eight properties  to which the test  
 coefficient of determination ${\rm R}^2$ 
attains at least 0.76:
experimental amorphous density ({\sc AmD}),
dielectric constant ({\sc DeC}),  
heat capacity liquid ({\sc HcL}), 
heat capacity solid ({\sc HcS}), 
mol volume ({\sc MlV}),
permittivity ({\sc Prm}),
refractive index ({\sc RfId})  and 
glass transition({\sc Tg}).
All these  data sets are provided by Bicerano~\cite{Bicerano}, 
where we did not include any polymer whose chemical formula
could not be found by its name in the book. 
For property {\sc RfId}, we remove the following   polymer 
as an outlier  from the original data set: 
2-decyl-1$\underline{~}$4-butadiene  $\C$ with $a(\C)=0.4899$.

We implemented Stages~1, 2 and 3 in Phase~1 as follows.

\medskip \noindent
{\bf Stage~1.  }
We set  a graph class   $ \mathcal{G}$   to be
the set of all polymers with any graph structure, 
and set a branch-parameter ${\rho}$ to be 2. 
We represent a polymer as a monomer representation. 

For each of the properties,  
 we first select a set $\Lambda$ of chemical elements 
 and then collect  a  data set  $D_{\pi}$ on  the polymers 
 over the set $\Lambda$ of chemical elements.  
 To construct the data set $D_{\pi}$,
  we eliminated  chemical compounds such that 
  the monomer representation $\C$ that does not satisfy one of the following:  
  $\C$ is connected;  
   the number of non-hydrogen neighbors of each atom  $\C$ is  at most 4; and   
  the number of end-vertices of the linked-edges in  $\C$ is at least two
  (i.e., no self-loop is a link-edge in the monomer form). 
Since the observed values of property {\sc Prm} are
measured by different frequencies, 
we include an extra descriptor $\mathrm{fq}$  that represents the frequency
used for each polymer $\C_i\in D_\pi$ 
in our feature vector $f(\C_i)$.  

 Table~\ref{table:phase1a} shows
  the size and range of data sets   that 
 we prepared for each chemical property in Stage~1,
 where  we denote the following:  
\begin{enumerate}[nosep,  leftmargin=*]
\item[-] 
  $\Lambda$: the set of elements used in the data set $D_{\pi}$; 
 $\Lambda$ is one of the following six sets: 
 $\Lambda_1=\{\ttH, \ttC, \ttO, \ttN \}$; 
 $\Lambda_2=\{\ttH, \ttC, \ttO_{(1)}, \ttO_{(2)}, \ttN \}$; 
   $\Lambda_3=\{\ttH, \ttC, \ttO, \ttN, \ttCl \}$;
   $\Lambda_4=\{\ttH, \ttC, \ttO, \ttN,  \ttCl, \ttS_{(2)} \}$;   
   $\Lambda_5=\{\ttH, \ttC, \ttO, \ttN,  \ttCl, \ttS_{(2)}, \ttS_{(6)} \}$; and
   $\Lambda_6=\{\ttH, \ttC, \ttO_{(1)}, \ttO_{(2)}, \ttN,  \ttCl, \ttSi_{(4)}, \ttF \}$, 
 where ${\tt a}_{(i)}$ for a chemical element ${\tt a}$ and an integer $i\geq 1$ 
 means that  a chemical element ${\tt a}$ with valence $i$. 

\item[-] 
 $|D_{\pi}|$:  the size of data set $D_{\pi}$ over $\Lambda$
  for the property $\pi$.
  
\item[-]   $ \underline{n},~\overline{n} $:  
  the minimum and maximum  values of the number $n(\C)$ of non-hydrogen atoms in 
  the polymers $\C$ in $D_{\pi}$.
\item[-] $\underline{a},~\overline{a} $:  the minimum and maximum values
of $a(\C)$ for $\pi$ over  the polymers  $\C$ in  $D_{\pi}$.
\item[-]    $|\Gamma|$: 
the number of different edge-configurations
of interior-edges over the compounds in~$D_{\pi}$. 
\item[-]  $|\mathcal{F}|$: the number of non-isomorphic chemical rooted trees
 in the set of all 2-fringe-trees in  the polymers in $D_{\pi}$.
 
\item[-] $K$: the number of descriptors in a feature vector $f(\C)$.  
\end{enumerate}

\medskip \noindent
{\bf Stage~2.  }
We used  the  new  feature function  defined 
in our chemical model without suppressing hydrogen 
(see Appendix~\ref{sec:descriptor} for the detail).
We standardize the range of each descriptor and
 the range $\{t\in \R \mid \underline{a}\leq t\leq \overline{a}\}$ 
  of property values $a(\C), \C\in D_\pi$.

\medskip \noindent
{\bf Stage~3.  }
For each chemical property $\pi$, we select a penalty value $\lambda_\pi$
in the Lasso function from 36 different values from 0 to 100   
by conducting linear regression as a preliminary experiment.
 
We conducted an experiment in Stage~3 to evaluate the performance
of the prediction function based on cross-validation.
For a property $\pi$, 
an execution of a {\em cross-validation}  consists of five trials of
constructing a prediction function as follows.
First partition the data set $D_{\pi}$ 
 into five subsets $D_{\pi}^{(k)}$, $k\in[1,5]$ randomly;
 for each $k\in [1,5]$,  the $i$-th trial 
  constructs a prediction function $\eta(k)$  by conducting 
  a linear regression with the penalty term $\lambda_\pi$
using the set $D_{\pi}\setminus D_{\pi}^{(k)}$ as a training data set.
We used   scikit-learn version 0.23.2  with Python 3.8.5 for executing linear regression 
with Lasso function.
For each property, we executed ten cross-validations and
we show the median of test coefficient of determination $\mathrm{R}^2(\eta(k),D_{\pi}^{(k)}), k\in[1,5]$
over all ten cross-validations (see Appendix~\ref{sec:linear_regression} for the definition coefficient 
of determination $\mathrm{R}^2(\eta,D)$ for a prediction function $\eta$ over a data set $D$).
Recall that a subset of descriptors is selected in linear regression with Lasso function
and let $K'$ denote the average number of selected descriptors over all 50 trials.  
The running time per trial in a cross-validation was at most one second.

\begin{table}[h!]\caption{Results in Phase 1.} 
  \begin{center}
    \begin{tabular}{@{} c c r c  c  r r r c r  c   @{}}\hline
      $\pi$ & $\Lambda$  &  $|D_{\pi}|$  &  $ \underline{n},~\overline{n} $ &   $\underline{a},~\overline{a}$ &
              $|\Gamma|$   &  $|\mathcal{F}|$ &   $K$ &  
              $\lambda_\pi$ & $K'$~ &    test $\mathrm{R}^2$  \\ \hline
      {\sc AmD} & $\Lambda_1$  & 86  &  4,\,45  & 0.838,\,1.34 & 28  & 25    & 83 & $5.0\mathrm{E}{-4}$  & 17.7  & 0.914    \\ 
      {\sc AmD} &  $\Lambda_4$ &  93 &   4,\,45  & 0.838,\,1.45 & 31   &  30  & 94 & $6.0\mathrm{E}{-4}$ & 17.0 &  0.918    \\
      {\sc DeC} &  $\Lambda_4$  & 37 &  4,\,22  & 2.13,\,3.4  & 22  &19   & 72 & $4.0\mathrm{E}{-3}$ & 6.7 & 0.761   \\
      {\sc HcL} & $\Lambda_1$ &  52  & 4,\,25 & 105.7,\,677.8 &  22  & 17   & 67 &   $7.0\mathrm{E}{-4}$ & 14.2 & 0.990   \\
      {\sc HcL} & $\Lambda_5$  & 55  &  4,\,32 & 105.7,\,678.1 & 27  & 20  & 81 &  $2.0\mathrm{E}{-4}$& 28.3 & 0.987    \\
      {\sc HcS} & $\Lambda_1$ &  54  &  4,\,45  & 84.5,\,720.5 &  26  & 20 & 75 &  $5.0\mathrm{E}{-4}$ & 16.4 & 0.968   \\      
      {\sc HcS} & $\Lambda_5$  & 59   &4,\,45  & 84.5,\,720.5 & 32 & 24 & 92 & $5.0\mathrm{E}{-4}$ & 18.9 & 0.961    \\
      {\sc MlV} & $\Lambda_1$ & 86 & 4,\,45 &  60.7,\,466.6  & 28 & 25  & 83 & $2.0\mathrm{E}{-5}$ & 39.1 & 0.996   \\
      {\sc MlV} & $\Lambda_4$ & 93  &  4,\,45 &  60.7,\,466.6 &  31  & 30   & 94 & $2.0\mathrm{E}{-6}$ & 60.8 &  0.994     \\
      {\sc Prm} & $\Lambda_1$ &  112  &  4,\,45  & 2.23,\,4.91  & 25 &   15  & 69 &  $4.0\mathrm{E}{-5}$ & 23.7 & 0.801    \\
      {\sc Prm} &  $\Lambda_3$  & 131 &  4,\,45  & 2.23,\,4.91  & 25  &17   & 73 & $5.0\mathrm{E}{-5}$ & 27.3 & 0.784   \\
      {\sc RfId} &  $\Lambda_2$  & 91 &  4,\,29 &  1.4507,\,1.683 & 26  & 35  & 96 &  $9.0\mathrm{E}{-4}$ & 22.0 & 0.852  \\ 
      {\sc RfId} &  $\Lambda_6$  & 124 &  4,\,29 & 1.339,\,1.683  & 32  & 50  & 124 &   $9.0\mathrm{E}{-4}$  & 27.8 & 0.832  \\ 
      {\sc Tg} & $\Lambda_1$  & 204  &  4,\,58  &    171,\,673 & 32   & 36 & 101 & $9.0\mathrm{E}{-5}$  & 40.0 & 0.902  \\
      {\sc Tg} & $\Lambda_5$  & 232  & 4,\,58 &   171,\,673 & 36   & 43  & 118   & $9.0\mathrm{E}{-5}$  & 45.8 & 0.894    \\ 
      \hline
  \end{tabular}\end{center}\label{table:phase1a}
\end{table}

 Table~\ref{table:phase1a}   shows the results on Stages~2 and 3,
 where  we denote the following:     
\begin{enumerate}[nosep,  leftmargin=*]
\item[-] 
  $\lambda_\pi$: the penalty value in the Lasso function selected
for a property $\pi$, where $a\mathrm{E}{b}$ means $a\times 10 ^{b}$;

\item[-] 
 $K'$: the average of the number of descriptors selected in the linear regression
  over all 50 trials 
  in ten cross-validations;
\item[-]
test $\mathrm{R}^2$: the median of test coefficient of determination $\mathrm{R}^2$ over all 50 trials
  in ten cross-validations.  
\end{enumerate}
  
From  Table~\ref{table:phase1a}, we see that the number $K'$ of selected descriptors 
is around 15 to 50 over all properties $\pi$ and that  the number $K'$ becomes slightly larger
when the set $\Lambda$ of specified chemical elements is large for the same property $\pi$.

%% file: Result_2LMM_LLR_Phase2_polymer.tex
\begin{figure}[!htb]
\begin{center} 
 \includegraphics[width=.89\columnwidth]{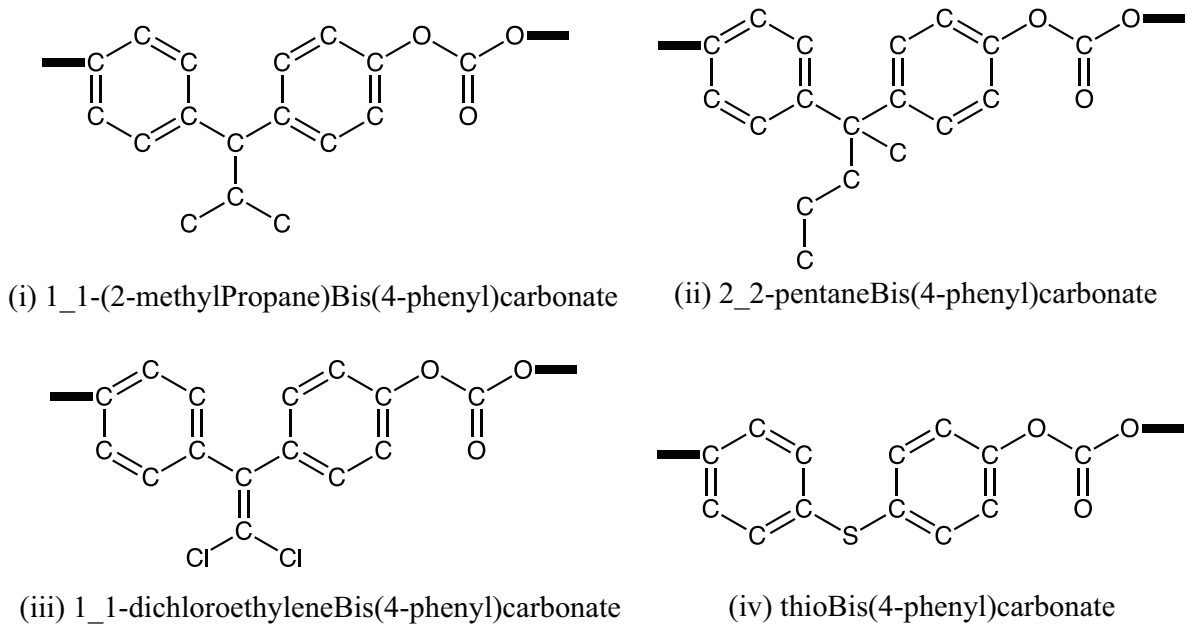}
\end{center}
\caption{An illustration of  four polymers: 
(i)  1$\underline{~}$1-(2-methylPropane)Bis(4-phenyl)carbonate;
(ii) 2$\underline{~}$2-pentaneBis(4-phenyl)carbonate;
(iii)  1$\underline{~}$1-dichloroethyleneBis(4-phenyl)carbonate;
(iv)  thioBis(4-phenyl)carbonate,
where hydrogens are omitted and connecting edges are depicted with thick lines. }
\label{fig:four_polymers}  
\end{figure}
\begin{figure}[h!] \begin{center}
\includegraphics[width=.95\columnwidth]{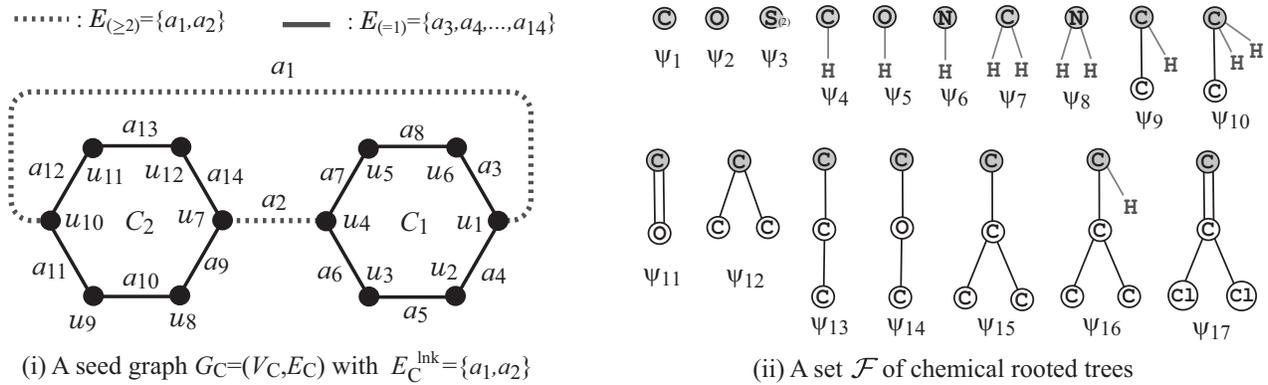}
\end{center} \caption{
(i)  A seed graph   $\GC$ for $I_{\mathrm{b}}$;
(ii) A set $\mathcal{F}$ of chemical rooted trees. }
\label{fig:test_set_fringe-trees_polymers}
\end{figure}

\medskip \noindent
{\bf Results on Phase~2.  }
To execute  Stages~4  and 5 in Phase~2, 
we used a set of two instances
$I_{\mathrm{a}}$ and $I_{\mathrm{b}}$.
We here present their seed graphs $\GC$ 
(see Appendices~\ref{sec:specification} and \ref{sec:test_instances} 
for the details of them).
The seed graph  $\GC$ of instance $I_{\mathrm{a}}$ is given
 by the graph in Figure~\ref{fig:specification_example_1}(a).
 Instance  $I_{\mathrm{b}}$ is introduced to 
  represent a set of polymers that includes the four examples of polymers in 
Figure~\ref{fig:four_polymers}.
The seed graph  of instance $I_{\mathrm{b}}$ is illustrated in 
 Figure~\ref{fig:test_set_fringe-trees_polymers}(a).

\medskip \noindent
{\bf Stage~4.  } 
We executed Stage~4 for  four
properties $\pi\in \{${\sc  AmD, HcL,  RfId, Tg}$\}$.  
 For the MILP formulation  $\mathcal{M}(x,y;\mathcal{C}_1)$ 
 in Section~\ref{sec:linear_regression}, 
we use the prediction function $\eta_{\w,b}$  
 that attained the median  test $\mathrm{R}^2$ in Table~\ref{table:phase1a}.
 To solve an MILP   in Stage~4, we used 
{\tt  CPLEX} version 12.10.

For property   {\sc  Prm}, we also need to specify
   the frequency $\mathrm{fq}$ under which the value $a(\C)$ is observed,    and 
  set lower and upper bounds 
 $\underline{\mathrm{fq}}, \overline{\mathrm{fq}} \in \R$ 
  on the frequency 
  to be  $\underline{\mathrm{fq}}:=60$ and  $\overline{\mathrm{fq}}:= 1.0\times 10^7$
  in this experiment. 
  
Tables~\ref{table:stages_4_5}  shows
  the computational results of the experiment in Stage~4 
for the four properties {\sc AmD, HcL,  Prm, RfId} and {\sc Tg}, respectively, 
 where we denote the following:
\begin{enumerate} [nosep,  leftmargin=*]
\item[-]  
$\pi$: a property $\pi\in \{${\sc  AmD, HcL,  RfId, Tg}$\}$; 

\item[-]  
inst.: instance $I_{\mathrm{a}}$ or $I_{\mathrm{b}}$; 

\item[-]  
$n_\LB$: a lower bound on the number of non-hydrogen atoms;

\item[-]  
  $ \underline{y}^*,~\overline{y}^* $:  
 lower and upper bounds $\underline{y}^*, \overline{y}^*\in \R$ 
  on the value $a(\C)$ of a polymer $\C$ to be inferred; 
  
\item[-]  
 $\#$v (resp.,  $\#$c): 
 the number  of variables (resp., constraints)  in the MILP  in Stage~4;  
 
\item[-]   
 I-time: the time (sec.) to solve the MILP  in Stage~4;  

\item[-]  
    $n$:  the number  $n(\C^\dagger)$  of  non-hydrogen atoms
     in the monomer representation $\C^\dagger$   inferred in Stage~4,
     where ``none'' means that no desired polymer exists for the topological specification;
     
\item[-]  
 $\nint$:  the number  $\nint(\C^\dagger)$ of interior-vertices in
  the monomer representation $\C^\dagger$   inferred in Stage~4;     and
  
\item[-]  
$\eta $: the predicted property value 
$\eta(f(\C^\dagger))$ of the polymer $\C^\dagger$ inferred 
in Stage~4.
\end{enumerate}

\begin{table}[h!]\caption{Results of Stages~4 and 5.}  
 \begin{center}
 \begin{tabular}{@{} c  c  c c  r  r c r   r   r c r     @{}}\hline                
$\pi$ & inst. & $n_\LB$ &  $ \underline{y}^*,~\overline{y}^* $ &  $\#$v~   &  $\#$c~~  &  
   {\small I-time} & ~$n$~~$\nint$  & {\small $\eta $}~~~  & 
                 {\small  D-time} &  {\small $\C$-LB } &  {\small $\#\C$}    \\ \hline
 {\sc  AmD} &  $I_{\mathrm{a}}$ & 30   &  0.885,\,0.890  &  11247 & 12964     & 6.20   &  49~30     &  0.889 &  0.285 & 64 &   64    \\%
                   &   $I_{\mathrm{b}}$ &  25    &  1.344,\,1.350  & 7125   &  7690  & 2.54  & 28~22  & 1.347  & 0.188 & 2610 & 100   \\ \hline 
 {\sc  HcL} &  $I_{\mathrm{a}}$ & 30 & 105.7,\,678.1   &  12171 & 13017  & 31.0  & none~     & - & - & -   & -      \\%
                  &  $I_{\mathrm{b}}$ & 30  &  658.8,\,660.2  &  8469  &  9916   & 1.51   &  32~20    & 660.0    & 0.189 & 576 &  100 \\ \hline 
 {\sc  Prm} &  $I_{\mathrm{a}}$ & 30    &  4.128\,4.150  &  9878 & 12547  &  10.7  &  50~30 &  4.150    &   0.166 & 24 & 24 \\%
                  &  $I_{\mathrm{b}}$ & 35   &  3.158\,3.188    &  8999 & 12112      &  2.03  &  41~24    & 3.188   & 0.190  &$1.5\mathrm{E}4$ & 100  \\   \hline 
 {\sc  RfId} &  $I_{\mathrm{a}}$ & 30 &  1.339,\,1.683    & 9979 & 12661  &  92.1  & none~    & - & - & -   & -         \\%
                 &   $I_{\mathrm{b}}$ & 40  & 1.406,\,1.422  & 10460 & 15035   &  2.61  &  47~27   & 1.413   &   0.202  & $7.8\mathrm{E}5$ &  100  \\ \hline 
 {\sc  Tg}   &  $I_{\mathrm{a}}$ & 30   &  180.0,\,181.6    &  12245 & 13102    & 17.0   &  50~30   &  181.06   & 0.220 & 36 & 36\\%
                 &  $I_{\mathrm{b}}$ &  45 &   180.6,\,182.8 &  12953& 18549 & 32.8   &  55~28    &  182.20 &  0.196 & $6.3\mathrm{E}5$ &   100 \\  \hline  
   \end{tabular}\end{center}\label{table:stages_4_5}
\end{table}
 
In Table~\ref{table:stages_4_5},  $\eta(f(\C^\dagger))$ is the predicted value of property $\pi$
of a polymer $\C^\dagger$ constructed by solving an MILP in Stage~4,
where we see that each $\eta(f(\C^\dagger))$ actually satisfies
the specified lower and upper bounds on a target chemical value.

We set lower and upper bounds on a target chemical value for property
   {\sc  HcL} with $\Lambda_1$  
 so that $(\underline{y}^*, \overline{y}^*)$ is the maximal range of
 the observed values over the data set $D_\pi$;
 i.e.,  $(\underline{y}^*, \overline{y}^*):=(\underline{a}, \overline{a})=(105.7,678.1)$. 
 Similarly for property    {\sc  RfId}  with  $\Lambda_6$,
 we set   $(\underline{y}^*, \overline{y}^*):= (\underline{a}, \overline{a})=(1.339,1.683 )$. 
 For an example of    $I_{\mathrm{a}}$ with  {\sc  AmD}, it holds that 
$\underline{y}^* \leq \eta(f(\C^\dagger)) \leq \overline{y}^* $  
with 
$\underline{y}^*=0.885$, $\overline{y}^* = 0.890$ and $\eta(f(\C^\dagger))= 0.889$.  
 For  instance  $I_{\mathrm{a}}$ with {\sc  HcL} and {\sc  RfId}, 
 Table~\ref{table:stages_4_5} reveals that there is no chemical graph that satisfies 
 the topological specification $I_{\mathrm{a}}$.  
These infeasible instance and instance $I_\mathrm{b}$ with $\pi=${\sc Tg}  took around 
30 to 90 seconds.
For the other cases, solving an MILP for inferring a polymer with around 50 non-hydrogen atoms
in the monomer form is around 2 to 15 seconds.

\begin{figure}[!htb]
\begin{center} 
 \includegraphics[width=.98\columnwidth]{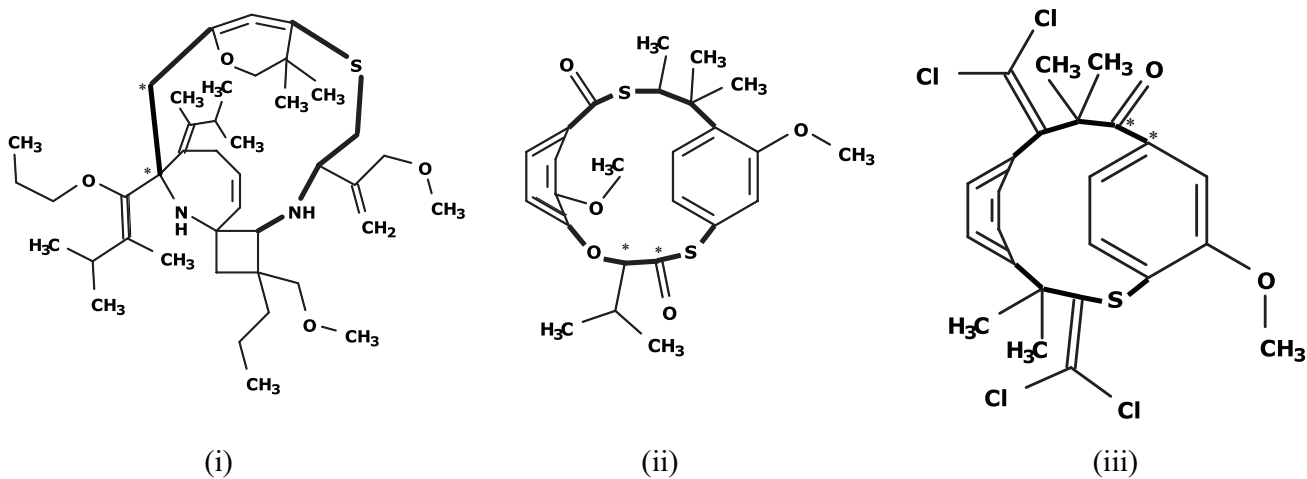}
\end{center}
\caption{ 
Illustrations of polymers, 
where the link-edges are depicted with thick lines 
 and the connecting-vertices are marked with asterisks.  
(i) A polymer  $\C^\dagger$ with $\eta(f(\C^\dagger))=0.889$ inferred
 from   $I_{\mathrm{a}}$ with $(\underline{y}^*, \overline{y}^*) =(0.885,0.890)$  of  {\sc AmD};   
(ii)  A polymer  $\C^\dagger$ with $\eta(f(\C^\dagger))= 660.0$ inferred
 from   $I_{\mathrm{b}}$ with $(\underline{y}^*, \overline{y}^*) =(658.8,660.2)$  of  {\sc HcL}.   
(iii)  A polymer  $\C^\dagger$ inferred 
from   $I_{\mathrm{b}}$ with lower and upper bounds on 
the predicted property value $\eta_\pi(f(\C^\dagger))$ of   property 
  $\pi\in\{${\sc AmD, HcL, Tg}$\}$ in Table~\ref{table:stage_4_three_properties}. 
}
\label{fig:MILP_solutions_polymer}  
\end{figure}

\bigskip 
Figure~\ref{fig:MILP_solutions_polymer}(i)  (resp., (ii))
 illustrates   the chemical graph  $\C^\dagger$  inferred   from   $I_{\mathrm{a}}$  (resp.,   $I_{\mathrm{b}}$)
with $(\underline{y}^*, \overline{y}^*) =(0.885,0.890)$    of  {\sc AmD}
   (resp., $(\underline{y}^*, \overline{y}^*) =(658.8,660.2)$  of  {\sc HcL})
  in Table~\ref{table:stages_4_5}.

 From Table~\ref{table:stages_4_5}, we observe that  instances
 with around  30 to 55  non-hydrogen atoms
 in the monomer representation are solved
 in around 2 to 30 seconds when they are feasible.

\input additional_experiment_Stage_4.tex

\medskip \noindent
{\bf Stage~5.  } 
We executed Stage~5 to generate a more number of target chemical graphs $\C^*$,
where we call a chemical graph $\C^*$ a {\em chemical isomer} of
a target chemical graph $\C^\dagger$ of a topological specification $\sigma$
if $f(\C^*)=f(\C^\dagger)$ and $\C^*$ also satisfies the same topological specification $\sigma$.
For this, we executed the same algorithm used by Zhu~et~al.~\cite{ZAHZNA21}. 
We computed  chemical isomers  $\C^*$ of 
each target chemical graph  $\C^\dagger$ inferred in Stage~4.
We execute an  algorithm for generating chemical isomers of   $\C^\dagger$
up to 100 when the number of all chemical isomers exceeds 100.
The algorithm can evaluate a lower bound 
on the total number of all chemical isomers $\C^\dagger$
without generating all of them.

Tables~\ref{table:stages_4_5}  shows 
   the computational results of the experiment
in Stage~5 for properties {\sc AmD, HcL,  RfId} and {\sc  Tg}, respectively,  
 where we denote the following:
\begin{enumerate}[nosep,  leftmargin=*]
\item[-]
 D-time: the running time (sec.) to execute the dynamic programming algorithm
 in Stage~5 to compute a lower bound on the number 
 of all polymers  $\C^*$ of  $\C^\dagger$   
 and generate all (or up to 100) chemical isomers $\C^*$;
 
\item[-]
 $\C$-LB: a lower bound on the number of all chemical isomers $\C^*$ of
$\C^\dagger$, where $a\mathrm{E}b$ means $a\times 10^{b}$;~and

\item[-]
 $\#\C$: the number of all (or up to 100) chemical isomers $\C^*$ of  $\C^\dagger$  
 generated in Stage~5.
\end{enumerate} 
 
 From Table~\ref{table:stages_4_5}, we observe that 
 the number of isomers $\C^*$ of  an output polymer $\C^\dagger$ 
 varies on each case, where
 the polymer  $\C^\dagger$ admits 
  only 24 isomers $\C^*$ for instance $I_\mathrm{a}$ and $\pi=${\sc Prm}
  and over $6.3\times 10^5$ for  instance $I_\mathrm{b}$ and $\pi=${\sc Tg}.
  The computation time for generating at most 100 isomers $\C^*$ and
  estimating a lower bound  $\C$-LB is at most 0.3 second for all cases in our experiment.

%% file: additional_experiment_Stage_4.tex
\bigskip
\noindent 
{\bf Inferring a polymer with target values in multiple properties}\\
Once we obtained prediction functions $\eta_\pi$ for several properties $\pi$,
it is easy to include MILP formulations for these functions $\eta_\pi$
into a single MILP $\mathcal{M}(x,y;\mathcal{C}_1)$ so as to
infer a chemical graph that satisfies given target values $y^*$ for
these properties at the same time.
As an additional experiment in Stage~4, 
we conducted a computational experiment for inferring a polymer
  that has a desired predicted value
each of some three properties $\pi_1,\pi_2$ and $\pi_3$. 
For a combination of three properties, we selected two sets 
$\mathrm{P}_1=\{${\sc AmD, HcL, Tg}$\}$  and 
$\mathrm{P}_2=\{${\sc HcS, MlV, RfId}$\}$, 
where we used the prediction function
$\eta_\pi$ for each property $\pi\in\mathrm{P}_i$ constructed in Stage~3.
Table~\ref{table:stage_4_three_properties} shows
 the result of Stage~4 
for inferring a chemical graph $\C^\dagger$
 from instance  $I_{\mathrm{b}}$ 
with  a set 
$\Lambda(\mathrm{P}_i)$ of chemical elements
for the set $\mathrm{P}_i$ of properties
such that 
$\Lambda(\mathrm{P}_1)
      =\Lambda_3=\{ \ttH, \ttC, \ttO, \ttN,  \ttCl, \ttS_{(2)} \}$ and 
$\Lambda(\mathrm{P}_2)
      =\{\ttH, \ttC, \ttO_{(2)}, \ttN,  \ttCl \}$, 
 where we denote the following: 
\begin{enumerate} [nosep,  leftmargin=*]
\item[-]  
  $\mathrm{P}_i, i=1,2$: a combination of three properties, 
  where $\mathrm{P}_1=\{${\sc AmD, HcL, Tg}$\}$  and 
$\mathrm{P}_2=\{${\sc HcS, MlV, RfId}$\}$;  
  
\item[-]  
$\pi$: one of the three properties in $\mathrm{P}_i, i=1,2$
 used in the experiment;

\item[-]  
  $ \underline{y}^*_\pi,~\overline{y}^*_\pi$:  
 lower and upper bounds $\underline{y}^*_\pi, \overline{y}^*_\pi\in \R$ 
  on  the predicted property value $\eta_\pi(f(\C^\dagger))$ 
  of  property  $\pi\in\mathrm{P}_i, i=1,2$
   for  a polymer $\C^\dagger$ to be inferred;
  
\item[-]  
 $\#$v (resp.,  $\#$c): 
 the number  of variables (resp., constraints)  in the MILP  in Stage~4;  
 
\item[-]   
 I-time: the  time (sec.) to solve the MILP  in Stage~4;  
 
\item[-]  
    $n$:  the number  $n(\C^\dagger)$  of  non-hydrogen atoms
     in the monomer representation $\C^\dagger$   inferred in Stage~4;   and 
\item[-]  
 $\nint$:  the number  $\nint(\C^\dagger)$ of interior-vertices in
  the monomer representation $\C^\dagger$   inferred in Stage~4; 
      
\item[-]  
$\eta_\pi$: the predicted property value 
$\eta_\pi(f(\C^\dagger))$ 
  of  property  $\pi\in\mathrm{P}_i, i=1,2$
   for the polymer $\C^\dagger$ inferred in Stage~4.
\end{enumerate} 

 \begin{table}[h!]\caption{ Results of Stage~4 for instance $I_{\mathrm{b}}$ 
 with specified target values of the three properties in $\mathrm{P}_i, i=1,2$.} 
 \begin{center}
 \begin{tabular}{@{} c  c  c c  c  r r r r r c    @{}}\hline                
$\mathrm{P}_i$  & $n_\LB$ & $\pi$  & $ \underline{y}_\pi^*,~\overline{y}_\pi^* $ & $\#$v  &  $\#$c~   &  
   {\small I-time}  & $n$~  &   $\nint$  &   $\eta_\pi$~~    \\ \hline
                       & & {\sc AmD}  &  1.200,\,1.224   &   &   &   &   &                            &   1.217    \\%
$\mathrm{P}_1$  & 25 & {\sc HcL}  & 624.0,\,628.0   &  7525 & 8211  & 3.09 & 31  &  18  &   625.9    \\%
                       && {\sc Tg}   & 171.0,\,174.0    &   &   &   &   &                            &   171.55  \\%
 \hline 
                       &&   {\sc HcS}  &   539,\,541  &   &   &   &   &                            &    540.7   \\%
$\mathrm{P}_2$  & 45 &   {\sc MlV}  & 393,\,395    & 12162  &  18536 & 210.2 & 45  & 29 &   394.3  \\%
                       &&  {\sc RfId}   & 1.4507,\,1.479    &   &   &   &   &                            & 1.46 \\%
   \hline
   \end{tabular}\end{center}\label{table:stage_4_three_properties}
\end{table}

Fig.~\ref{fig:MILP_solutions_polymer}(iii)
 illustrates  the polymer  $\C^\dagger$  inferred 
from  $I_{\mathrm{b}}$  with 
$(\underline{y}^*_{\pi_1}, \overline{y}^*_{\pi_1}) =(1.200, 1.224)$, 
$(\underline{y}^*_{\pi_2}, \overline{y}^*_{\pi_2}) =(624.0, 628.0)$ and 
$(\underline{y}^*_{\pi_3}, \overline{y}^*_{\pi_3}) =(171.0, 174.0)$
for $\pi_1=${\sc AmD}, $\pi_2=${\sc HcL} and  $\pi_3=${\sc Tg},
 respectively.

%% file: Conclusion_2LMM_LLR_polymer.tex
\section{Concluding Remarks}\label{sec:conclude}
  
  In this paper, we designed a method for inferring  polymers 
  based on the framework for monomers proposed by 
Akutsu and Nagamochi~\cite{AN19}.
To treat a polymer as a form of monomers with no connecting-edges,
we introduce a new way of representing a polymer with a monomer form
by distinguishing link-edges from other edges in polymers.  
Since the link-edges of a polymer are characteristic to the polymer,
we included new descriptors that feature the link-edges of a polymer
into our feature vector. 
We constructed prediction functions by linear regression 
for eight chemical properties on polymers in Phase~1 of the framework.
We inferred polymers for the first time in Phase~2 of the framework.
The results of our computational experiments suggest that
the method still can infer a polymer with 50 non-hydrogen atoms
in the monomer form in a reasonable running time.  
 
 There are some chemical properties on polymers to which linear regression 
 did not provide a good prediction function.
 It is left as a future work to use other learning methods such as
 decision trees and neural networks
 and find new effective descriptors in order
  to  construct a prediction function with a better performance 
  for these  chemical properties on polymers.

%% file: Linear_Regression.tex
\section{Linear Regressions}\label{sec:linear_regression} 

This section reviews the method for linear regression used
by Zhu~et~al.~\cite{ZAHZNA21} in the framework of inferring chemical graphs.

For an integer $p\geq 1$ and a vector $\x\in \R^p$, the $j$-th entry of $\x$ 
is denoted by $\x(j), j\in [1,p]$.

Let $D$ be a data set   of chemical graphs $\C$ with
an observed value $a(\C)\in \R$,
where we denote by $a_i=a(\C_i)$ 
for an indexed graph $\C_i$. 

Let $ f$ be a feature function that maps a chemical graph $\C$
to a vector $ f(\C)\in \RK$
where we denote by $\x_i= f(\C_i)$ 
for an indexed graph $\C_i$. 
For  a prediction function $\eta: \RK\to \R$, 
define an error function 
\[ \mathrm{Err}(\eta;D)  \triangleq 
\sum_{\C_i\in D}(a_i - \eta(f(\C_i)))^2=\sum_{\C_i\in D}(a_i - \eta(\x_i))^2, \]
and define the {\em coefficient of determination}
 $\mathrm{R}^2(\eta,D)$ 
  to be 
\[ \displaystyle{ \mathrm{R}^2(\eta,D)\triangleq 
  1- \frac{\mathrm{Err}(\eta;D) } 
  {\sum_{ \C_i\in D  } (a_i-\widetilde{a})^2}  
  \mbox{   for  }
   \widetilde{a}= \frac{1}{|D |}\sum_{ \C\in D }a(\C).  } \]

For a feature space $\RK$, a hyperplane is defined to be 
a pair  $(\w,b)$ of a vector $\w\in \RK$ and a real $b\in \R$.
Given a hyperplane $(\w,b)\in \RKw$,
a prediction function $\eta_{\w,b}:\RK\to \R$ is defined by setting
\[ \eta_{\w,b}(\x) \triangleq \w\cdot \x +b=\sum_{j\in [1,K]}\w(j)\x(j) +b. \]
We  observe that such a prediction function can be represented as an ANN
with an input layer with $K$ nodes $u_{j}, j\in [1,K]$
and an output layer with a single node $v$
such that the weight of edge arc $(u_{j},v)$ is set to be $w(j)$, 
 the bias of node $u$ is set to be $b$ 
 and the activation function at node $u$ is set to be a linear function.
 However, a learning algorithm for an ANN may not find a set of weights
 $w(j),j\in[1,K]$ and $b$ that minimizes the error function, since
 the algorithm simply iterates modification of the current weights and biases
 until it terminates at a local optima in the minimization. 

We wish to find a hyperplane $(\w,b)$  that minimizes the error function
$\mathrm{Err}(\eta_{\w,b};D)$.
In many cases, a feature vector $f$ contains descriptors that do not play
an essential role in constructing a good prediction function.
When we solve the minimization problem, the entries $\w(j)$ for some descriptors $j\in [1,K]$
in the resulting hyperplane $(\w,b)$ become zero, which means that these descriptors were
not necessarily important for finding a prediction function $\eta_{\w,b}$. 
It is proposed that solving the minimization with an additional penalty term $\tau$ to the error function 
often results in a more number of  entries $\w(j)=0$, reducing a set of descriptors
necessary for defining   a prediction function $\eta_{\w,b}$. 
For an error function with such a penalty term,  
a Ridge function 
 $\frac{1}{2|D|}\mathrm{Err}(\eta_{\w,b};D)
 +\lambda [\sum_{j\in[1,K]}w(j)^2 +  b^2]$~\cite{Ridge88}
 and  a Lasso function 
$\frac{1}{2|D|}\mathrm{Err}(\eta_{\w,b};D)
+ \lambda [\sum_{j\in[1,K]}  |w(j)| +  |b| ]$~\cite{Lasso96} 
are known, where $\lambda \in \R$ is a given real number.

Given a prediction function $\eta_{\w,b}$,
we can simulate a  process of computing 
the output $\eta_{\w,b}(\x)$ for an input $\x\in \RK$
as an MILP   $\mathcal{M}(x,y;\mathcal{C}_1)$ in the framework. 
By solving such an MILP for a specified target value $y^*$,   
we can find a vector $x^*\in \RK$ such that $\eta_{\w,b}(\x^*)=y^*$.
Instead of specifying a single target value $y^*$, we use
  lower and upper bounds $\underline{y}^*, \overline{y}^*\in \R$ 
  on the value $a(\C)$ of a chemical graph $\C$ to be inferred.
We can control the range between $\underline{y}^*$ and $\overline{y}^*$
for searching a chemical graph $\C$  
by setting  $\underline{y}^*$ and $\overline{y}^*$ to be close or different values.  
A desired MILP is formulated as follows. 

\paragraph{$\mathcal{M}(x,y;\mathcal{C}_1)$: 
 An MILP formulation  for the inverse problem to prediction function }
~\\

\smallskip\noindent
{\bf constants: } 
\begin{enumerate}  [nosep,  leftmargin=*]
\item[-]  
A hyperplane $(\w,b)$ with $\w\in \RK$ and $b\in \R$;

  
\item[-]
Real values $\underline{y}^*, \overline{y}^*\in \R$ such that $\underline{y}^*< \overline{y}^*$; 

\item[-] 
A set $I_{\Z}$ of indices $j\in [1,K]$ such that 
the $j$-th descriptor $\dcp_j(\C)$ is always an integer;

\item[-] 
A set $I_{+}$ of indices $j\in [1,K]$ such that 
the $j$-th descriptor $\dcp_j(\C)$ is always non-negative;

\item[-] 
$\ell(j), u(j)\in \R, j\in [1,K]$: lower and upper bounds
 on the $j$th-descriptor;
 \end{enumerate}

\smallskip\noindent
{\bf variables: } 
\begin{enumerate} [nosep,  leftmargin=*]

\item[-] 
Non-negative integer variable $x(j)\in \Z_+, j\in I_{\Z}\cap I_+$;

\item[-] 
Integer variable $x(j)\in \Z, j\in I_{\Z}\setminus I_+$;

\item[-] 
Non-negative real variable $x(j)\in \Z_+, j\in  I_+\setminus I_{\Z}$;

\item[-] 
Real variable $x(j)\in \Z, j\in [1,K] \setminus (I_{\Z}\cup I_+)$;
 \end{enumerate}
 
\smallskip\noindent
{\bf constraints: }   
\begin{align} 
 \ell(j) \leq x(j)  \leq  u(j), j\in [1,K], \label{eq:1} 
\end{align}   
\begin{align} 
\underline{y}^* \leq 
\sum_{j\in [1,K]} w(j)x(j)+b \leq  \overline{y}^*,  \label{eq:2} 
\end{align}   
 
\smallskip\noindent
{\bf objective function: }   \\
~~ none.\\

The number of variables and constraints in the above MILP formulation is $O(K)$. 
It is not difficult to see that the above MILP is an NP-hard problem. 

The entire MILP for Stage~4 consists of the two MILPs
$\mathcal{M}(x,y;\mathcal{C}_1)$ and $\mathcal{M}(g,x;\mathcal{C}_2)$ 
with no objective function. 
The latter  represents the computation process of our feature function $f$ and 
a given topological specification.   
See Appendix~\ref{sec:full_milp}  for the details of  MILP $\mathcal{M}(g,x;\mathcal{C}_2)$.

%% file: Descriptor_M_lfac_polymer.tex
\section{A Full Description of Descriptors}\label{sec:descriptor}

Our definition of feature function is analogous with the one  by 
 Zhu~et~al.~\cite{ZAHZNA21}
   except for a necessary modification due to our polymer model 
   with link-edges.

Associated with the two functions 
$\alpha$ and $\beta$ in a chemical graph $\Co=(H,\alpha,\beta)$,
we introduce   functions  
 $\ac: V(E)\to (\Lambda\setminus\{\ttH\})\times (\Lambda\setminus\{\ttH\})\times [1,3]$, 
 $\cs: V(E)\to (\Lambda\setminus\{\ttH\})\times [1,6]$ and
$\ec: V(E)\to ((\Lambda\setminus\{\ttH\})\times [1,6])
\times ((\Lambda\setminus\{\ttH\})\times [1,6])\times [1,3]$
in the following. 

 To represent  a feature of the exterior  of  $\Co$, 
  a  chemical rooted tree in $\mathcal{T}(\Co)$ is
  called a {\em fringe-configuration} of $\Co$. 

We also represent leaf-edges in the exterior of $\Co$.
For a leaf-edge $uv\in E(\anC)$ with $\deg_{\anC}(u)=1$, we define
the {\em adjacency-configuration} of $e$ to be an ordered tuple
$(\alpha(u),\alpha(v),\beta(uv))$. 
Define 
\[ \Gac^\lf\triangleq \{(\ta,\tb,m)\mid \ta,\tb\in\Lambda, 
m\in[1,\min\{\val(\ta),\val(\tb)\}]\} \]
as a set of possible adjacency-configurations for leaf-edges. 

To  represent a feature of an interior-vertex $v\in V^\inte(\Co)$ such that
$\alpha(v)=\ta$  and  $\deg_{\anC}(v)=d$
(i.e., the number of non-hydrogen atoms adjacent to $v$ is $d$) 
   in a chemical   graph  $\Co=(H,\alpha,\beta)$,
 we use  a pair $(\ta, d)\in (\Lambda\setminus\{{\tt H}\})\times [1,4]$,
 which we call the {\em chemical symbol} $\cs(v)$ of the vertex $v$.
 We treat $(\ta, d)$ as a single symbol $\ta d$,  and  
define $\Ldg$   to be  the set of all chemical symbols
$\mu=\ta d\in  (\Lambda\setminus\{{\tt H}\})\times [1,4]$.  

We define a method for featuring interior-edges  as follows.
Let $e=uv\in E^\inte(\Co)$  be 
 an interior-edge $e=uv\in E^\inte(\Co)$ 
 such that $\alpha(u)=\ta$, $\alpha(v)=\tb$ and $\beta(e)=m$ 
   in a chemical graph  $\Co=(H,\alpha,\beta)$.
To feature this edge $e$, 
 we use a tuple $(\ta,\tb,m)\in (\Lambda\setminus\{{\tt H}\})
    \times (\Lambda\setminus\{{\tt H}\})\times [1,3]$,
 which we call the {\em adjacency-configuration} $\ac(e)$ of the edge $e$. 
 We introduce a total order $<$ over the elements in $\Lambda$
 to distinguish  between $(\ta,\tb, m)$ and $(\tb,\ta, m)$ 
 $(\ta\neq \tb)$ notationally.
 For a tuple  $\nu=(\ta,\tb, m)$,
 let $\overline{\nu}$ denote the tuple $(\tb,\ta, m)$.  

Let $e=uv\in E^\inte(\Co)$  be 
an  interior-edge $e=uv\in E^\inte(\Co)$ 
 such that $\cs(u)=\mu$, $\cs(v)=\mu'$ and $\beta(e)=m$ 
   in a chemical  graph  $\Co=(H,\alpha,\beta)$.
To feature this edge $e$, 
 we use a tuple $(\mu,\mu',m)\in \Ldg\times \Ldg\times [1,3]$, 
 which we call  the {\em edge-configuration} $\ec(e)$ of the edge $e$. 
 We introduce a total order $<$ over the elements in $\Ldg$
 to distinguish between $(\mu,\mu', m)$ and $(\mu', \mu, m)$ 
 $(\mu \neq \mu')$ notationally. 
 For a tuple  $\gamma=(\mu,\mu', m)$,
 let $\overline{\gamma}$ denote the tuple $(\mu', \mu, m)$. 
   
Let $\pi$ be a chemical property for which we will construct
a prediction function $\eta$ from a feature
vector $f(\C)$ of a chemical graph $\C$ 
to a predicted value $y\in \mathbb{R}$
for the  chemical property of $\C$.

We first choose a set $\Lambda$ of chemical elements
 and then collect a data set  $D_{\pi}$ of
  chemical compounds  $C$ whose 
  chemical elements belong to $\Lambda$,
  where we regard  $D_{\pi}$ as a set of chemical graphs $\C$
  that represent the chemical compounds $C$  in  $D_{\pi}$.
To define the interior/exterior of 
chemical graphs  $\C\in D_{\pi}$,
we  next choose a branch-parameter ${\rho}$, where
 we recommend ${\rho}=2$.  
 
Let $\Lambda^\inte(D_\pi)\subseteq \Lambda$  
(resp., 
$\Lambda^\ex(D_\pi)\subseteq \Lambda$)
denote the set  of chemical elements  used in
the set $V^\inte(\C)$ of interior-vertices
(resp., the set $V^\ex(\C)$ of  exterior-vertices) of $\C$
 over all chemical graphs $\C\in D_\pi$, 
and $\Gamma^\inte(D_\pi)$
(resp., $\Gamma^\lnk(D_\pi)$) 
denote the set of edge-configurations used in
the set $E^\inte(\C)$  of interior-edges
(resp., the set $\Elnk(\C)$ of linked-edges) in $\C$
 over all chemical graphs $\C\in D_\pi$. 
Let $\mathcal{F}(D_\pi)$ denote the set of
chemical rooted trees $\psi$  
r-isomorphic to a chemical rooted tree in $\mathcal{T}(\C)$
  over all chemical graphs $\C\in D_\pi$,
  where possibly a chemical rooted tree $\psi\in \mathcal{F}(D_\pi)$
  consists of a single chemical element $\ta\in \Lambda\setminus \{{\tt H}\}$.
  
We define an integer encoding of a finite set $A$ of elements
to be a bijection $\sigma: A \to [1, |A|]$, 
where we denote by $[A]$   the set $[1, |A|]$ of integers.
Introduce  an integer coding of each of the   sets 
$\Lambda^\inte(D_\pi)$, $\Lambda^\ex(D_\pi)$, 
$\Gamma^\inte(D_\pi)$ and $\mathcal{F}(D_\pi)$. 
Let $[\ta]^\inte$  
(resp., $[\ta]^\ex$)  denote   
the coded integer of  an element $\ta\in \Lambda^\inte(D_\pi)$
(resp., $\ta\in \Lambda^\ex(D_\pi)$),  
$[\gamma]$   denote  
the coded integer of  an element $\gamma$ in $\Gamma^\inte(D_\pi)$
and 
$[\psi]$   denote  an element $\psi$ in $\mathcal{F}(D_\pi)$. 
 
We assume that a chemical graph $\C$
 treated in this paper satisfies  $\deg_{\anC}(v)\leq 4$
in the hydrogen-suppressed graph $\anC$.
 
In our model, we  use an integer 
  $\mathrm{mass}^*(\ta)=\lfloor 10\cdot \mathrm{mass}(\ta)\rfloor$, 
 for each $\ta\in \Lambda$.
 
  We define the {\em feature vector} $f(\C)$ 
  of a polymer $\C=(H,\alpha,\beta)\in D_{\pi}$ 
  to be a vector that consists of the following  
non-negative integer descriptors $\dcp_i(\C)$, $i\in [1,K]$, where 
$K=14+ |\Lambda^\inte(D_\pi)|+|\Lambda^\ex(D_\pi)|
+|\Gamma^\inte(D_\pi)|+|\Gamma^\lnk(D_\pi)|+|\Ldg|
+|\mathcal{F}(D_\pi)|+|\Gac^\lf|$. 


\begin{enumerate}  
\item   
$\dcp_1(\C)$: the number  $|V(H)|-|\VH|$ of non-hydrogen atoms  in  $\C$.  
 
\item 
$\dcp_2(\C)$:  the number $|V^\inte(\C)|$ of interior-vertices in  $\C$.
  
\item 
$\dcp_3(\C)$:  the number $|\Elnk(\C)|$ of link-edges in  $\C$.
This descriptor is newly introduced in this paper to feature
a structure of polymers. 

\item 
$\dcp_4 (\C)$: 
the average $\overline{\mathrm{ms}}(\C)$ of mass$^*$ 
over all atoms in $\C$; \\
 i.e., $\overline{\mathrm{ms}}(\C)\triangleq 
 \frac{1}{|V(H)|}\sum_{v\in V(H)}\mathrm{mass}^*(\alpha(v))$. 

\item 
$\dcp_i(\C)$,  $i=4+d,   d\in [1,4]$: 
the number $\dg_d^{\oH} (\C)$  of non-hydrogen vertices $v\in V(H)\setminus \VH$
 of degree $\deg_{\anC}(v)=d$
 in the hydrogen-suppressed chemical graph $\anC$.  
 
\item 
$\dcp_i(\C)$,  $i=8+d,   d\in [1,4]$: 
the number $\dg_d^\inte(\C)$
 of interior-vertices of interior-degree  $\deg_{\C^\inte}(v)=d$
  in the interior $\C^\inte=(V^\inte(\C),E^\inte(\C))$ of  $\C$. 
  
   
\item $\dcp_i(\C)$, $i=12+m$,  $m\in[2,3]$: 
the number $\bd_m^\inte(\C)$
 of  interior-edges with bond multiplicity $m$ in  $\C$; 
 i.e., $\bd_m^\inte(\C)\triangleq \{e\in E^\inte(\C)\mid \beta(e)=m\}$.

\item $\dcp_i(\C)$, $i=14+[\ta]^\inte$, 
 $\ta\in \Lambda^\inte(D_\pi)$: 
 the frequency $\na_\ta^\inte(\C)=|V_\ta(\C)\cap V^\inte(\C) |$ 
 of chemical element $\ta$ in
 the set $V^\inte(\C)$ of  interior-vertices in  $\C$. 
 
\item $\dcp_i(\C)$, 
$i=14+|\Lambda^\inte(D_\pi)|+[\ta]^\ex$, 
 $\ta\in \Lambda^\ex(D_\pi)$: 
 the frequency $\na_\ta^\ex(\C)=|V_\ta(\C)\cap V^\ex(\C) |$
  of chemical element $\ta$ in
 the set $V^\ex(\C)$ of  exterior-vertices in  $\C$. 
 
\item $\dcp_i(\C)$, 
$i=14+|\Lambda^\inte(D_\pi)|+|\Lambda^\ex(D_\pi)|+ [\gamma]$, 
$\gamma \in \Gamma^\inte(D_\pi)$: 
the frequency $\ec_{\gamma} (\C)$ of edge-configuration $\gamma$
in the set $E^\inte(\C)$ of interior-edges   in  $\C$. 

\item $\dcp_i(\C)$, 
$i=14+|\Lambda^\inte(D_\pi)|+|\Lambda^\ex(D_\pi)|+ |\Gamma^\inte(D_\pi)|
+ [\gamma]$, 
$\gamma \in \Gamma^\lnk(D_\pi)$: 
the frequency $\ec_{\gamma} (\C)$ of edge-configuration $\gamma$
in the set $\Elnk(\C)$ of link-edges   in  $\C$. 
This descriptor is newly introduced in this paper to feature
link-edges of polymers. 
 
\item $\dcp_i(\C)$, 
$i=14+|\Lambda^\inte(D_\pi)|+|\Lambda^\ex(D_\pi)|+ |\Gamma^\inte(D_\pi)|
+ [\mu]$, 
$\mu\in \Ldg^\inte$: 
the frequency of chemical symbols $\mu=\alpha(u)\deg_{\anC}(u)$ 
 of connecting-vertices $u$   in  $\C$.

\item $\dcp_i(\C)$, 
$i= 14+|\Lambda^\inte(D_\pi)|+|\Lambda^\ex(D_\pi)|
+|\Gamma^\inte(D_\pi)|+|\Gamma^\lnk(D_\pi)|+|\Ldg|+ [\psi]$,  
 $\psi \in \mathcal{F}(D_\pi)$: 
the frequency $\fc_{\psi}(\C)$ of fringe-configuration $\psi $
in the set of ${\rho}$-fringe-trees in  $\C$. 

\item $\dcp_i(\C)$, 
$i= 14+|\Lambda^\inte(D_\pi)|+|\Lambda^\ex(D_\pi)|
+ |\Gamma^\inte(D_\pi)|+|\Gamma^\lnk(D_\pi)|+|\Ldg|
+|\mathcal{F}(D_\pi)|+ [\nu]$,  
 $\nu \in \Gac^\lf$: 
the frequency $\ac_{\nu}^\lf(\C)$ of adjacency-configuration $\nu$
in the set of leaf-edges in  $\anC$. 
\end{enumerate}

%% file: Specification_2LMM_polymer.tex
\section{Specifying Target Chemical Graphs}\label{sec:specification}

Our definition of topological specification is analogous with the one  by 
 Zhu~et~al.~\cite{ZAHZNA21}
   except for a necessary modification due to our polymer model 
   with link-edges.

\subsection*{Seed Graph}

A  {\em seed graph} for a polymer  is defined
to be a graph $\GC=(\VC,\EC)$  with a specified edge subset $\EC^\lnk$
 such that 
the edge set $\EC$ consists of four sets 
$\Et$, $\Ew$, $\Ez$ and $\Eew$, 
where each of them can be empty, and
 $\EC^\lnk$ is a circular  set in $\GC$ such that 
  $\emptyset\neq \EC^\lnk\subseteq \Et\cup \Ew\cup \Eew$. 
Figure~\ref{fig:specification_example_1}(a) illustrates an example of a seed graph,
where $\VC=\{u_1,u_2,\ldots,u_{14}\}$, 
$\Et=\{a_1,a_2,a_3,a_4\}$, 
$\Ew=\{a_5,a_6,\ldots,a_9\}$,
$\Ez=\{a_{10}\}$,
$\Eew=\{a_{11},a_{12},\ldots,a_{18}\}$ and 
$\EC^\lnk=\{a_1,a_2\}$.

 A {\em subdivision} $S$ of $\GC$  
is a graph constructed from a seed graph $\GC$ 
according to the following rules:
\begin{enumerate}[leftmargin=*]
\item[-]
Each edge $e=uv\in \Et$ is replaced
with a $u,v$-path $P_e$ of length at least 2;

\item[-] 
Each edge $e=uv\in \Ew$ is replaced
with a $u,v$-path $P_e$ of length at least 1
(equivalently $e$ is directly used or replaced with
a $u,v$-path $P_e$ of length at least 2);

\item[-] 
Each edge $e\in \Ez$ is either used or discarded;   and

\item[-]
Each edge $e\in \Eew$ is always used directly.
\end{enumerate}

The set of link-edges in the monomer representation  $\C$ of 
an inferred polymer 
consists of edges in $\EC^\lnk\cap( \Eew\cup \Ew)$
or edges  in   paths $P_e$ for all edges $e=uv\in \EC^\lnk\cap (\Ew\cup \Et)$
in a  subdivision  $S$ of $\GC$. 
 
A target chemical graph $\C=(H,\alpha,\beta)$ will contain  $S$  as a subgraph
of the interior $H^\inte$ of $\C$.


\subsection*{Interior-specification}

A graph $H^*$ that serves as the interior $H^\inte$ of
a target chemical graph $\C$ will be constructed as follows.
First construct a subdivision  $S$ of a seed graph $\GC$ 
by replacing each edge $e=u u'\in \Et\cup\Ew$
with a pure $u,u'$-path $P_e$.
Next construct a supergraph $H^*$ of $S$ by 
attaching a leaf path $Q_v$ at each vertex $v\in \VC$ or
at an internal vertex $v\in V(P_e)\setminus\{u,u'\}$ 
of each pure $u,u'$-path $P_e$ for some edge $e=uu'\in \Et\cup\Ew$,
where possibly $Q_v=(v), E(Q_v)=\emptyset$ 
(i.e., we do not attach any new edges to $v$).
We introduce the following rules for specifying
 the size of $H^*$, the length $|E(P_e)|$  of
a pure path  $P_e$,  the length $|E(Q_v)|$ of
a   leaf path $Q_v$, the number of  leaf paths $Q_v$
and a bond-multiplicity of each interior-edge,
where we call the set of prescribed constants  
 an  {\em interior-specification}   $\sint$: 
\begin{enumerate}[leftmargin=*]
 \item[-]
  Lower and upper bounds $\nint_\LB, \nint_\UB\in \Z_+$ 
  on   the number of interior-vertices 
of a target chemical graph~$\C$. 

 \item[-]
  Lower and upper bounds $\nlnk_\LB, \nlnk_\UB\in \Z_+$ 
  on   the number of link-edges 
of a target chemical graph~$\C$. 
  
\item[-] 
For each edge $e=u u'\in \Et\cup\Ew$, 
\begin{description} 
\item[]
 a lower bound $\ell_{\LB}(e)$ and 
 an upper bound $\ell_{\UB}(e)$  on the length $|E(P_e)|$ of
 a pure $u,u'$-path $P_e$. 
(For a notational convenience, set 
$\ell_\LB(e):=0$, $\ell_\UB(e):=1$, $e\in \Ez$ and
$\ell_\LB(e):=1$, $\ell_\UB(e):=1$, $e\in \Eew$.)
   
\item[]  
 a lower bound $\bl_{\LB}(e)$ and 
 an upper bound $\bl_{\UB}(e)$ on the number of leaf paths $Q_v$ attached 
 at  internal vertices $v$ of a pure $u,u'$-path $P_e$.   

\item[] 
 a lower bound $\ch_{\LB}(e)$ and 
 an upper bound $\ch_{\UB}(e)$  on the maximum 
 length  $|E(Q_v)|$ of a leaf path $Q_v$ attached  
 at an internal vertex $v\in V(P_e)\setminus\{u,u'\}$ 
 of a pure $u,u'$-path $P_e$.   
\end{description} 

\item[-]
For each vertex $v\in \VC$, 
\begin{description} 
\item[]
 a lower bound $\ch_{\LB}(v)$ and 
 an upper bound $\ch_{\UB}(v)$  on  
 the number of leaf paths $Q_v$ attached to $v$,
 where $0\leq \ch_{\LB}(v)\leq \ch_{\UB}(v)\leq 1$.
 
\item[]
 a lower bound $\ch_{\LB}(v)$ and 
 an upper bound $\ch_{\UB}(v)$  on the
 length $|E(Q_v)|$ of a leaf path $Q_v$ attached to $v$. 
\end{description}  

\item[-] 
For each edge $e=u u'\in \EC$, 
a lower bound $\bd_{m, \LB}(e)$ 
and an  upper bound $\bd_{m, \UB}(e)$  on
the number of edges with bond-multiplicity $m\in [2,3]$ in
$u,u'$-path $P_e$, where we regard $P_e$, $e  \in \Ez\cup \Eew$ 
as single edge $e$.
\end{enumerate}

We call a graph $H^*$ that satisfies an interior-specification $\sint$
a {\em $\sint$-extension of $\GC$}, 
where the bond-multiplicity of each edge has been determined.

Table~\ref{table:interior-spec}  shows an example of 
an interior-specification  $\sint$ to the seed graph  $\GC$ in 
Figure~\ref{fig:specification_example_1}. 

\begin{table}[h!]\caption{Example~1 of an interior-specification  $\sint$. }
\begin{tabular}{ |  c | c | c | c |  } \hline 
$\nint_\LB=20$ & $\nint_\UB = 30$ & 
$\nlnk_\LB=2$ & $\nlnk_\UB = 24$ \\\hline 
\end{tabular}

 \begin{tabular}{ |  c | c c c c c c c c c |  } \hline
              & $a_1$ &  $a_2$ &   $a_3$ &   $a_4$ &   $a_5$ &   $a_6$ &   $a_7$ &   $a_8$  &   $a_9$   \\\hline
 $\ell_\LB(a_i)$ &  2 &  4 &  3 & 2 &  2 &  1  &  1 &  1 &   1\\ \hline
 $\ell_\UB(a_i)$ &  3 & 6 &  6 & 5 &  3 &  3  &  6 &  2 &   6 \\\hline
 $\bl_\LB(a_i)$ &   0 &  1 & 1 & 0 &  0 &   0 &  0 &   0 &  0 \\ \hline
 $\bl_\UB(a_i)$ &  1 &  4 &  4 & 3 &  2 &   1 &  1 &  1  &  1 \\\hline
 $\ch_\LB(a_i)$ &   0 &  2 &  1 & 0 &  0 &  0 &  0 &   0 &   0 \\ \hline
 $\ch_\UB(a_i)$ &  3 &  6 &  6 & 3 &  3 &   3 &  3 &   0 &   0 \\\hline
\end{tabular} 

\begin{tabular}{ |  c | c c c c c c   c c c c  c c c  c|  } \hline
                        & $u_1$ &  $u_2$ &   $u_3$ &   $u_4$ &   $u_5$ &   $u_6$ 
                       & $u_7$ &   $u_8$ &   $u_9$ &   $u_{10}$ &   $u_{11}$ 
                       &   $u_{12}$ &   $u_{13}$ &   $u_{14}$ \\\hline 
 $\bl_\LB(u_i)$ &  0 & 0 & 0 & 0 & 0 &  0 & 0 & 0 & 1 & 0 & 0 & 0 & 0 & 0 \\ \hline
 $\bl_\UB(u_i)$&  1 & 1 & 1 & 1 & 1 &  1 & 1 & 1 & 1 & 1 & 1 & 1 & 1 & 1 \\\hline
 $\ch_\LB(u_i)$&  0 & 0 & 0 & 0 & 0 &  0 & 0 & 0 & 1 & 0 & 0 & 0 & 0 & 0 \\\hline
 $\ch_\UB(u_i)$& 4 & 4 & 4 & 4 & 4 &  4 & 4 & 4 & 6 & 4 & 4 & 4 & 4 & 4 \\\hline
\end{tabular} 

\begin{tabular}{ |  c | c c c c c c   c c c c c c  c c c c c c |  } \hline
                               & $a_1$ &  $a_2$ &   $a_3$ &   $a_4$ &   $a_5$ &   $a_6$ 
                               & $a_7$ &  $a_8$ &   $a_9$ &   $a_{10}$ &   $a_{11}$ &   $a_{12}$ 
                               & $a_{13}$ &   $a_{14}$ &   $a_{15}$ &   $a_{16}$   &   $a_{17}$ &   $a_{18}$
                                    \\\hline
 $\bd_{2, \LB}(a_i)$ &  0 & 0 & 0 & 0 & 0 &  0 & 0 & 0 & 0 & 0 & 0 & 0 & 0 & 1  & 0 & 0 & 0 & 0\\ \hline 
 $ \bd_{2, \UB}(a_i)$& 1 & 2 & 1 & 1 & 1 &  1 & 1 & 1 & 1 & 1 & 1 &  1 & 1 & 1 & 1 & 1 &  1 & 1\\ \hline
 $\bd_{3, \LB}(a_i)$ &  0 & 0 & 0 & 0 & 0 &  0 & 0 & 0 & 0 & 0 & 0 & 0 & 0 & 0  & 0 & 0 & 0 & 0\\ \hline
 $ \bd_{3, \UB}(a_i)$& 1 & 1 & 1 & 1 & 1 &  1 & 1 & 1 & 1 & 1 & 1 &  1 & 1 & 1 & 1 &  1 & 1 & 1\\ \hline
\end{tabular} 
\label{table:interior-spec}  
\end{table}

Figure~\ref{fig:test_subgraph_polymer} illustrates an example of 
an $\sint$-extension $H^*$ of seed graph  $\GC$ in 
Figure~\ref{fig:specification_example_1}(a)
under the interior-specification  $\sint$ in 
Table~\ref{table:interior-spec}.


\subsection*{Chemical-specification}
 
 Let $H^*$ be a graph that serves as 
 the interior $H^\inte$ of a target chemical graph $\C$,
 where the bond-multiplicity of each edge in $H^*$ has be determined.
 Finally we introduce a set of rules for constructing 
   a target chemical graph $\C$ from $H^*$ 
   by choosing  a chemical element $\ta\in \Lambda$ 
and assigning a ${\rho}$-fringe-tree $\psi$
 to each interior-vertex $v\in V^\inte$. 
We introduce the following rules for specifying
the size of $\C$, a set of chemical rooted trees  
that are allowed to use as  ${\rho}$-fringe-trees 
and lower and upper bounds on the frequency of
a chemical element, a chemical symbol, 
 an edge-configuration, and a fringe-configuration
where we call the set of prescribed constants   
 a  {\em chemical specification} $\sce$:   
 
\begin{enumerate}[leftmargin=*]
\item[-] 
Lower and upper bounds $n_\LB,  n^*\in \Z_+$
on the number of vertices, where $\nint_\LB \leq n_\LB\leq n^*$.
 
\item[-] 
A subset $\mathcal{F}^* \subseteq \mathcal{F}(D_\pi)$  
 of chemical rooted trees $\psi$ with $\h(\anpsi)\leq {\rho}$, where 
 we require that 
 every ${\rho}$-fringe-tree $\C[v]$ rooted at an interior-vertex $v$ 
    in  $\C$  belongs to $\mathcal{F}^*$.  
Let   
$\Lambda^\ex$ denote the set of  chemical elements assigned to non-root
vertices over all chemical rooted trees in $\mathcal{F}^*$.  
 
\item[-] 
A subset  $\Lambda^\inte\subseteq \Lambda^\inte(D_\pi)$, where 
 we require that every chemical element $\alpha(v)$ 
 assigned to an interior-vertex  $v$ in $\C$ belongs to $\Lambda^\inte$.
Let $\Lambda:= \Lambda^\inte\cup \Lambda^\ex$ and
 $\na_\ta(\C)$ (resp., $\na_\ta^\inte(\C)$ and $\na_\ta^\ex(\C)$) 
 denote the number of vertices   (resp.,   interior-vertices and  exterior-vertices)
  $v$ such that $\alpha(v)=\ta$   in  $\C$.
 
\item[-] 
A set $\Ldg^\inte\subseteq \Lambda\times [1,4]$  of chemical  symbols.

\item[-] 
Subsets $\Gamma^\lnk\subseteq \Gamma^\inte$ of $\Gamma^\inte(D_\pi)$  
of  edge-configurations  $(\mu,\mu' ,m)$ with $\mu \leq \mu'$, where 
 we require that the edge-configuration $\ec(e)$ of an interior-edge (resp., a link-edge) $e$ in $\C$ 
 belongs to $\Gamma^\inte$ (resp.,    $\Gamma^\lnk$).
We do not distinguish  $(\mu,\mu' ,m)$ and $(\mu' , \mu,m)$.  
  
\item[-] 
Define  $\Gac^\inte$  (resp.,    $\Gac^\lnk$)  to be the set of   adjacency-configurations such that  
$\Gac^\typ:=\{(\ta, \tb, m) \mid (\ta d, \tb d',m)\in \Gamma^\typ\}, \typ\in\{\inte,\lnk\}$.   
Let  $\ac_\nu^\inte(\C), \nu\in \Gac^\inte$  
(resp.,  $\ac_\nu^\lnk(\C), \nu\in \Gac^\lnk$)   
denote  the number of  interior-edges (resp.,  link-edges) $e$
 such that $\ac(e)=\nu$  in $\C$.
  
\item[-] 
 Subsets $\Lambda^*(v)\subseteq \{\ta\in \Lambda^\inte\mid \val(\ta)\geq 2\}$, 
 $v\in \VC$,  
 we require that every chemical element $\alpha(v)$ 
 assigned to   a vertex $v\in  \VC$
 in the seed graph  belongs to $\Lambda^*(v)$.  

\item[-] Lower and upper bound functions 
$\na_\LB,\na_\UB: \Lambda\to  [0,n^*]$  and 
$\na_\LB^\inte,\na_\UB^\inte: \Lambda^\inte\to  [0,n^*]$ 
on the number of   interior-vertices  $v$ such that  $\alpha(v)=\ta$  in $\C$. 

\item[-] Lower and upper bound functions  
$\ns_\LB^\inte,\ns_\UB^\inte: \Ldg^\inte\to  [0,n^*]$ 
  on the number of   interior-vertices $v$ such that $\cs(v)=\mu$  in $\C$.   

\item[-] Lower and upper bound functions  
$\ns_\LB^\cnt,\ns_\UB^\cnt: \Ldg^\inte\to  [0,2]$ 
  on the number of connecting-vertices $v$ such that $\cs(v)=\mu$  in $\C$.   
  
\item[-] Lower and upper bound functions  
$\ac_\LB^\inte,\ac_\UB^\inte: \Gac^\inte \to  \Z_+$ 
($\ac_\LB^\lnk,\ac_\UB^\lnk: \Gac^\lnk \to  \Z_+$)
 on the number of  interior-edges (resp., link-edges) $e$ such that $\ac(e)=\nu$  in $\C$. 

\item[-] Lower and upper bound functions  
$\ec_\LB^\inte,\ec_\UB^\inte: \Gamma^\inte \to  \Z_+$ 
(resp., $\ec_\LB^\lnk,\ec_\UB^\lnk: \Gamma^\lnk \to  \Z_+$)  
 on the number of  interior-edges  (resp., link-edges)  $e$ such that $\ec(e)=\gamma$  in $\C$.  
 
\item[-] Lower and upper bound functions  
$\fc_\LB,\fc_\UB: \mathcal{F}^*\to  [0,n^*]$ 
  on the number of   interior-vertices $v$ 
  such that $\C[v]^\fr$ is r-isomorphic to $\psi\in \mathcal{F}^*$  in $\C$.  
  
 \item[-] Lower and upper bound functions  
$\ac^\lf_\LB,\ac^\lf_\UB: \Gac^\lf \to  [0,n^*]$ 
  on the number of  leaf-edges $uv$ in $\acC$
  with adjacency-configuration $\nu$.  
\end{enumerate}
 
We call a chemical graph $\C$ that satisfies a chemical specification $\sce$
a {\em $(\sint,\sce)$-extension of $\GC$},
and denote by $\mathcal{G}(\GC, \sint,\sce)$ the set of
all $(\sint,\sce)$-extensions of $\GC$. 

Table~\ref{table:chemical_spec}  shows an example of 
a chemical-specification  $\sce$ to the seed graph  $\GC$
 in Figure~\ref{fig:specification_example_1}.

\begin{table}[h!]\caption{Example~2 of a chemical-specification  $\sce$.  
}
\begin{tabular}{ |  l |  } \hline
 $n_\LB=30$,  $n^* =50$. \\\hline
  branch-parameter:   ${\rho}=2$  \\\hline
\end{tabular}

\begin{tabular}{ |  l |  } \hline
 Each of sets $\mathcal{F}(v), v\in \VC$ and
 $\mathcal{F}_E$ is set to be \\
 the set $\mathcal{F}$  of chemical rooted trees $\psi$ with $\h(\anpsi)\leq {\rho}=2$
in Figure~\ref{fig:specification_example_1}(b). \\\hline
\end{tabular}

\begin{tabular}{ |  c | c |   } \hline
  $\Lambda=\{ \ttH,\ttC,\ttN,\ttO, \ttS_{(2)},\ttS_{(6)}, \ttP=\ttP_{(6)},\ttCl\}$ & 
  $\Ldg^\inte =\{ \ttC2 , \ttC3,  \ttC4, \ttN2, \ttN3, \ttO2,
    \ttS_{(2)}2,  \ttS_{(6)}3, \ttP4   \}$  
\\\hline
\end{tabular}

\begin{tabular}{ |  c | l |  } \hline
  $\Gac^{\inte}$ &
  $ \nu_1 \!=\!(\ttC   , \ttC  , 1) ,   \nu_2 \!=\!(\ttC   , \ttC  , 2) ,   
   \nu_3 \!=\!(\ttC   , \ttN  , 1) ,  \nu_4 \!=\!(\ttC  , \ttO  , 1), 
    \nu_5 \!=\! (\ttC, \ttS_{(2)}, 1),\nu_6 \!=\!(\ttC  , \ttS_{(6)}, 1), 
    \nu_7 \!=\! (\ttC  , \ttP  , 1) $  \\ \hline
\end{tabular}

\begin{tabular}{ |  c | l |  } \hline
  $\Gamma^{\inte}$ &
  $ \gamma_1 \!=\! (\ttC 2 , \ttC 2, 1) ,
    \gamma_{2} \!=\!(\ttC 2 , \ttC 2, 2),  
   \gamma_3 \!=\!(\ttC 2 , \ttC 3, 1) ,  
   \gamma_4 \!=\!(\ttC 2 , \ttC 3, 2) ,  
   \gamma_5 \!=\!(\ttC 2 , \ttC 4, 1) , 
   \gamma_6 \!=\!(\ttC 3 , \ttC 3, 1) , 
 $ \\
   &
  $  \gamma_7 \!=\!(\ttC 3 , \ttC 3, 2) ,
    \gamma_8 \!=\!(\ttC 3 , \ttC 4, 1), 
   \gamma_9 \!=\!(\ttC 2 , \ttN 3, 1) ,  
   \gamma_{10} \!=\!(\ttC 3 , \ttN 2, 1) ,   
    \gamma_{11} \!=\!(\ttC 4 , \ttN2, 1), 
   \gamma_{12} \!=\!(\ttC 2 , \ttO 2, 1), $ \\ 
   &
  $  
    \gamma_{13} \!=\!(\ttC 3 , \ttO 2, 1) ,    
    \gamma_{14} \!=\!(\ttC 2, \ttS_{(2)} 2, 1),  
    \gamma_{15} \!=\!(\ttC 3, \ttS_{(2)} 2, 1),  
    \gamma_{16} \!=\!(\ttC 4, \ttS_{(2)} 2, 1),  
    \gamma_{17} \!=\!(\ttC 3 , \ttS_{(6)}3, 1),   $ \\ 
   &
  $  
   \gamma_{18} \!=\!(\ttC 4, \ttS_{(6)}3, 1), 
    \gamma_{19} \!=\!(\ttC 2, \ttP4, 1), 
    \gamma_{20} \!=\!(\ttC 3, \ttP4, 1)  
     $ \\ \hline
\end{tabular}

\begin{tabular}{ |  c | l |  } \hline
  $\Gac^{\lnk}$ &
  $ \nu'_1 \!=\!(\ttC   , \ttC  , 1) ,   \nu'_2 \!=\!(\ttC   , \ttC  , 2) ,   
   \nu'_3 \!=\!(\ttC   , \ttN  , 1),  \nu'_4 \!=\!(\ttC   , \ttS_{(2)}  , 1)  $  \\ \hline
\end{tabular}

\begin{tabular}{ |  c | l |  } \hline
  $\Gamma^{\lnk}$ &
  $ \gamma'_1 \!=\! (\ttC 2 , \ttC 2, 1) ,
   \gamma'_2 \!=\!(\ttC 2 , \ttC 3, 1) ,  
   \gamma'_3 \!=\!(\ttC 2 , \ttC 4, 1) ,  
   \gamma'_4 \!=\!(\ttC 3 , \ttC 3, 1) , 
   \gamma'_5 \!=\!(\ttC 3 , \ttC 3, 2) ,
   \gamma'_6 \!=\!(\ttC 2 , \ttN 3, 1) ,   $ \\
   &
  $   
   \gamma'_7 \!=\!(\ttC 3 , \ttN 2, 1), 
    \gamma'_8 \!=\!(\ttC 2, \ttS_{(2)} 2, 1),  
    \gamma'_9 \!=\!(\ttC 3, \ttS_{(2)} 2, 1),  
    \gamma'_{10} \!=\!(\ttC 4, \ttS_{(2)} 2, 1) $  \\\hline
\end{tabular}

\begin{tabular}{ |  l|  } \hline
$\Lambda^*(u_i)=\{\ttC\}, i\in\{1,2,3,4,5,6,9\}$, 
$\Lambda^*(u_8)=\{{\ttO}\}$, 
$\Lambda^*(u_{12})=\{{\tt C, P}\}$, \\
$\Lambda^*(u_i)=\{\ttC,\ttO,\ttN\}$, $i\in [1,14]\setminus\{1,2,3,4,5,6,8,9,12\}$
   \\\hline
\end{tabular}
  
\begin{tabular}{ |  c | c c c c  c c c c |  } \hline
                         & ${\tt H}$  & ${\tt C}$ &   ${\tt N}$ &     ${\tt O}$ 
                         & $\ttS_{(2)}$ & $\ttS_{(6)}$ & $\ttP$ & $\ttCl$ \\\hline
 $\na_\LB(\ta)$ & 40 &  25 & 1 &  1 & 0 & 0 & 0  & 0 \\ \hline 
 $\na_\UB(\ta)$ & 80 & 50 & 8 &  8  & 4 & 4 & 4 & 4 \\\hline
\end{tabular} 
\begin{tabular}{ |  c | c c c  c c c   |  } \hline
   & $\ttC$ &   $\ttN$ &     $\ttO$  & $\ttS_{(2)}$ & $\ttS_{(6)}$ & $\ttP$  \\\hline
 $\na_\LB^{\inte}(\ta)$ &  10 &  1 & 0  & 0 & 0 & 0      \\ \hline
 $\na_\UB^{\inte}(\ta) $&  25 & 4 & 5 & 2 & 2 & 2  \\\hline
\end{tabular} 

\begin{tabular}{ |  c | c c c c c c  c c c   |  } \hline
    & $\ttC2$ &  $\ttC3$ &   $\ttC4$ & $\ttN2$ &   $\ttN3$ &   $\ttO2$
   & $\ttS_{(2)}2$ & $\ttS_{(6)}3$ & $\ttP4$  \\\hline
 $\ns_\LB^{\inte}(\mu)$ &  3 &   5 & 0  & 0 &  0 &  0 & 0 &  0 &  0    \\ \hline
 $\ns_\UB^{\inte}(\mu) $& 12 & 15 & 5 & 5 &  3 &  5  & 1 & 1 &  1   \\\hline
\end{tabular} 

\begin{tabular}{ |  c | c c c c c c  c c c   |  } \hline
    & $\ttC2$ &  $\ttC3$ &   $\ttC4$ & $\ttN2$ &   $\ttN3$ &   $\ttO2$
   & $\ttS_{(2)}2$ & $\ttS_{(6)}3$ & $\ttP4$  \\\hline
 $\ns_\LB^{\cnt}(\mu)$ &  0 &   0 & 0  & 0 &  0 &  0 & 0 &  0 &  0    \\ \hline
 $\ns_\UB^{\cnt}(\mu) $& 2 & 2 & 2 & 2 & 2 &  2  & 1 & 1 &  0   \\\hline
\end{tabular}   

\begin{tabular}{ |  c | c c c c c c c |  } \hline
         & $\nu_1 $ &   $\nu_2 $ & $\nu_3 $   & $\nu_4 $
         &   $\nu_5 $ & $\nu_6 $   & $\nu_7 $ \\\hline
 $\ac_\LB^{\inte}(\nu)$  &  0  &  0  & 0  &  0  & 0  &  0 & 0     \\ \hline
 $\ac_\UB^{\inte}(\nu)$ & 30 & 10 & 10 &  10 & 2 &  3 &  3 \\\hline
\end{tabular} 

\begin{tabular}{ |  c | c c c c c  c c  |  } \hline
    & $\gamma_1 $ &   $\gamma_2 $ & $\gamma_3 $   & $\gamma_4 $  & $\gamma_5 $
    & $\gamma_i, i\in[6,13]$ &   $\gamma_i, i\in[14,20]$    \\\hline
 $\ec_\LB^{\inte}(\gamma)$ &
    0 &  0 & 0 &  0  & 0 &  0 &  0    \\ \hline
 $\ec_\UB^{\inte}(\gamma) $ &
   4 & 15 & 5 &  5  & 10 & 5  &   2  \\\hline
\end{tabular} 

\begin{tabular}{ |  c | c c c  c  |  } \hline
         & $\nu'_1 $ &   $\nu'_2 $ & $\nu'_3 $  & $\nu'_4 $  \\\hline
 $\ac_\LB^{\lnk}(\nu')$  &  0  &  0  & 0  & 0      \\ \hline
 $\ac_\UB^{\lnk}(\nu')$ &  10 &  5 &  5  &  5   \\\hline
\end{tabular} 
\begin{tabular}{ |  c | c    |  } \hline
    & $\gamma'_i, i\in[1,10]$   \\\hline
 $\ec_\LB^{\lnk}(\gamma')$ &  0    \\ \hline
 $\ec_\UB^{\lnk}(\gamma') $& 4   \\\hline
\end{tabular}

\begin{tabular}{ |  c | c   c   |  } \hline 
& $\psi\in\{\psi_i\mid i=1,6,11\}$ 
& $\psi\in \mathcal{F}^*\setminus \{\psi_i\mid i=1,6,11\}$ \\\hline
 $\fc_\LB(\psi)$  &  1 &    0   \\ \hline 
 $\fc_\UB(\psi)$ &  10 &  3\\\hline
\end{tabular} 

\begin{tabular}{ |  c | c   c   |  } \hline 
& $\nu\in\{(\ttC,\ttC,1),(\ttC,\ttC,2)  \}$ 
& $\nu\in \Gac^\lf \setminus \{(\ttC,\ttC,1),(\ttC,\ttC,2)  \}$   \\\hline
 $\ac^\lf_\LB(\nu)$  &  0 &    0   \\ \hline 
 $\ac^\lf_\UB(\nu)$ &  10 &  8 \\\hline
\end{tabular} 

\label{table:chemical_spec}
\end{table}

Figure~\ref{fig:example_polymer} 
 illustrates an example of 
a   $(\sint,\sce)$-extension of $\GC$   obtained 
from the  $\sint$-extension $H^*$  
 in Figure~\ref{fig:test_subgraph_polymer} 
under the chemical-specification $\sce$ in Table~\ref{table:chemical_spec}.  
  

%% file: Test_instances_2LMM_polymer.tex
\section{Test Instances for Stages~4 and 5}\label{sec:test_instances} 

We prepared the following instances $I_{\mathrm{a}}$ and $I_{\mathrm{b}}$
 for conducting experiments
of  Stages~4  and 5 in Phase~2. 
 
 In Stages~4 and 5, we  use four properties 
 $\pi\in \{${\sc AmD,  HcL, RfId, Tg}$\}$
 and define a set $\Lambda(\pi)$ of chemical elements as follows:  
  $\Lambda(${\sc AmD}$)=\Lambda_4=\{\ttH,\ttC,\ttN,\ttO, \ttCl, \ttS_{(2)} \}$,  
   $\Lambda(${\sc HcL}$)= \Lambda(${\sc Tg}$)=\Lambda_5= \{\ttH, \ttC, \ttO, \ttN,  \ttCl, \ttS_{(2)}, \ttS_{(6)} \}$,
  $\Lambda(${\sc RfId}$)=\Lambda_6=\{\ttH, \ttC, \ttO_{(1)}, \ttO_{(2)}, \ttN,  \ttCl, \ttSi_{(4)}, \ttF \}$  and
  $\Lambda(${\sc Prm}$)=\Lambda_3=\{\ttH, \ttC,  \ttO, \ttN,  \ttCl\}$.

\begin{itemize} 
  \item[(a)]  $I_{\mathrm{a}} =(\GC,\sint,\sce)$: The instance
  used in Appendix~\ref{sec:specification} to explain the target specification.
 For each property $\pi\in \{${\sc AmD,  HcL, RfId, Tg, Prm}$\}$, we replace
 $\Lambda=\{ \ttH,\ttC,\ttN,\ttO, \ttS_{(2)},\ttS_{(6)},\ttP_{(5)}, \ttCl\}$
in Table~\ref{table:chemical_spec} 
 with $\Lambda(\pi)\cap \{\ttS_{(2)},\ttS_{(6)},\ttP_{(5)}, \ttCl\}$
 and  remove from the $\sce$
 all chemical symbols,  edge-configurations and fringe-configurations
  that cannot be constructed from the replaced element set 
 (i.e., those containing a chemical element in 
 $\{\ttS_{(2)},\ttS_{(6)}, \ttP_{(5)}, \ttCl\}\setminus \Lambda(\pi)$).
 
 \end{itemize}
 
\begin{itemize} 
  \item[(b)]  $I_{\mathrm{b}} =(\GC,\sint,\sce)$: An instance that
  represents  a set of polymers that includes the four examples of polymers in 
Fig.~\ref{fig:four_polymers}.
We set a seed graph  $\GC=(\VC,\EC=\Eew)$ to be the graph  
 with two cycles $C_1$ and $C_2$ in Fig.~\ref{fig:test_set_fringe-trees_polymers}(a),
 where we set  
$\Et=\EC^\lnk=\{a_1,a_2\}$ and 
$\Eew=\{a_{3},a_{12},\ldots,a_{14}\}$. \\
Set  $\Lambda:=\Lambda(\pi)$ for each property $\pi\in \{${\sc AmD,  HcL, RfId, Tg}$\}$,
and  set $\Ldg^\inte$ to be
the set of all possible chemical symbols in $\Lambda\times[1,4]$.\\
Set 
$\Gamma^\inte$ (resp.,  $\Gamma^\lnk$)
to be the set of edge-configurations of the interior-edges
(resp.,  the link-edges)
used in the four examples of polymers in Fig.~\ref{fig:four_polymers}.  
Set 
$\Gamma^\inte_\ac$ (resp.,  $\Gamma^\lnk_\ac$) to be
 the set of the adjacency-configurations of the edge-configurations in 
$\Gamma^\inte$ (resp.,  $\Gamma^\lnk$). \\
We specify $n_\LB$ for each property $\pi$
and 
set 
$\nint_\LB:=14$, $\nint_\UB:=n^*:=n_\LB+10$,  
$\nlnk_\LB:=2$,  $\nlnk_\UB: =2+\max\{ n_\LB-15, 0\}$.  \\
For each link-edge $a_i\in\Et=\EC^\lnk=\{a_1,a_2\}$, 
set 
 $\ell_\LB(a_i):=2+\max\{\lfloor (n_\LB-15)/4\rfloor,0\}$,  
 $\ell_\UB(a_i):=\ell_\LB(a_i)+5$,
 $\bl_\LB(a_i):=0, \bl_\UB(a_i):=3$, 
 $\ch_\LB(a_i):=0, \ch_\UB(a_i):=5$,  
 $\bd_{2,\LB}(a_i):=0$ and $\bd_{2,\UB}(a_i):= \lfloor \ell_\LB(a_i)/3 \rfloor$.\\
To form two benzene rings from the two cycles $C_1$ and $C_2$, set 
   $\Lambda^*(u):=\{{\tt C}\}$, 
 $\bl_\LB(u):=\bl_\UB(u):=\ch_\LB(u):=\ch_\UB(u):=0$, $u\in \VC$,
 $\bd_{2,\LB}(a_i):=\bd_{2,\UB}(a_i):=0,  i\in\{3,5,7,9,11,13\}$,
 $\bd_{2,\LB}(a_i):=\bd_{2,\UB}(a_i):=1,  i\in\{4,6,8,10,12,14\}$.\\
Not to include any triple-bond, set 
 $\bd_{3,\LB}(a):=\bd_{3,\UB}(a):=0, a\in \EC$.
 \\
Set lower bounds
 $\na_\LB$,  $\na^\inte_\LB$,  $\ns^\inte_\LB$,  $\ns^\cnt_\LB$, 
$\ac^\inte_\LB$, $\ac_\LB^\lnk$, $\ec_\LB^\inte$, $\ec_\LB^\lnk$ and  $\ac^\lf_\LB$  to be 0. \\
Set  upper bounds   
 $\na_\UB(\ta):=n^*, \na\in\{\ttH,\ttC\}$,   
 $\na_\UB(\ta):=5+\max\{ n_\LB-15, 0\}, \ta\in\{\ttO,\ttN\}$,
 $\na_\UB(\ta):=2+\max\{\lfloor (n_\LB-15)/4\rfloor,0\}, 
 \ta\in\Lambda\setminus \{\ttH,\ttC,\ttO,\ttN\}$,  
  $\ns^\cnt_\UB(\mu):=2, \mu\in\Ldg^\inte$, 
 and   $\na^\inte_\UB$,  $\ns^\inte_\UB$, 
$\ac^\inte_\UB$, $\ac_\LB^\lnk$, $\ec_\UB^\inte$,  $\ec_\UB^\lnk$ 
and  $\ac^\lf_\UB$ to be  $n^*$. \\
Set $\mathcal{F}$ to be the set of the 17 chemical rooted trees $\psi_i,i\in[1,17]$
 in  Fig.~\ref{fig:test_set_fringe-trees_polymers}(b).  
Set $\mathcal{F}_E :=\mathcal{F}(v) := \mathcal{F}$, $v\in \VC$ and 
$\fc_\LB(\psi):=0, \psi\in \mathcal{F}$,
$\fc_\UB(\psi_i):=12+\max\{ n_\LB-15, 0\}, i\in[1,4]$, 
$\fc_\UB(\psi_i):=8+\max\{\lfloor (n_\LB-15)/2\rfloor,0\}, i\in[5,12]$ and
$\fc_\UB(\psi_i):=5+\max\{\lfloor (n_\LB-15)/4\rfloor,0\}, i\in[13,17], \psi_i\in \mathcal{F}$. 

 \end{itemize}

%% file: Constraints_MILP_2LMM_base_polymer.tex
\section{All Constraints in an MILP Formulation for Chemical Graphs}\label{sec:full_milp}

Our definition of an MILP formulation MILP $\mathcal{M}(g,x;\mathcal{C}_2)$
  is analogous with the one  by  Zhu~et~al.~\cite{ZAHZNA21}
   except for a necessary modification due to our polymer model 
   with link-edges. 
 
We define a standard encoding of a finite set $A$ of elements
to be a bijection $\sigma: A \to [1, |A|]$, 
where we denote by $[A]$   the set $[1, |A|]$ of integers
and by $[{\tt e}]$ the encoded element $\sigma({\tt e})$.
Let $\epsilon$ denote {\em null}, a fictitious chemical element 
that does not belong to any set of chemical elements,
chemical symbols, adjacency-configurations and
edge-configurations in the following formulation.
Given a finite set $A$, let $A_\epsilon$ denote the set $A\cup\{\epsilon\}$
and define a standard encoding of $A_\epsilon$
  to be a bijection $\sigma: A \to [0, |A|]$ such that
$\sigma(\epsilon)=0$, 
where we denote by $[A_\epsilon]$   the set $[0, |A|]$ of integers
and by $[{\tt e}]$ the encoded element $\sigma({\tt e})$,
where $[\epsilon]=0$.

 \bigskip 
 Let $\sigma=(\GC,\sint,\sce)$ be a target specification,
 ${\rho}$ denote  the branch-parameter in the specification $\sigma$
 and  $\C$ denote a chemical   graph in $\mathcal{G}(\GC, \sint,\sce)$. 

 \subsection{Selecting  a Cyclical-base} 
\label{sec:co}
 
Recall that  
\[ \begin{array}{ll}
   \Eew = \{e\in \EC\mid \ell_\LB(e)=\ell_\UB(e)=1 \}; &
   \Ez =\{e\in \EC\mid \ell_\LB(e)=0, \ell_\UB(e)=1 \}; \\
  \Ew=\{e\in \EC\mid \ell_\LB(e)=1,  \ell_\UB(e)\geq 2 \}; &
  \Et= \{e\in \EC\mid \ell_\LB(e)\geq 2 \}; \end{array} \]
A subset $\EC^\lnk\subseteq \Eew\cup \Ew\cup \Et$ is given
for introducing link-edges in the monomer representation  $\C$ of 
an inferred polymer.
%
\begin{enumerate} [leftmargin=*]
\item[-]
Every edge $a_i\in \Eew$ is  included in  $\anC$;

\item[-]
Each edge $a_i\in \Ez$ is   included in $\anC$ if necessary;
 
\item[-]
For each edge  $a_i  \in \Et$, edge $a_i$ is not included in $\anC$
and instead a path 
\[P_i=(\vC_{\tail(i)}, \vT_{j-1},\vT_{j},\ldots,
    \vT_{j+t}, \vC_{\hd(i)})\]
     of length at least 2
  from vertex $\vC_{\tail(i)}$ to vertex $\vC_{\hd(i)}$ 
  visiting some  vertices in $\VT$ is constructed in $\anC$; and  
 
\item[-]
For each edge $a_i  \in \Ew$, either  edge $a_i$   is directly used in $\anC$ or
the above path $P_i$ of length at least 2   is constructed in $\anC$.  
 \end{enumerate}
 
Let  $\tC\triangleq |\VC|$ and denote $\VC$ by 
$\{\vC_{i}\mid i\in [1,\tC]\}$.
Regard the seed graph $\GC$ as a digraph such that
each edge $a_i$ with end-vertices $\vC_{j}$ and $\vC_{j'}$
is directed from  $\vC_{j}$ to $\vC_{j'}$ when $j<j'$.
 For each directed edge $a_i  \in \EC $,
 let $\hd(i)$ and $\tail(i)$ denote the head and tail of $\eC(i)$;
 i.e., $a_i=(\vC_{\tail(i)}, \vC_{\hd(i)})$. 
  
Define 
 \[ \kC \triangleq  |\Et\cup \Ew| , ~~ \widetilde{\kC} \triangleq  |\Et| ,\]
 and denote   $\EC=\{a_i\mid i\in[1,\mC]\}$,
 \[\mbox{
$\Et=\{a_k\mid k\in[1,\widetilde{\kC}]\}$,
$\Ew=\{a_k\mid k\in[\widetilde{\kC}+1,\kC]\}$, }\]
 \[\mbox{
$\Ez=\{a_i\mid i\in[\kC+1,\kC+|\Ez|]\}$ and 
$\Eew=\{a_i\mid i\in[\kC+|\Ez|+1,\mC]\}$.}\]
Let $\Iew$ denote the set of indices $i$ of edges $a_i\in \Eew$.
Similarly for $\Iz$, $\Iw$  and $\It$.
Let $\Ilnk$ denote the set of indices $i$ of edges $a_i\in \EC^\lnk$.

To control the construction of such a path $P_i$
 for each edge  $a_k\in  \Et\cup \Ew $,
we regard the index $k\in [1,\kC]$ of each edge $a_k\in  \Et\cup \Ew$
as the ``color'' of the edge.
To introduce necessary linear constraints 
that can construct such a path $P_k$ properly   in our MILP,
we assign the color $k$ to the vertices $\vT_{j-1},\vT_{j},\ldots,$ 
$\vT_{j+t}$ in $\VT$
when the above path  $P_k$ is used in $\anC$.
 
For each index $s\in [1,\tC]$, let  
$\IC(s)$ denote the set of edges $e\in \EC$ incident to vertex $\vC_{s}$,
and 
 $\Eew^+(s)$ (resp., $\Eew^-(s)$) denote the set of 
 edges $a_i\in \Eew$ such that 
the tail (resp., head) of $a_i$ is vertex $\vC_{s}$.
Similarly for 
$\Ez^+(s)$,  $\Ez^-(s)$, $\Ew^+(s)$,  $\Ew^-(s)$,
$\Et^+(s)$ and $\Et^-(s)$.
Let $\IC(s)$ denote the set of indices $i$ of edges $a_i\in \IC(s)$.
Similarly for   
$\Iew^+(s)$,  $\Iew^-(s)$,
$\Iz^+(s)$,  $\Iz^-(s)$, 
$\Iw^+(s)$,  $\Iw^-(s)$,
$\It^+(s)$ and $\It^-(s)$.
Note that $[1, \kC]=\It\cup \Iw$ and 
$[\widetilde{\kC}+1,\mC]=\Iw\cup \Iz\cup\Iew$.

\smallskip\noindent
{\bf constants: } 
\begin{enumerate} [leftmargin=*]
\item[-]
 $n^* \in \Z$: an upper bound  
on the number $n(\C)$ of non-hydrogen atoms in $\C$;  

\item[-] $\tC=|\VC|$, $\widetilde{\kC}=  |\Et|$, $\kC= |\Et\cup \Ew|$,
      $\tT=\nint_\UB-|\VC|$, $\mC=|\EC|$.
      Note that 
      $a_i\in \EC\setminus (\Et\cup \Ew)$ holds $i\in [\kC+1,\mC]$;   

\item[-] 
$\ell_\LB(k), \ell_\UB(k)\in [1, \tT]$, $k\in [1,\kC]$: 
lower and upper bounds on the length of path $P_k$;  
       
\item[-] 
$n_\lnk^{(=1)}=|\Ilnk\cap \Eew|=|\Ilnk\cap \{[\kC+|\Ez|+1,\mC\}|$: 
the number of link-edges created from $\Eew$; 

\item[-] 
$\nlnk_\LB, \nlnk_\UB\in [0,n^*]$:
lower and upper bounds   on   the number of link-edges 
of a target polymer~$\C$;  
\end{enumerate}

\smallskip\noindent
{\bf variables: } 
\begin{enumerate}[leftmargin=*]
\item[-] $\eC(i)\in[0,1]$,  $i\in [1, \mC]$: 
$\eC(i)$ represents edge $a_i\in \EC$, $i\in [1,\mC]$  
 ($\eC(i)=1$, $i\in \Iew$;  $\eC(i)=0$, $i\in \It$)     
  ($\eC(i)=1$ $\Leftrightarrow $   edge $a_i$ is  used in  $\anC$);    
\item[-]  $\vT(i)\in[0,1]$,   $i\in [1,\tT]$:  
  $\vT(i)=1$ $\Leftrightarrow $ vertex $\vT_{i}$ is used in  $\anC$;   
\item[-]  $\eT(i)\in[0,1]$, $i\in [1,\tT+1]$:  $\eT(i)$ represents edge 
$\eT_{i}=(\vT_{i-1}, \vT_{i})\in \ET$,  
where $\eT_{1}$ and $\eT_{\tT+1}$ are fictitious edges
  ($\eT(i)=1$ $\Leftrightarrow $   edge $\eT_{i}$ is  used in  $\anC$);    
\item[-]  $\chiT(i)\in [0,\kC]$, $i\in [1,\tT]$: $\chiT(i)$ represents
 the color assigned to vertex $\vT_{i}$ 
  ($\chiT(i)=k>0$
   $\Leftrightarrow $  vertex $\vT_{i}$ is  assigned color $k$;
   $\chiT(i)=0$ means that vertex $\vT_{i}$ is not used in $\anC$);    
   
\item[-]  $\clrT(k)\in [\ell_\LB(k)-1, \ell_\UB(k)-1]$, $k\in [1,\kC]$, 
$\clrT(0)\in [0, \tT]$: the number of vertices 
$\vT_{i}\in \VT$  with color $c$;
%
\item[-]  $\dclrT(k)\in [0,1]$,   $k\in [0,\kC]$:
      $\dclrT(k)=1$    $\Leftrightarrow $ $\chiT(i)=k$ 
      for some $i\in [1,\tT]$;
      
\item[-]    $\chiT(i,k)\in[0,1]$,  $i\in [1,\tT]$, $k\in [0,\kC]$  
  ($\chiT(i,k)=1$    $\Leftrightarrow $ $\chiT(i)=k$);  
\item[-]   $\tldgC^+(i)\in [0,4]$, $i\in [1,\tC]$: 
the out-degree of vertex $\vC_{i}$ with the used edges $\eC$ in $\EC$; 

\item[-]   $\tldgC^-(i)\in [0,4]$, $i\in [1,\tC]$: 
the in-degree of vertex $\vC_{i}$  with the used edges $\eC$ in $\EC$; 

\item[-]   $n_\lnk\in [\nlnk_\LB, \nlnk_\UB]$: 
the number of link-edges  in $\C$; 
\end{enumerate}
  
\smallskip\noindent
{\bf constraints: }   
\begin{align} 
  \eC(i)=1,  ~~~  i\in \Iew,       &&  \label{eq:co_first}  \\
  \eC(i)=0,  ~~ \clrT(i)\geq 1,   ~~~  i\in \It,     &&   \label{eq:co_first} \\
  \eC(i)+ \clrT(i)\geq 1,  ~~~~~  \clrT(i)\leq \tT\cdot (1-\eC(i) ), 
~~~  i\in \Iw,    &&    \label{eq:co1c} 
\end{align}   
  
\begin{align}  
\sum_{ c\in \Iw^-(i)\cup \Iz^-(i)\cup \Iew^-(i) }\!\!\!\!\!\! \eC(c) 
 = \tldgC^-(i),  ~~ 
\sum_{ c\in \Iw^+(i)\cup \Iz^+(i)\cup \Iew^+(i) }\!\!\!\!\!\! \eC(c) 
 = \tldgC^+(i),  &&   i\in [1,\tC],   \label{eq:co_5}
\end{align}   
\begin{align} 
\chiT(i,0)=1 -\vT(i), ~~~
\sum_{k\in [0,\kC]} \chiT(i,k)=1,  ~~~ 
\sum_{k\in [0,\kC]}k\cdot \chiT(i,k)=\chiT(i),  && i\in[1,\tT],  \label{eq:co2} 
\end{align}   

\begin{align}  
\sum_{i\in[1,\tT]} \chiT(i,k)=\clrT(k), ~~
\tT\cdot \dclrT(k)\geq  \sum_{i\in [1,\tT]} \chiT(i,k)
\geq \dclrT(k), &&  k\in [0,\kC],    \label{eq:co3}   
\end{align}     
 
\begin{align}  
\vT(i-1)\geq \vT(i), && \notag \\
 \kC\cdot (\vT(i-1)-\eT(i )) \geq \chiT(i-1)-\chiT(i )
  \geq \vT(i-1) - \eT(i ), && i\in[2,\tT], \label{eq:co4} 
 \end{align}    
 
\begin{align}  
 \sum_{k\in \Ilnk\cap [1,\kC]} (\clrT(k)+1) + n_\lnk^{(=1)} = n_\lnk. \label{eq:co_last} 
 \end{align}    

\subsection{Constraints for Including Leaf Paths} 
\label{sec:int}

Let
$\widetilde{\tC}$  denote the number of vertices $u\in \VC$ such that 
$\bl_\UB(u)=1$ and assume that 
$\VC=\{u_1,u_2,\ldots, u_p\}$ so that 
\[ \mbox{ 
$\bl_\UB(u_i)=1$, $i\in [1,\widetilde{\tC}]$ and 
$\bl_\UB(u_i)=0$, $i\in[\widetilde{\tC}+1, \tC]$. }\]
Define the set of colors for the vertex set 
$\{u_i\mid i\in [1,\widetilde{\tC}] \}\cup \VT$
 to be $[1,\cF]$ with 
\[ \cF \triangleq \widetilde{\tC} + \tT 
=|\{u_i\mid i\in[1,\widetilde{\tC}]\}\cup \VT|. \]
Let each  vertex   $\vC_{i}$, $i\in[1,\widetilde{\tC}]$ 
(resp., $\vT_{i}\in \VT$)
  correspond to 
a color $i\in [1,\cF]$ (resp., $i+\widetilde{\tC} \in [1,\cF]$). 
When a path $P=(u, \vF_{j}, \vF_{j+1},\ldots, \vF_{j+t})$ 
from a vertex $u\in \VC\cup \VT$ 
  is used in $\anC$, we assign the color $i\in [1,\cF]$ of the vertex $u$
to the vertices $\vF_{j}, \vF_{j+1},\ldots, \vF_{j+t}\in \VF$.

\smallskip\noindent
{\bf constants: } 
\begin{enumerate}[leftmargin=*]
\item[-] $\cF$: the maximum number of different colors 
assigned to the vertices in $\VF$;  

\item[-] $\nint_\LB, \nint_\UB \in [2,n^* ]$:
 lower and upper bounds on
the number of interior-vertices in $\C$; 

\item[-] $\bl_\LB(i) \in [0,1]$,  $i\in [1, \widetilde{\tC}]$: 
a lower   bound  on the number of leaf ${\rho}$-branches  in
the leaf path rooted  at a vertex $\vC_{i}$; 

\item[-]  $\bl_\LB(k),\bl_\UB(k)\in [0,\ell_\UB(k)-1]$, 
 $k\in[1,\kC]=\It\cup\Iw$: 
lower and upper bounds on the number of 
leaf ${\rho}$-branches in the trees rooted at internal vertices 
of a pure path $P_k$  for an edge $a_k\in \Ew\cup \Et$; 
\end{enumerate}

\smallskip\noindent
{\bf variables: } 
\begin{enumerate}[leftmargin=*]
  
\item[-]   $\nint_G\in [\nint_\LB, \nint_\UB]$: 
the number of interior-vertices in $\C$; 

\item[-]  $\vF(i)\in[0,1]$,   $i\in [1,\tF]$:
 $\vF(i)=1$ $\Leftrightarrow $ vertex $\vF_{i}$ is used in  $\C$;   
 
\item[-]  $\eF(i)\in[0,1]$, $i\in [1,\tF+1]$:  $\eF(i)$ represents edge 
$\eF_{i}=\vF_{i-1} \vF_{i}$,  
where $\eF_{1}$ and $\eF_{\tF+1}$ are fictitious edges
  ($\eF(i)=1$ $\Leftrightarrow $   edge $\eF_{i}$ is  used in  $\C$);    
\item[-]  $\chiF(i)\in [0,\cF]$, $i\in [1,\tF]$: $\chiF(i)$ represents
 the color assigned to  vertex $\vF_{i}$  
  ($\chiF(i)=c$ $\Leftrightarrow $  vertex $\vF_{i}$ is  assigned color $c$);   
  
\item[-]  $\clrF(c)\in [0, \tF]$, $c\in [0,\cF]$: the number of vertices $\vF_{i}$
 with color $c$;  
\item[-]  $\dclrF(c)\in [\bl_\LB(c), 1]$,  $c\in [1, \widetilde{\tC}]$:
      $\dclrF(c)=1$    $\Leftrightarrow $ $\chiF(i)=c$ for some $i\in [1,\tF]$;  
\item[-]  $\dclrF(c)\in[0,1]$,  $c\in [\widetilde{\tC}+1,\cF]$:
      $\dclrF(c)=1$    $\Leftrightarrow $ $\chiF(i)=c$ for some $i\in [1,\tF]$;  
\item[-]    $\chiF(i,c)\in[0,1]$,
 $i\in [1,\tF]$, $c\in [0,\cF]$:  
   $\chiF(i,c)=1$    $\Leftrightarrow $ $\chiF(i)=c$;     
\item[-]  $\bl(k,i)\in [0,1]$, $k\in[1,\kC]= \It\cup\Iw$,  $i\in[1,\tT]$: 
    $\bl(k,i)=1$ $\Leftrightarrow$ path $P_k$ contains vertex $\vT_{i}$ 
    as an internal vertex
    and the ${\rho}$-fringe-tree rooted at $\vT_{i}$ contains a leaf ${\rho}$-branch;
\end{enumerate}
  
\smallskip\noindent
{\bf constraints: }   
\begin{align} 
\chiF(i,0)=1 -\vF(i), ~~~
\sum_{c\in [0,\cF]} \chiF(i,c)=1,  ~~~ 
\sum_{c\in [0,\cF]}c\cdot \chiF(i,c)=\chiF(i),  &&  i\in[1,\tF],  \label{eq:int_first} 
\end{align}   

\begin{align}  
\sum_{i\in[1,\tF]} \chiF(i,c)=\clrF(c), ~~~ \tF\cdot \dclrF(c)\geq
\sum_{i\in [1,\tF]} \chiF(i,c)\geq \dclrF(c), &&  c\in [0,\cF],    \label{eq:int3}   
\end{align}   
 
\begin{align}  
 \eF(1)=\eF(\tF+1)=0,  && \label{eq:int4} 
 \end{align}   
 
\begin{align}  
\vF(i-1)\geq \vF(i), && \notag \\
 \cF\cdot (\vF(i-1)-\eF(i)) \geq \chiF(i-1)-\chiF(i) 
 \geq \vF(i-1)- \eF(i), && i\in[2,\tF], \label{eq:int6} 
 \end{align}



\begin{align}  
 \bl(k,i)\geq  \dclrF(\widetilde{\tC} + i)+\chiT(i,k)-1 , ~~~
 ~~~~~   k \in[1,\kC],   i\in[1,\tT], &&    
 \end{align}   
 
\begin{align}  
 \sum_{k \in[1,\kC],  i\in[1,\tT]} \bl(k,i)
 \leq \sum_{i\in[1,\tT]}\dclrF( \widetilde{\tC} +i),   &&    
  \label{eq:int12} 
 \end{align}   
  
  \begin{align}  
 \bl_\LB(k)\leq  \sum_{ i\in[1,\tT]} \bl(k,i) \leq  \bl_\UB(k) , ~~~~~~
     k \in[1,\kC], &&      
       \label{eq:int13} 
 \end{align}

\begin{align}  
 \tC +\sum_{i\in [1,\tT]} \vT(i) + \sum_{i\in [1,\tF]} \vF(i) =\nint_G.  &&  
  \label{eq:int_last} 
 \end{align}   
 

\subsection{Constraints for Including Fringe-trees} \label{sec:ex}
 
 Recall that   $\mathcal{F}(D_\pi)$ denotes the set of 
chemical rooted trees $\psi$  
r-isomorphic to a chemical rooted tree in $\mathcal{T}(\C)$
  over all chemical graphs $\C\in D_\pi$,
  where possibly a chemical rooted tree $\psi\in \mathcal{F}(D_\pi)$
  consists of a single chemical element $\ta\in \Lambda\setminus \{{\tt H}\}$.

To express the condition that
the ${\rho}$-fringe-tree is chosen from a rooted tree $C_i$, $T_i$  or  $F_i$, 
we introduce the following set of variables and constraints.  
  
\smallskip\noindent
{\bf constants: } 
\begin{enumerate}[leftmargin=*]
\item[-]   $n_\LB$: a lower  bound 
on the number $n(\C)$ of non-hydrogen atoms in $\C$,
where $n_\LB, n^*\geq \nint_\LB$;  
\item[-]   $\ch_{\LB}(i),\ch_{\UB}(i)\in [0,n^* ]$, $i\in [1,\tT]$: 
lower and upper bounds on $\h(\langle T_i\rangle)$ of the tree $T_i$ rooted 
at a vertex $\vC_{i}$; 

\item[-]   $\ch_{\LB}(k),\ch_{\UB}(k)\in [0,n^* ]$, $k \in[1,\kC]= \It\cup\Iw$: 
lower and upper bounds on the maximum  
 height $\h(\langle T \rangle)$ of the tree $T\in \F(P_k)$ rooted at 
an internal vertex of a path $P_k$   for an edge $a_k\in \Ew\cup \Et$;  


\item[-]  Prepare a coding of the set  $\mathcal{F}(D_\pi)$ and let 
    $[\psi]$ denote  the coded integer of  
     an element $\psi$ in $\mathcal{F}(D_\pi)$;  

\item[-]   Sets  $\mathcal{F}(v) \subseteq \mathcal{F}(D_\pi), v\in \VC$
and $\mathcal{F}_E \subseteq \mathcal{F}(D_\pi)$ 
 of chemical rooted trees $T$ with $\h(T)\in [1,{\rho}]$;  
 
\item[-]  Define
$\mathcal{F}^*:=\bigcup_{v\in \VC}\mathcal{F}(v)\cup \mathcal{F}_E$, 
 $\FrC_i:= \mathcal{F}(\vC_i)$, $i\in[1,\tC]$,
$\FrT_i:= \mathcal{F}_E$, $i\in[1,\tT]$  and 
$\FrF_i:= \mathcal{F}_E$, $i\in[1,\tF]$;

\item[-]    
 $\fc_\LB(\psi),\fc_\UB(\psi)\in[0,n^*], \psi\in \mathcal{F}^*$:
lower and upper bound functions  
  on the number of   interior-vertices $v$ 
  such that $\C[v]$ is r-isomorphic to $\psi$  in $\C$; 
  
\item[-]  
$\FrX_i[p], p\in [1,{\rho}], \mathrm{X}\in\{\mathrm{C,T,F}\}$:
the set of  chemical rooted trees  $T\in  \FrX_i$
with   $\h(\langle T\rangle)= p$;  

\item[-]  
$n_{\oH}([\psi])\in [0, 3^{\rho}], \psi\in \mathcal{F}^*$: 
the number $n(\langle \psi\rangle)$ 
of non-root hydrogen vertices in a chemical rooted tree  $\psi$; 
 
\item[-]  
$\h_{\oH}([\psi])\in [0,{\rho}], \psi\in \mathcal{F}^*$: 
 the height $\h(\langle \psi\rangle)$  of the
 hydrogen-suppressed chemical rooted tree  $\langle \psi\rangle$; 

\item[-]  
$\deg_\mathrm{r}^{\oH}([\psi])\in [0,3], \psi\in \mathcal{F}^*$: 
the number $\deg_\mathrm{r}(\anpsi)$ of non-hydrogen children of the root $r$
 of a chemical rooted tree  $\psi$; 
 
\item[-]  
$\deghyd_\mathrm{r}([\psi])\in [0,3], \psi\in \mathcal{F}^*$: 
the number $\deg_\mathrm{r}(\psi)-\deg_\mathrm{r}(\anpsi)$ 
of hydrogen children of the root $r$ of a chemical rooted tree  $\psi$; 
 
\item[-] 
$\vion(\psi)\in [-3,+3], \psi\in \mathcal{F}^*$: 
  the ion-valence of the root in  $\psi$;

\item[-] 
  $\ac^\lf_\nu(\psi), \nu\in \Gac^\lf$:
the frequency of leaf-edges with adjacency-configuration $\nu$ in $\psi$;
  
 \item[-] 
$\ac^\lf_\LB,\ac^\lf_\UB: \Gac^\lf \to  [0,n^*]$:
lower and upper bound functions    on the number of  leaf-edges $uv$ in $\acC$
  with adjacency-configuration $\nu$; 
\end{enumerate}

\smallskip\noindent
{\bf variables: }   
\begin{enumerate}[leftmargin=*]
\item[-]
  $n_G\in [n_\LB, n^*]$: the number $n(\C)$ of non-hydrogen atoms in $\C$;  
\item[-] $\vX(i)\in[0,1], i\in [1,\tX]$,   $\mathrm{X}\in\{\mathrm{T,F}\}$: 
 $\vX(i)=1$ $\Leftrightarrow $ vertex $\vX_{i}$ is used in $\C$; 
       
\item[-] 
$\dlfrX(i,[\psi])\in [0,1], 
    i\in[1,\tX], \psi\in \FrX_i, \mathrm{X}\in \{\mathrm{C,T,F}\}$:  
$\dlfrX(i,[\psi])=1$  $\Leftrightarrow $
 $\psi$ is  the ${\rho}$-fringe-tree rooted at vertex $\vX_i$ in $\C$;  

\item[-]
$\fc([\psi])\in [\fc_\LB(\psi),\fc_\UB(\psi)], \psi\in \mathcal{F}^*$:
    the number of   interior-vertices $v$ 
  such that $\C[v]$ is r-isomorphic to $\psi $  in $\C$;  
  
\item[-]    
$\ac^\lf([\nu])\in [\ac^\lf_\LB(\nu),\ac^\lf_\UB(\nu)], \nu\in \Gac^\lf$: 
  the number of leaf-edge with adjacency-configuration $\nu$  in $\C$;  
   
\item[-]
 $\degXex(i)\in [0,3],  i\in [1,\tX],     \mathrm{X}\in\{\mathrm{C,T,F}\}$:
the number of non-hydrogen children of the root
 of  the ${\rho}$-fringe-tree rooted at vertex $\vX_i$ in $\C$;  
  
\item[-]  $\hyddegX(i)\in [0,4]$,  $i\in [1,\tX]$, 
 $\mathrm{X}\in \{\mathrm{C,T,F}\}$: 
 the number of  hydrogen atoms adjacent to  vertex $\vX_{i}$
 (i.e.,  $\hyddeg(\vX_{i})$) in $\C=(H,\alpha,\beta)$; 
 
\item[-] 
 $\eledegX(i)\in [-3,+3]$,  $i\in [1,\tX]$, 
 $\mathrm{X}\in \{\mathrm{C,T,F}\}$: 
 the  ion-valence $\vion(\psi)$ of vertex $\vX_{i}$
 (i.e.,  $\eledegX(i)=\vion(\psi)$ 
 for the ${\rho}$-fringe-tree $\psi$ rooted at $\vX_{i}$) in $\C=(H,\alpha,\beta)$;

\item[-] $\hX(i)\in [0,{\rho}]$, $i\in [1,\tX]$,
$\mathrm{X}\in \{\mathrm{C,T,F}\}$: the height $\h(\langle T\rangle)$ of
the hydrogen-suppressed chemical rooted tree $\langle T\rangle$ of  
 the ${\rho}$-fringe-tree $T$ rooted at vertex $\vX_i$ in $\C$;  
\item[-] $\sigma(k,i)\in[0,1]$, $k \in[1,\kC]=\It\cup\Iw,  i\in [1,\tT]$: 
    $\sigma(k,i)=1$ $\Leftrightarrow$ 
    the ${\rho}$-fringe-tree $T_v$ rooted at  vertex $v=\vT_{i}$ 
      with color $k$  has the largest height $\h(\langle \T_v \rangle)$ among such trees
      $T_v, v\in \VT$;
\end{enumerate}

\smallskip\noindent
{\bf constraints: }    
\begin{align}    
\sum_{\psi\in \FrC_i}\!\!\dlfrC(i,[\psi]) =1, &&  i\in [1,\tC], \notag \\
\sum_{\psi\in \FrX_i }\!\!\dlfrX(i,[\psi]) =\vX(i),  
&&   i\in [1,\tX],  \mathrm{X}\in\{\mathrm{T,F}\},  \label{eq:ex_first}   
\end{align}     
 
\begin{align}    
\sum_{\psi\in \FrX_i }\!\! \deg_\mathrm{r}^{\oH}([\psi]) \cdot \dlfrX(i,[\psi]) 
 = \degXex(i),  &&  \notag \\
\sum_{\psi\in \FrX_i }\!\!   \deghyd_\mathrm{r}([\psi]) \cdot \dlfrX(i,[\psi])
 = \hyddegX(i),   &&  \notag \\
\sum_{\psi\in \FrX_i }\!\! \vion([\psi]) \cdot \dlfrX(i,[\psi]) 
 = \eledegX(i),
&&   i\in [1,\tX],  \mathrm{X}\in\{\mathrm{C,T,F}\},  \label{eq:ex_1}  
\end{align}    
 
\begin{align}    
\sum_{\psi\in \FrF_i[{\rho}] }\dlfrF(i,[\psi]) 
\geq \vF(i) - \eF(i+1),
      && i\in [1,\tF]~(\eF(\tF+1)=0), \label{eq:ex3}  
\end{align}   
 
\begin{align}   
\sum_{\psi\in \FrX_i } \h_{\oH}([\psi]) \cdot \dlfrX(i,[\psi]) =  \hX(i), && 
   i\in[1,\tX],  \mathrm{X}\in \{\mathrm{C,T,F}\}, \label{eq:ex5}  
\end{align}   

\begin{align}     
\sum\limits_{\substack{  \psi\in \FrX_i \\
              i\in [1,\tX],    \mathrm{X}\in \{\mathrm{C,T,F}\}  }}\!\!
                 n_{\oH}([\psi])  \cdot \dlfrX(i,[\psi])    
 +    \sum_{  i\in [1,\tX], \mathrm{X}\in \{\mathrm{T,F}\} } \vX(i)
   +\tC
  = n_G, ~~
 && 
  \label{eq:ex2} 
\end{align}   

\begin{align} 
\sum_{ i\in [1,\tX], \mathrm{X}\in\{\mathrm{C,T,F}\}} 
\dlfrX(i,[\psi]) =  \fc([\psi]), &&  \psi\in \mathcal{F}^*, 
 \label{eq:ex3} 
\end{align}    

\begin{align}    
\sum_{\psi\in \FrX_i, i\in[1,\tX], \mathrm{X}\in \{\mathrm{C,T,F}\}}
\ac^\lf_\nu(\psi)\cdot \dlfrX(i,[\psi]) = \ac^\lf([\nu]), &&
\nu\in \Gac^\lf, \label{eq:ex4}  
\end{align}   

\begin{align}  
\hC(i)    \geq \ch_\LB(i)- n^* \cdot \dclrF(i),  ~~
\clrF(i)+{\rho} \geq \ch_\LB(i) , ~~~~~~~~~~~~~~   &&\notag \\
\hC(i)          \leq \ch_\UB(i) ,  ~~
\clrF(i)+{\rho} \leq \ch_\UB(i)+ n^* \cdot (1-\dclrF(i)),  ~ 
        &&   i\in [1,\widetilde{\tC}],         
            \label{eq:int14} 
 \end{align}   
 
\begin{align}  
 \ch_\LB(i) \leq  \hC(i)   \leq  \ch_\UB(i) ,  ~~ 
        &&    i\in [\widetilde{\tC}+1,\tC],       
             \label{eq:int14} 
 \end{align}   
 
\begin{align}   
 \hT(i)    \leq \ch_\UB(k) 
  + n^*\cdot (\dclrF( \widetilde{\tC}+ i)+1-\chiT(i,k)),  &&\notag \\
\clrF(\widetilde{\tC}+i)+{\rho}
 \leq \ch_\UB(k)+ n^*\cdot (2-\dclrF( \widetilde{\tC}+ i)-\chiT(i,k)),    
&& k \in[1,\kC],  i\in [1,\tT],      
    \label{eq:int15} 
 \end{align}   
 
\begin{align}   
 \sum_{i\in[1,\tT]}\sigma(k,i) =\dclrT(k),   &&   k \in[1,\kC],      
 \label{eq:int16} 
 \end{align}   
 
\begin{align}  
 \chiT(i,k)\geq \sigma(k,i), && \notag\\
 \hT(i)    \geq \ch_\LB(k) - n^*\cdot (\dclrF( \widetilde{\tC}+ i)+1-\sigma(k,i) ),
  && \notag\\ 
\clrF(\widetilde{\tC}+i)+{\rho}
 \geq \ch_\LB(k) - n^* \cdot (2-\dclrF( \widetilde{\tC}+ i)-\sigma(k,i)),   
&&  k \in[1,\kC],  i\in [1,\tT]. 
    \label{eq:ex_last} 
 \end{align}   

\subsection{Descriptor for the  Number of Specified Degree} 
\label{sec:Deg}

We include constraints to compute descriptors for degrees in $\C$. \\

\smallskip\noindent
{\bf variables: } 
\begin{enumerate}[leftmargin=*]
\item[-]  $\degX(i)\in [0,4]$,  $i\in [1,\tX]$, 
 $\mathrm{X}\in \{\mathrm{C,T,F}\}$: 
 the number of non-hydrogen atoms adjacent to  vertex $v=\vX_{i}$
 (i.e.,  $\deg_{\anC}(v)=\deg_H(v)-\hyddeg_{\C}(v)$) in $\C=(H,\alpha,\beta)$; 

\item[-] $\degCT(i)\in [0,4]$,  $i\in [1, \tC]$: the number of edges
from vertex $\vC_{i}$ to vertices $\vT_{j}$, $j\in [1,\tT]$;  
\item[-]  $\degTC(i)\in [0,4]$,  $i\in [1, \tC]$: the number of edges
from  vertices $\vT_{j}$, $j\in [1,\tT]$ to vertex $\vC_{i}$;    
\item[-]    $\ddgC(i,d)\in[0,1]$,  $i\in [1,\tC]$, $d\in [1,4]$, 
  $\ddgX(i,d)\in[0,1]$,  $i\in [1,\tX]$,
 $d\in [0,4]$,  $\mathrm{X}\in \{\mathrm{T,F}\}$: 
        $\ddgX(i,d)=1$ $\Leftrightarrow$   $\degX(i)+\hyddegX(i)=d$;  
       
\item[-]   $\dg(d)\in[\dg_\LB(d),\dg_\UB(d)]$,  $d \in[1,4]$:
    the number  of interior-vertices $v$ with 
       $\mathrm{deg}_H(\vX_{i})=d$   in $\C=(H,\alpha,\beta)$;

\item[-] $\degCint(i)\in [1,4]$,  $i\in [1, \tC]$, 
 $\degXint(i)\in [0,4]$,  $i\in [1, \tX], \mathrm{X}\in \{\mathrm{T,F}\}$: 
the interior-degree $\deg_{H^\inte}(\vX_i)$ 
  in the interior $H^\inte=(V^\inte(\C),E^\inte(\C))$ of  $\C$; i.e., 
the number of interior-edges incident to vertex $\vX_{i}$;

\item[-]    $\ddgCint(i,d)\in[0,1]$,  $i\in [1,\tC]$,  $d\in [1,4]$,  
  $\ddgXint(i,d)\in[0,1]$,  $i\in [1,\tX]$,
 $d\in [0,4]$,  $\mathrm{X}\in \{\mathrm{T,F}\}$: 
       $\ddgXint(i,d)=1$ $\Leftrightarrow$   $\degXint(i)=d$;  
       
\item[-]   $\dg^\inte(d)\in[\dg_\LB(d),\dg_\UB(d)]$,  $d \in[1,4]$:
    the number  of interior-vertices $v$ with
    the interior-degree  $\deg_{H^\inte}(v)=d$
  in the interior $H^\inte=(V^\inte(\C),E^\inte(\C))$ of  $\C=(H,\alpha,\beta)$.
  
\end{enumerate}
   
\smallskip\noindent
{\bf constraints: }   
\begin{align}   
\sum_{   k\in \It^+(i)\cup \Iw^+(i)} \dclrT(k) = \degCT(i), ~~
 \sum_{   k\in \It^-(i)\cup \Iw^-(i)} \dclrT(k) = \degTC(i), 
    &&    i\in [1, \tC],     \label{eq:Deg_first}  
\end{align}

\begin{align}   
\tldgC^-(i)+\tldgC^+(i)   + \degCT(i)  + \degTC(i) + \dclrF(i) = \degCint(i),  
    &&    i\in [1, \widetilde{\tC}],     \label{eq:Deg2}  
\end{align}   

\begin{align}      
\tldgC^-(i)+\tldgC^+(i)  + \degCT(i)  + \degTC(i)   = \degCint(i),  
     &&   i\in [\widetilde{\tC}+1,\tC],     \label{eq:Deg2b}  
\end{align}   

\begin{align}      
  \degCint(i)+ \degCex(i) = \degC(i),  
    &&    i\in [1, \tC],     \label{eq:Deg2c}  
\end{align}   

\begin{align}    
\sum_{\psi\in \FrC_i[{\rho}] }\dlfrC(i,[\psi]) \geq 2-\degCint(i)
      &&  i\in [1, \tC],    \label{eq:Deg2d} 
\end{align}   

\begin{align}   
  2\vT(i)   + \dclrF(\widetilde{\tC}+i)    =\degTint(i),   && \notag \\
 \degTint(i)+ \degTex(i)  =\degT(i),   && 
  i\in [1,\tT]~(\eT(1)=\eT(\tT+1)=0), \label{eq:Deg3}  
\end{align}   

\begin{align}
   \vF(i) +\eF(i+1)  =\degFint(i),  && \notag \\
   \degFint(i)  +\degFex(i)   = \degF(i),   && 
  i\in [1,\tF] ~(\eF(1)=\eF(\tF+1)=0),   \label{eq:Deg4}  
\end{align} 

\begin{align}   
\sum_{d\in [0,4]}\ddgX(i,d)=1, ~
\sum_{d\in [1,4]}d\cdot\ddgX(i,d)=\degX(i)+\hyddegX(i), && \notag \\
\sum_{d\in [0,4]}\ddgXint(i,d)=1, ~
\sum_{d\in [1,4]}d\cdot\ddgXint(i,d)=\degXint(i), && 
 i\in [1,\tX],  \mathrm{X}\in \{\mathrm{T, C, F}\}, \label{eq:Deg5}  
\end{align}   
 
\begin{align}   
\sum_{ i\in [1,\tC]} \ddgC(i,d) + \sum_{ i\in [1,\tT]} \ddgT(i,d) 
 + \sum_{ i\in [1,\tF] }  \ddgF(i,d) = \dg(d),   &&   \notag \\
\sum_{ i\in [1,\tC]} \ddgCint(i,d) + \sum_{ i\in [1,\tT]} \ddgTint(i,d) 
 + \sum_{ i\in [1,\tF] }  \ddgFint(i,d) = \dg^\inte(d),     
    && d\in [1,4].  
  \label{eq:Deg_last}  
\end{align}   

\subsection{Assigning Multiplicity} 
\label{sec:beta}

 We prepare an integer variable $\beta(e)$  
 for each edge $e$ in the scheme graph $\mathrm{SG}$ 
 to denote the bond-multiplicity of $e$ in a selected graph $H$ and
 include necessary constraints for the variables to satisfy in $H$. 
 
\smallskip\noindent
{\bf constants: }
\begin{enumerate}[leftmargin=*]
\item[-]
$\betar([\psi])$: the sum $\beta_\psi(r)$ of bond-multiplicities of edges
incident to  the root $r$ of a chemical rooted tree $\psi\in \mathcal{F}^*$; 
\end{enumerate}

\smallskip\noindent
{\bf variables: } 
\begin{enumerate}[leftmargin=*]
\item[-] $\bX(i)\in [0,3]$,   $i\in [2,\tX]$, $\mathrm{X}\in \{\mathrm{T,F}\}$:   
 the bond-multiplicity of edge  $\eX_{i}$ in $\C$;  
 
\item[-] $\bC(i)\in [0,3]$,     $i\in [\widetilde{\kC}+1,\mC]= \Iw\cup \Iz\cup\Iew$:    
     the bond-multiplicity of 
     edge  $a_{i}\in \Ew\cup \Ez\cup\Eew$ in $\C$;        
\item[-]
   $\bCT(k), \bTC(k)\in [0,3]$, $k\in [1, \kC]=\It\cup \Iw$: 
   the bond-multiplicity of the first (resp., last) edge of the pure path $P_k$ in $\C$;    
   
\item[-]
   $\bsF(c)\in [0,3], c\in [1,\cF=\widetilde{\tC} + \tT ]$:  
   the bond-multiplicity of the first edge of the leaf path $Q_c$
   rooted at vertex $\vC_{c}, c\leq\widetilde{\tC} $
    or $\vT_{c-\widetilde{\tC}}, c>\widetilde{\tC} $  in $\C$;   
    
\item[-] $\bXex(i)\in [0,4],  i\in [1,\tX],   \mathrm{X}\in\{\mathrm{C,T,F}\}$:
the sum $\beta_{\C[v]}(v)$ of bond-multiplicities of edges in the ${\rho}$-fringe-tree
$\C[v]$ rooted at  interior-vertex $v=\vX_{i}$;  

\item[-] $\delbX(i,m)\in [0,1]$, $i\in [2,\tX]$,   $m\in[0,3]$, 
       $\mathrm{X}\in \{\mathrm{T,F}\}$:  
  $\delbX(i,m)=1$  $\Leftrightarrow$  $\bX(i)=m$; 
\item[-] $\delbC(i,m)\in [0,1]$,  
   $i\in [\widetilde{\kC},\mC]=\Iw\cup \Iz\cup\Iew$,  $m\in[0,3]$:  
   $\delbC(i,m)=1$  $\Leftrightarrow$  $\bC(i)=m$; 
\item[-]
   $\delbCT(k,m), \delbTC(k,m)\in [0,1]$, $k\in [1, \kC]=\It\cup \Iw$,  $m\in[0,3]$:
     $\delbCT(k,m)=1$   (resp., $\delbTC(k,m)=1$)    $\Leftrightarrow$  
           $\bCT(k)=m$ (resp., $\bTC(k)=m$); 
           
\item[-]
   $\delbsF(c,m)\in [0,1]$, $c\in [1,\cF]$, 
    $m\in[0,3], \mathrm{X}\in \{\mathrm{C,T}\}$: 
     $\delbsF(c,m)=1$ $\Leftrightarrow$  $\bsF(c)=m$;  
\item[-] $\bd^\inte(m)\in[0, 2\nint_\UB]$, $m\in[1,3]$:
      the number of interior-edges with bond-multiplicity  $m$ in $\C$;  
      
\item[-] $\bdX(m)\in [0,2\nint_\UB],  \mathrm{X}\in \{\mathrm{C,T,CT,TC}\}$,
      $\bdX(m)\in [0,2\nint_\UB], \mathrm{X}\in \{\mathrm{F,CF,TF}\}$, $m\in[1,3]$:  
 the number of interior-edges $e\in \EX$ with bond-multiplicity  $m$ in  $\C$; 
\end{enumerate}
 
\smallskip\noindent
{\bf constraints: } 
\begin{align}    
\eC(i)\leq \bC(i)\leq 3\eC(i), 
  i\in [\widetilde{\kC}+1,\mC]=\Iw\cup \Iz\cup\Iew, \label{eq:beta_first} 
\end{align}   

\begin{align}   
  \eX(i)\leq \bX(i)\leq 3 \eX(i), 
  &&    i\in [2,\tX],   \mathrm{X}\in \{\mathrm{T, F}\},    \label{eq:beta1}  
\end{align}   

\begin{align}   
\dclrT(k)\leq \bCT(k)\leq 3 \dclrT(k), ~~~ 
\dclrT(k)\leq \bTC(k)\leq 3 \dclrT(k), &&  k\in [1, \kC], \label{eq:beta8} \\ 
\dclrF(c)\leq \bXF(c)\leq 3 \dclrF(c), &&   c\in [1,\cF] \label{eq:beta8}  
\end{align}

\begin{align} 
\sum_{m\in[0,3]} \delbX(i,m)=1,  ~~
\sum_{m\in[0,3]}m\cdot \delbX(i,m)=\bX(i), &&   i\in [2,\tX],  
  \mathrm{X}\in \{\mathrm{T,F}\},  \label{eq:beta10}    
\end{align}   

\begin{align} 
\sum_{m\in[0,3]} \delbC(i,m)=1,  ~~
\sum_{m\in[0,3]}m\cdot \delbC(i,m)=\bC(i), &&   i\in [\widetilde{\kC}+1,\mC],   \label{eq:beta11}    
\end{align}   
 
\begin{align}   
\sum_{m\in[0,3]} \delbCT(k,m)=1,    ~~ 
\sum_{m\in[0,3]}m\cdot\delbCT(k,m)=\bCT(k),
&&     k\in [1, \kC],  \notag \\      
\sum_{m\in[0,3]} \delbTC(k,m)=1,  ~~ 
\sum_{m\in[0,3]} m\cdot\delbTC(k,m)=\bTC(k),  &&
      k\in [1, \kC], \notag \\
\sum_{m\in[0,3]} \delbsF(c,m)=1,  ~~ 
\sum_{m\in[0,3]} m\cdot\delbsF(c,m)=\bsF(c),  &&    c\in [1,\cF],
   \label{eq:beta15}    
\end{align}    

 \begin{align}           
\sum_{\psi\in \FrX_i } \betar([\psi]) \cdot  \dlfrX(i,[\psi]) = \bXex(i),     
 &&  i\in [1,\tX],     \mathrm{X}\in\{\mathrm{C,T,F}\}, 
 \label{eq:beta16a}   
\end{align}

\begin{align} 
 \sum_{i\in [\widetilde{\kC}+1,\mC]} \delbC(i,m) =\bdC(m), ~~
  \sum_{i\in [2,\tT]} \delbT(i,m)    =\bdT(m),   \notag \\ 
   \sum_{k\in [1, \kC]}\delbCT(k,m)=\bdCT(m), ~~
   \sum_{k\in [1, \kC]}\delbTC(k,m)=\bdTC(m),    \notag \\ 
\sum_{i\in [2,\tF]}\!\!\! \delbF(i,m) =\bdF(m), ~~
 \sum_{c\in [1,\widetilde{\tC}]} \delbsF(c,m)  =\bdCF(m),   \notag \\ 
  \sum_{c\in [\widetilde{\tC}+1,\cF]}  \delbsF(c,m) =\bdTF(m),  
   \notag \\ 
 \bdC(m)+\bdT(m) + \bdF(m)
 +\bdCT(m)+\bdTC(m) +\bdTF(m)+\bdCF(m) = \bd^\inte(m),   \notag \\ 
  m\in [1,3].       \label{eq:beta_last} 
\end{align}   

\subsection{Assigning Chemical Elements and  Valence Condition}
\label{sec:alpha}

We include constraints so that each vertex $v$ in a selected graph $H$
satisfies the valence condition; i.e., 
$\beta_\C(v)=  \val(\alpha(v)) +\eledeg_\C(v)$, 
where $\eledeg_\C(v)=\vion(\psi)$ for the ${\rho}$-fringe-tree $\C[v]$
r-isomorphic to $\psi$. 
With these constraints, a chemical graph
   $\C=(H,\alpha,\beta)$ on a selected subgraph $H$
will be constructed. 
 
\smallskip\noindent
{\bf constants: }
 \begin{enumerate}[leftmargin=*]
\item[-] Subsets
 $\Lambda^\inte \subseteq \Lambda\setminus\{{\tt H}\}, 
 \Lambda^\ex \subseteq \Lambda$ of chemical elements,
 where we denote by $[{\tt e}]$ (resp., $[{\tt e}]^\inte$ and $[{\tt e}]^\ex$)  
 of a standard encoding of an element ${\tt e}$ in the set $\Lambda$ 
 (resp.,    $\Lambda^\inte_\epsilon$ and  $\Lambda^\ex_\epsilon$);  
\item[-]  A valence function: $\val: \Lambda \to [1,6]$;  

\item[-]  A function $\mathrm{mass}^*:\Lambda\to \mathbb{Z}$ 
(we let $\mathrm{mass}(\ta)$ denote  the observed mass of a chemical element  
$\ta\in \Lambda$, and define 
   $\mathrm{mass}^*(\ta)\triangleq
    \lfloor 10\cdot \mathrm{mass}(\ta)\rfloor$);  
        
\item[-]  
 Subsets $\Lambda^*(i)\subseteq \Lambda^\inte$, $i\in[1,\tC]$; 
 
\item[-] 
 $\na_\LB(\ta),\na_\UB(\ta)\in [0,n^* ]$,  $\ta\in  \Lambda$:
lower and upper bounds on the number of vertices  $v$    with $\alpha(v)=\ta$;  
\item[-] 
  $\na_\LB^\inte(\ta),\na_\UB^\inte(\ta)\in [0,n^* ]$,
 $\ta\in  \Lambda^\inte$:
lower and upper bounds on the number  of interior-vertices  
 $v$ with $\alpha(v)=\ta$; 

\item[-] 
$\alpha_\mathrm{r}([\psi])\in [\Lambda^\ex], \in \mathcal{F}^*$:
 the chemical element $\alpha(r)$  of the root $r$ of  $\psi$;

\item[-]   $\na_\ta^\ex([\psi])\in [0,n^*]$,  
$\ta\in \Lambda^\ex, \psi\in \mathcal{F}^*$: 
the frequency of chemical element $\ta$ in the set of  
non-rooted vertices   in   $\psi$, where possibly $\ta={\tt H}$;


\item[-]  A positive integer $\mathrm{M}\in \Z_+$: 
an upper bound for the average $\overline{\mathrm{ms}}(\C)$ of mass$^*$ 
over all atoms in $\C$;
\end{enumerate}
      
\smallskip\noindent
{\bf variables: } 
\begin{enumerate}[leftmargin=*]
\item[-]
   $\bCT(i),\bTC(i)\in [0,3], i\in [1,\tT]$:
the bond-multiplicity of edge $\eCT_{j,i}$ (resp., $\eTC_{j,i}$)
if one exists;  

\item[-] 
 $\bCF(i), \bTF(i)\in [0,3], i\in [1,\tF]$:
the bond-multiplicity of $\eCF_{j,i}$ (resp., $\eTF_{j,i}$)
if one exists;  

\item[-]  $\aX(i)\in [\Lambda^\inte_\epsilon ],
       \delaX(i,[\ta]^\inte)\in [0,1],  \ta\in \Lambda^\inte_\epsilon, i\in [1,\tX],
         \mathrm{X}\in \{\mathrm{C,T,F}\}$:  
$\aX(i)= [\ta]^\inte\geq 1$  (resp., $\aX(i)=0$)
  $\Leftrightarrow$ $\delaX(i,[\ta]^\inte)=1$ (resp., $\delaX(i,0)=0$)  
  $\Leftrightarrow$ $\alpha(\vX_{i})= \ta\in \Lambda$ 
(resp., vertex $\vX_{i}$ is not used in $\C$); 

\item[-] 
$\delaX(i,[\ta]^\inte)\in [0,1], i\in [1,\tX],  
\ta  \in \Lambda^\inte,   \mathrm{X}\in\{\mathrm{C,T,F}\}$:   
   $\delaX(i,[\ta]^\typ)=1$   $\Leftrightarrow$   $\alpha(\vX_{i})=\ta$;  
\item[-]  $\mathrm{Mass}\in \mathbb{Z}_+$: 
 $\sum_{v\in V(H)} \mathrm{mass}^*(\alpha(v))$;  
%
\item[-]  $\overline{\mathrm{ms}}\in \mathbb{R}_+$: 
 $\sum_{v\in V(H)} \mathrm{mass}^*(\alpha(v)) / |V(H)|$;  
 \item[-] 
$\delta_{\mathrm{atm}}(i)\in [0,1], i\in [n_\LB + \na_\LB({\tt H}), 
n^* + \na_\UB({\tt H})]$:   
   $\delta_{\mathrm{atm}}(i)=1$   $\Leftrightarrow$   $|V(H)| = i$;  

\item[-]   $\na([\ta])\in[\na_\LB(\ta),\na_\UB(\ta)]$,
 $\ta \in \Lambda$:
    the number  of vertices $v\in V(H)$
     with $\alpha(v)=\ta$, where possibly $\ta={\tt H}$; 
    
\item[-]   $\na^{\inte}([\ta]^\inte) \in[\na_\LB^\inte(\ta),\na_\UB^\inte(\ta)]$,
 $\ta \in \Lambda, \mathrm{X}\in \{\mathrm{C,T,F}\}$:
    the number  of interior-vertices  $v\in V(\C)$ 
    with $\alpha(v)=\ta$; 
    
\item[-]   $\naX^\ex([\ta]^\ex) , \na ^\ex([\ta]^\ex) \in [0,\na_\UB(\ta)]$,
 $\ta \in \Lambda$,   $\mathrm{X}\in \{\mathrm{C,T,F}\}$: 
    the number    of   exterior-vertices rooted at vertices $v\in\VX$ 
  and  the number    of   exterior-vertices $v$
     such that  $\alpha(v)=\ta$;
     

\end{enumerate} 
    
\smallskip\noindent
{\bf constraints: } 
\begin{align}    
 \bCT(k)-3(\eT(i)-\chiT(i,k)+1) \leq 
\bCT(i)\leq \bCT(k)+3(\eT(i)-\chiT(i,k)+1),  i\in [1,\tT], && \notag\\   
  \bTC(k)-3(\eT(i+1)-\chiT(i,k)+1) \leq 
\bTC(i)\leq \bTC(k)+3(\eT(i+1)-\chiT(i,k)+1),  i\in [1,\tT], && \notag\\ 
   k\in [1, \kC],   &&  \label{eq:alpha_first}  
\end{align}

\begin{align}   
 \bsF(c)-3(\eF(i)-\chiF(i,c)+1) \leq 
\bCF(i)\leq \bsF(c)+3(\eF(i)-\chiF(i,c)+1),   i\in [1,\tF], 
&&  c\in[1,\widetilde{\tC}] ,    \notag\\  
  \bsF(c)-3(\eF(i)-\chiF(i,c)+1) \leq 
\bTF(i)\leq \bsF(c)+3(\eF(i)-\chiF(i,c)+1),    i\in [1,\tF], 
&& c\in[\widetilde{\tC}+1,\cF] ,    \notag\\  
  \label{eq:alpha2}  
\end{align}

\begin{align}  
   \sum_{\ta\in \Lambda^\inte} \delaC(i,[\ta]^\inte)=1, ~~ 
   \sum_{\ta\in \Lambda^\inte} [\ta]^\inte\cdot\delaX(i,[\ta]^\inte)=\aC(i),   
     &&  i\in [1,\tC], \notag  \\
  \sum_{\ta\in \Lambda^\inte } \delaX(i,[\ta]^\inte)=\vX(i), ~~ 
   \sum_{\ta\in \Lambda^\inte} [\ta]^\inte\cdot\delaX(i,[\ta]^\inte)=\aX(i), 
    &&  i\in [1,\tX],  \mathrm{X}\in \{\mathrm{T,F}\},  
  \label{eq:alpha_first} 
\end{align}

\begin{align}  
\sum_{\psi\in \FrX_i } 
 \alpha_\mathrm{r}([\psi])\cdot \dlfrX(i,[\psi]) = \aX(i),   
 &&   i\in [1,\tX],   \mathrm{X}\in \{\mathrm{C,T,F}\},  
  \label{eq:alpha_1} 
\end{align}


\begin{align}  
\sum_{j\in \IC(i)}\bC(j)  
+ \sum_{  k\in \It^+(i)\cup \Iw^+(i)} \bCT(k)
+ \sum_{  k\in \It^-(i)\cup \Iw^-(i)} \bTC(k)   &&  \notag\\
     + \bsF(i)  +\bCex(i) -\eledegC(i) 
     =
     \sum_{\ta\in \Lambda^\inte}\val(\ta)\delaC(i,[\ta]^\inte),  
 &&  i\in [1,\widetilde{\tC}],  \label{eq:alpha3} 
\end{align}

\begin{align}   
\sum_{j\in \IC(i)}\bC(j)  
+ \sum_{  k\in \It^+(i)\cup \Iw^+(i)} \bCT(k)
+ \sum_{  k\in \It^-(i)\cup \Iw^-(i)} \bTC(k)   &&  \notag\\
+\bCex(i) -\eledegC(i) 
   =
      \sum_{\ta\in \Lambda^\inte}\val(\ta)\delaC(i,[\ta]^\inte),  
  &&  i\in [\widetilde{\tC}+1,\tC],   \label{eq:alpha3b} 
\end{align} 

\begin{align}  
 \bT(i)+\bT(i\!+\!1)   +   \bTex(i)  
  + \bCT(i) + \bTC(i) \hspace{1cm} \notag\\
  + \bsF(\widetilde{\tC}+i) -\eledegT(i) 
   =
      \sum_{\ta\in \Lambda^\inte}\val(\ta)\delaT(i,[\ta]^\inte), 
   \notag \\
  i\in [1,\tT]~  (\bT(1)=\bT(\tT+1)=0),   \label{eq:alpha4} 
\end{align}
 
\begin{align} 
 \bF(i)+\bF(i\!+\!1) +\bCF(i) +\bTF(i)  \hspace{1cm}    \notag\\
  +\bFex(i)  -\eledegF(i) 
   =
      \sum_{\ta\in \Lambda^\inte}\val(\ta)\delaF(i,[\ta]^\inte),  
   \notag \\
  i\in [1,\tF] ~  (\bF(1)=\bF(\tF+1)=0),   \label{eq:alpha5} 
\end{align}


\begin{align}  
 \sum_{i\in [1,\tX], i\in [1,\tX] } \delaX(i,[\ta]^\inte) = \naX([\ta]^\inte) ,  
 &&  \ta\in \Lambda^\inte, \mathrm{X}\in \{\mathrm{C,T,F}\},  
     \label{eq:alpha6} 
\end{align}    

\begin{align}   
\sum_{\psi\in \FrX_i  } \na_\ta^\ex([\psi])\cdot  \dlfrX(i,[\psi])  
   = \naX^\ex([\ta]^\ex),     && 
 \ta\in \Lambda^\ex,  \mathrm{X}\in \{\mathrm{C,T,F}\},  \label{eq:alpha6} 
\end{align}    
       
\begin{align}  
 \naC([\ta]^\inte)+ \naT([\ta]^\inte)+ \naF([\ta]^\inte) =  \na^\inte([\ta]^\inte),  
    &&   \ta\in \Lambda^\inte,       \notag \\
  \sum_{  \mathrm{X}\in \{\mathrm{C,T,F}\} }
   \naX^\ex([\ta]^\ex)       =\na^\ex([\ta]^\ex),  
  &&  \ta\in \Lambda^\ex,    \notag \\
  \na^\inte([\ta]^\inte) + \na^\ex([\ta]^\ex)=\na([\ta]),  
 &&   \ta\in \Lambda^\inte\cap \Lambda^\ex,     \notag \\
  \na^\inte([\ta]^\inte)  =\na([\ta]),  
 &&      \ta\in \Lambda^\inte \setminus \Lambda^\ex,     \notag \\ 
   \na^\ex([\ta]^\ex)  =\na([\ta]),  
 &&      \ta\in \Lambda^\ex \setminus \Lambda^\inte,        
     \label{eq:alpha6} 
\end{align}

 \begin{align}    
 \sum_{\ta\in \Lambda^*(i)} \delaC(i,[\ta]^\inte) = 1,  
  &&  i\in [1,\tC],   \label{eq:alpha8} 
\end{align}

\begin{align}   
\sum_{ \ta\in\Lambda }\mathrm{mass}^*(\ta )\cdot \na([\ta])
 =\mathrm{Mass}, &&    \label{eq:alpha7} 
\end{align}  

 \begin{align}
 \sum_{i \in [n_\LB + \na_\LB({\tt H}), n^* + \na_\UB({\tt H})]} \delta_{\mathrm{atm}}(i) = 1, && \\
 \sum_{i \in [n_\LB + \na_\LB({\tt H}), n^* + \na_\UB({\tt H})]} i \cdot \delta_{\mathrm{atm}}(i) 
 = n_G + \na^\ex([{\tt H}]^\ex), && \notag  \\
 \mathrm{Mass}  - \mathrm{M}\cdot (1 - \delta_{\mathrm{atm}}(i)) 
  \leq i \cdot \overline{\mathrm{ms}} \leq
   \mathrm{Mass}  + \mathrm{M}\cdot (1 - \delta_{\mathrm{atm}}(i)), 
   && i \in [n_\LB + \na_\LB({\tt H}), n^* + \na_\UB({\tt H})]. 
  \label{eq:alpha_last} 
 \end{align}

\subsection{Constraints for Bounds on the Number of Bonds}  
\label{sec:BDbond}

We include constraints for specification of lower and upper bounds
$\bd_\LB$ and $\bd_\UB$. 

\smallskip\noindent
{\bf constants: } 
\begin{enumerate}[leftmargin=*]
\item[-]
$\bd_{m, \LB}(i), \bd_{m, \UB}(i)\in [0,\nint_\UB]$, 
$i\in [1,\mC]$,  $m\in [2,3]$:  lower and upper bounds 
 on the number  of edges $e\in E(P_i)$ with bond-multiplicity $\beta(e)=m$
 in the pure path $P_i$ for edge $e_i\in \EC$; 
\end{enumerate}

\smallskip\noindent
{\bf variables : } 
\begin{enumerate}[leftmargin=*]
\item[-]
  $\bdT(k,i,m)\in [0,1]$, $k\in [1, \kC]$, $i\in [2,\tT]$, $m\in [2,3]$:  
  $\bdT(k,i,m)=1$  $\Leftrightarrow$ the pure path $P_k$ for edge $e_k\in \EC$ 
  contains edge $\eT_i$ with $\beta(\eT_i)=m$; 
\end{enumerate}
  
\smallskip\noindent
{\bf constraints: } 
\begin{align}    
\bd_{m,\LB}(i)\leq \delbC(i,m)\leq \bd_{m,\UB}(i), 
  i\in \Iew\cup \Iz, m\in [2,3], && 
  \label{eq:BDbond_first}  
\end{align}

\begin{align}   
\bdT(k,i,m)\geq \delbT(i,m)+\chiT(i,k)-1, 
~~~ k \in [1, \kC], i\in [2,\tT],  m\in [2,3], && 
 \label{eq:BDbond2}  
\end{align}   
 
\begin{align}    
\sum_{j\in[2,\tT]}\delbT(j,m) \geq 
\sum_{k\in[1, \kC], i\in [2,\tT]}\!\!\!\! \bdT(k,i,m) , 
~~ m\in [2,3],  \label{eq:BDbond3}  
\end{align}    

\begin{align}    
 \bd_{m, \LB}(k) \leq 
   \sum_{i\in [2,\tT]}\bdT(k,i,m) +\delbCT(k,m)+\delbTC(k,m)    
   \leq   \bd_{m, \UB}(k), ~~~~  \notag \\
    k\in [1, \kC],   m\in [2,3]. ~~ 
     \label{eq:BDbond_last}  
\end{align}

%% file: Constraints_MILP_2LMH_ac_cs_ec_polymer.tex
\subsection{Descriptor for the Number of  Adjacency-configurations}  
\label{sec:AC}

We call a tuple $(\ta,\tb,m)
\in (\Lambda\setminus\{{\tt H}\})\times (\Lambda\setminus\{{\tt H}\}) \times[1,3]$
an {\em adjacency-configuration}.
The adjacency-configuration of an edge-configuration
$(\mu=\ta d, \mu'=\tb d', m)$ is defined to be
 $(\ta,\tb,m)$.
We include constraints to compute the frequency of each adjacency-configuration
in an inferred chemical graph $\C$. 
 
\smallskip\noindent
{\bf constants: } 
\begin{enumerate}[leftmargin=*]
\item[-] A set  $\Gamma^\inte$ of edge-configurations $\gamma=(\mu,\mu',m)$ 
with  $\mu\leq \mu'$;

\item[-] 
Let  $\overline{\gamma}$ of an edge-configuration $\gamma=(\mu,\mu',m)$
denote the  edge-configuration $(\mu',\mu,m)$; 

\item[-] Let $\Gamma_{<}^\inte=\{(\mu,\mu',m)\in  \Gamma^\inte\mid \mu < \mu' \}$, 
$\Gamma_{=}^\inte=\{(\mu,\mu',m)\in  \Gamma^\inte\mid \mu= \mu' \}$
and    $\Gamma_{>}^\inte=\{\overline{\gamma}\mid 
    \gamma\in  \Gamma_{<}^\inte  \}$;
    
\item[-] 
Let  $\Gacs^\inte$, $\Gace^\inte$ and $\Gacl^\inte$ 
 denote  the sets of the adjacency-configurations of
edge-configurations in the sets 
$\Gamma_{<}^\inte$, $\Gamma_{=}^\inte$ and $\Gamma_{>}^\inte$, 
 respectively;
 
\item[-] 
Let  $\overline{\nu}$ of an adjacency-configuration $\nu=(\ta, \tb,m)$
denote the  adjacency-configuration $(\tb,\ta,m)$; 

\item[-] 
 Prepare a coding of   the  set 
$\Gac^\inte \cup \Gacl^\inte$  and let 
$[\nu]^\inte$   denote  
the coded integer of  an element $\nu$ in $\Gac^\inte \cup \Gacl^\inte$; 

\item[-] 
Choose subsets   $\tGacC,\tGacT,\tGacCT,\tGacTC, \tGacF, \tGacCF , \tGacTF
   \subseteq \Gac^\inte\cup\Gacl^\inte$;  
 To compute the frequency     of adjacency-configurations exactly,  set 
  $\tGacC:= \tGacT :=\tGacCT:=  \tGacTC :=\tGacF:=  \tGacCF := \tGacTF :=
  \Gac^\inte\cup\Gacl^\inte$; 

\item[-]  $\ac_\LB^\inte(\nu),  \ac_\UB^\inte(\nu) \in [0,2\nint_\UB ], 
\nu=(\ta,\tb,m)\in \Gac^\inte$: 
lower and upper bounds on the number 
  of interior-edges  $e=uv$  with $\alpha(u)=\ta$, 
 $\alpha(v)=\tb$ and $\beta(e)=m$; 
 
\item[-] 
 A subset $\Gac^\lnk\subseteq \Gac^\inte$ for adjacency-configurations of link-edges.
 Let $\Gacs^\lnk= \Gac^\lnk\cap \Gacs^\inte$, 
 $\Gace^\lnk= \Gac^\lnk\cap \Gace^\inte$ and  
$\Gacl^\inte=\{(\tb,\ta,m)\mid (\ta,\tb,m)\in \Gacs^\lnk\}$; 
 
\item[-]  $\ac_\LB^\lnk(\nu),  \ac_\UB^\lnk(\nu) \in [0,2\nint_\UB ], 
\nu=(\ta,\tb,m)\in \Gac^\lnk$: 
lower and upper bounds on the number 
  of link-edges  $e=uv$  with $\alpha(u)=\ta$, 
 $\alpha(v)=\tb$ and $\beta(e)=m$; 
\end{enumerate}

\smallskip\noindent
{\bf variables: } 
\begin{enumerate}[leftmargin=*]
\item[-]
$\ac^\inte([\nu]^\inte) \in [\ac_\LB^\inte(\nu), \ac_\UB^\inte(\nu)], 
\nu\in \Gac^\inte$: 
the number of interior-edges  with  adjacency-configuration $\nu$; 
\item[-]
$\acC([\nu]^\inte)\in [0,\mC],  \nu\in \tGacC$, 
$\acT([\nu]^\inte)\in [0,\tT],   \nu\in \tGacT$, 
$\acF([\nu]^\inte)\in [0,\tF], \nu\in \tGacF$:  
the number of  edges $\eC\in \EC$ (resp.,  edges $\eT\in \ET$
and   edges $\eF\in \EF$)  with  adjacency-configuration $\nu$; 

\item[-]
$\acCT([\nu]^\inte)\in [0, \min\{\kC,\tT\} ], \nu\in \tGacCT$,
$\acTC([\nu]^\inte)\in [0,\min\{\kC,\tT\} ], \nu\in \tGacCT$, 
$\acCF([\nu]^\inte)\in [0,\widetilde{\tC}],  \nu\in \tGacCF$,
$\acTF([\nu]^\inte)\in [0,\tT], \nu\in \tGacTF$: 
the number of  edges   $\eCT\in \ECT$  
(resp.,   edges $\eTC\in \ETC$
and  edges $\eCF\in \ECF$ and $\eTF\in \ETF$)  with  adjacency-configuration $\nu$; 
%
\item[-]
$\dlacC(i,[\nu]^\inte)\in [0,1], 
  i\in [\widetilde{\kC}+1,\mC]=\Iw\cup \Iz\cup\Iew, \nu\in \tGacC$, 
$\dlacT(i,[\nu]^\inte)\in [0,1],  i\in [2,\tT],   \nu\in \tGacT$, 
$\dlacF(i,[\nu]^\inte)\in [0,1] , i\in [2,\tF],\nu\in \tGacF$:
$\dlacX(i,[\nu]^\inte)=1$  $\Leftrightarrow$
edge  $\eX_i$ has  adjacency-configuration $\nu$; 
\item[-]
$\dlacCT(k,[\nu]^\inte),\dlacTC(k,[\nu]^\inte)\in [0,1],
k\in [1, \kC]=\It\cup \Iw,  \nu\in \tGacCT$: 
$\dlacCT(k,[\nu]^\inte)=1$   (resp., $\dlacTC(k,[\nu]^\inte)=1$)  $\Leftrightarrow$
edge  $\eCT_{\tail(k),j}$ (resp.,  $\eTC_{\hd(k),j}$) 
for some $j\in [1,\tT]$ has  adjacency-configuration $\nu$;

\item[-]
$\dlacCF(c,[\nu]^\inte)\in [0,1],  c\in [1,\widetilde{\tC}],\nu\in \tGacCF$:
$\dlacCF(c,[\nu]^\inte)=1$    $\Leftrightarrow$
edge   $\eCF_{c,i}$  for some $i\in [1,\tF]$ has  adjacency-configuration $\nu$;

\item[-]
  $\dlacTF(i,[\nu]^\inte)\in [0,1],  i\in [1,\tT],  \nu\in \tGacTF$: 
   $\dlacTF(i,[\nu]^\inte)=1$  $\Leftrightarrow$
edge   $\eTF_{i,j}$
 for some $j\in [1,\tF]$ has  adjacency-configuration $\nu$;   
\item[-]
$\aCT(k),\aTC(k)\in [0, |\Lambda^\inte|],   k\in [1, \kC]$: 
$\alpha(v)$  of the edge $(\vC_{\mathrm{tail}(k)},v)\in \ECT$   
 (resp., $(v,\vC_{\mathrm{head}(k)})\in \ETC$) if any; 
 
\item[-]
$\aCF(c)\in [0, |\Lambda^\inte|], c\in [1,\widetilde{\tC}]$: 
 $\alpha(v)$  of the edge $(\vC_{c},v)\in \ECF$   if any; 
 
\item[-]
$\aTF(i)\in [0, |\Lambda^\inte|], i\in [1,\tT]$: 
 $\alpha(v)$  of the edge $(\vT_{i},v)\in \ETF$   if any; 
\item[-]
$\DlacCp(i),  \DlacCm(i), \in [0,|\Lambda^\inte|], 
  i\in [\widetilde{\kC}+1,\mC]$, 
$\DlacTp(i),\DlacTm(i)\in [0,|\Lambda^\inte|],  i\in [2,\tT]$,
$\DlacFp(i),\DlacFm(i)\in [0,|\Lambda^\inte|] , i\in [2,\tF]$: 
$\DlacXp(i)=\DlacXm(i)=0$ (resp., 
 $\DlacXp(i)=\alpha(u)$ and $\DlacXm(i)=\alpha(v)$) $\Leftrightarrow$
edge  $\eX_i=(u,v)\in \EX$   is used in $\C$ (resp., $\eX_i\not\in E(G)$); 

\item[-]
$\DlacCTp(k),\DlacCTm(k)\in [0,|\Lambda^\inte|],
k\in [1, \kC]=\It\cup \Iw$: 
$\DlacCTp(k)=\DlacCTm(k) =0$ 
(resp.,  $\DlacCTp(k)=\alpha(u)$ and $\DlacCTm(k)=\alpha(v)$) 
 $\Leftrightarrow$  
edge  $\eCT_{\tail(k),j}=(u,v)\in \ECT$   
  for some $j\in [1,\tT]$ is used in $\C$ (resp., otherwise);  
  
\item[-]
$\DlacTCp(k),\DlacTCm(k)\in [0,|\Lambda^\inte|],
k\in [1, \kC]=\It\cup \Iw$: 
Analogous with $\DlacCTp(k)$ and $\DlacCTm(k)$;

\item[-]
$\DlacCFp(c)\in [0,|\Lambda^\inte|],
\DlacCFm(c) \in [0,|\Lambda^\inte|],  c\in [1,\widetilde{\tC}]$: 
$\DlacCFp(c)=\DlacCFm(c) =0$ (resp., 
 $\DlacCFp(c)=\alpha(u)$ and $\DlacCFm(c)=\alpha(v)$) 
 $\Leftrightarrow$ 
edge  $\eCF_{c,i}=(u,v)\in \ECF$  
   for some $i\in [1,\tF]$ is used in $\C$ (resp., otherwise);  
\item[-]
  $\DlacTFp(i)\in [0,|\Lambda^\inte|],
  \DlacTFm(i)\in [0,|\Lambda^\inte|],  i\in [1,\tT]$:  
  Analogous with $\DlacCFp(c)$ and $\DlacCFm(c)$;
    
\item[-]
$\ac^\lnk([\nu]^\inte) \in [\ac_\LB^\lnk(\nu),  
\ac_\UB^\lnk(\nu)],   \nu\in \Gac^\lnk$: 
the number of  link-edges   with  adjacency-configuration $\nu$;  

\item[-]
$\acC^\lnk([\nu]^\inte),\acT^\lnk([\nu]^\inte)
\in [0,\mC],  \nu\in \Gac^\lnk$:  
the number of  link-edges $\eC\in \EC$ (resp.,  edges $\eT\in \ET$)  
with  adjacency-configuration $\nu$;  

\item[-]
$\acCT^\lnk([\nu]^\inte)\in [0, \min\{\kC,\tT\} ]$,
$\acTC^\lnk([\nu]^\inte)\in [0,\min\{\kC,\tT\} ],   \nu\in \Gac^\lnk$: 
the number of  link-edges   $\eCT\in \ECT$  
(resp.,   link-edges $\eTC\in \ETC$)  with  adjacency-configuration $\nu$; 
 
\item[-] 
$\dlacTlnk(i,[\nu]^\inte)\in [0,1],  i\in [2,\tT],   \nu\in \Gac^\lnk$:
$\dlacTlnk(i,[\nu]^\inte)=1$  $\Leftrightarrow$
edge  $\eT_i \in \ET$ is a link-edge with adjacency-configuration $\nu$;  
\end{enumerate}

\smallskip\noindent
{\bf constraints: } 

\begin{align} 
 \acC([\nu]^\inte) =0,  &&  \nu \in \Gac^\inte\setminus \tGacC , \notag \\
 \acT([\nu]^\inte) =0,  &&  \nu \in \Gac^\inte\setminus \tGacT , \notag \\
 \acF([\nu]^\inte) =0,  &&  \nu \in \Gac^\inte\setminus \tGacF , \notag \\
 \acCT([\nu]^\inte) =0,  &&  \nu \in \Gac^\inte\setminus \tGacCT , \notag \\
 \acTC([\nu]^\inte) =0,  &&  \nu \in \Gac^\inte\setminus \tGacTC , \notag \\
 \acCF([\nu]^\inte) =0,  &&  \nu \in \Gac^\inte\setminus \tGacCF , \notag \\
 \acTF([\nu]^\inte) =0,  &&  \nu \in \Gac^\inte\setminus \tGacTF , \notag \\
 \label{eq:AC_first} 
\end{align}    

\begin{align} 
 \sum_{(\ta, \tb,m)=\nu\in  \Gac^\inte}\acC([\nu]^\inte) 
      =\sum_{i\in [\widetilde{\kC}+1,\mC]}\delbC(i,m),  &&   m\in [1,3]   , \notag \\
 \sum_{(\ta, \tb,m)=\nu\in  \Gac^\inte}\acT([\nu]^\inte) 
      =\sum_{i\in [2,\tT]}\delbT(i,m) ,  &&   m\in [1,3] , \notag \\
 \sum_{(\ta, \tb,m)=\nu\in \Gac^\inte}\acF([\nu]^\inte)
      =\sum_{i\in [2,\tF]}\delbF(i,m) ,  &&   m\in [1,3]  , \notag \\
 \sum_{(\ta, \tb,m)=\nu\in \Gac^\inte}\acCT([\nu]^\inte)
     =\sum_{k\in [1, \kC]} \delbCT(k,m),  &&   m\in [1,3]  , \notag \\
 \sum_{(\ta, \tb,m)=\nu\in \Gac^\inte}\acTC([\nu]^\inte)
    =\sum_{k\in [1, \kC]} \delbTC(k,m),  &&   m\in [1,3]  , \notag \\
 \sum_{(\ta, \tb,m)=\nu\in \Gac^\inte}\acCF([\nu]^\inte)
    =\sum_{c\in [1,\widetilde{\tC}]} \delbsF(c,m),  &&   m\in [1,3]  , \notag \\
 \sum_{(\ta, \tb,m)=\nu\in \Gac^\inte}\acTF([\nu]^\inte) 
    =\sum_{c\in [\widetilde{\tC}+1, \cF]} \delbsF(c,m),  &&   m\in [1,3]  , \notag \\
 \label{eq:AC_first2} 
\end{align}    

\begin{align} 
\sum_{\nu=(\ta,\tb,m) \in \tGacC }\!\!\!\! m\cdot \dlacC(i, [\nu]^\inte) 
=\bC(i), && \notag  \\  
\DlacCp(i) +\sum_{\nu=(\ta,\tb,m) \in \tGacC }\!\!\!\! [\ta]^\inte\dlacC(i, [\nu]^\inte) 
=\aC(\tail(i)),  && \notag  \\
\DlacCm(i) +\sum_{\nu=(\ta,\tb,m) \in \tGacC }\!\!\!\! [\tb]^\inte\dlacC(i, [\nu]^\inte) 
=\aC(\hd(i)),  &&\notag  \\
\DlacCp(i)+\DlacCm(i) \leq 2|\Lambda^\inte|(1 - \eC(i)),
&& i\in [\widetilde{\kC}+1,\mC],  \notag  \\
\sum_{i\in [\widetilde{\kC}+1,\mC]}\!\!\!\! \dlacC(i, [\nu]^\inte) =\acC([\nu]^\inte),  
&&  \nu \in \tGacC ,   \label{eq:AC1}   
\end{align}    

\begin{align}  
\sum_{i\in \Ilnk\cap [\widetilde{\kC}+1,\mC]}\!\!\!\! \dlacC(i, [\nu]^\inte) 
=\acC^\lnk([\nu]^\inte),  
&&  \nu \in \Gac^\lnk\cup \Gacl^\lnk,   \label{eq:AC1b}   
\end{align}     

\begin{align} 
\sum_{\nu=(\ta,\tb,m) \in \tGacT }\!\!\!\! m\cdot \dlacT(i, [\nu]^\inte) 
=\bT(i), && \notag  \\  
\DlacTp(i) +\sum_{\nu=(\ta,\tb,m) \in \tGacT }\!\!\!\! [\ta]^\inte\dlacT(i, [\nu]^\inte) 
=\aT(i-1),  && \notag  \\
\DlacTm(i) +\sum_{\nu=(\ta,\tb,m) \in \tGacT }\!\!\!\! [\tb]^\inte\dlacT(i, [\nu]^\inte) 
=\aT(i),  &&\notag  \\
\DlacTp(i)+\DlacTm(i) \leq 2|\Lambda^\inte|(1 - \eT(i)),
&& i\in [2,\tT],   \notag  \\
 \sum_{ i\in [2,\tT]} \!\! \dlacT(i, [\nu]^\inte) =\acT([\nu]^\inte),    
&& \nu \in \tGacT ,    \label{eq:AC2} 
\end{align}    

\begin{align}  
\dlacT(i, [\nu]^\inte)\!+\!\sum_{k\in \Ilnk\cap [1,\kC]}\chiT(i,k)
 \geq 2\dlacTlnk(i, [\nu]^\inte), 
~~  i\in [2,\tT],  &&  \notag  \\
\dlacTlnk(i, [\nu]^\inte) 
\geq \dlacT(i, [\nu]^\inte)\!+\!\sum_{k\in \Ilnk\cap [1,\kC]}\chiT(i,k)\!-\!1, ~
~~  i\in [2,\tT],  &&  \notag  \\
\sum_{i\in [2,\tT]}\!\!\!\! \dlacTlnk(i, [\nu]^\inte) 
=\acT^\lnk([\nu]^\inte),  
&&  \nu \in \Gac^\lnk\cup \Gacl^\lnk,   
\label{eq:AC1b}   
\end{align}     

\begin{align} 
\sum_{\nu=(\ta,\tb,m) \in \tGacF }\!\!\!\! m\cdot \dlacF(i, [\nu]^\inte) 
=\bF(i), && \notag  \\  
\DlacFp(i) +\sum_{\nu=(\ta,\tb,m) \in \tGacF }\!\!\!\! [\ta]^\inte\dlacF(i, [\nu]^\inte) 
=\aF(i-1),  && \notag  \\
\DlacFm(i) +\sum_{\nu=(\ta,\tb,m) \in \tGacF }\!\!\!\! [\tb]^\inte\dlacF(i, [\nu]^\inte) 
=\aF(i),  &&\notag  \\
\DlacFp(i)+\DlacFm(i) \leq 2|\Lambda^\ex|(1 - \eF(i)),
&& i\in [2,\tF],   \notag  \\
  \sum_{ i\in [2,\tF]} \!\! \dlacF(i, [\nu]^\inte) =\acF([\nu]^\inte),  
 &&  \nu \in \tGacF ,    \label{eq:AC3} 
\end{align}

\begin{align} 
 \aT(i)+|\Lambda^\inte|(1-\chiT(i,k)+\eT(i))\geq \aCT(k),  && \notag  \\  
\aCT(k)\geq \aT(i)- |\Lambda^\inte|(1-\chiT(i,k)+\eT(i)), 
&&  i\in [1,\tT],     \notag  \\  
\sum_{\nu=(\ta,\tb,m) \in \tGacCT }\!\!\!\! m\cdot \dlacCT(k, [\nu]^\inte) 
=\bCT(k), && \notag  \\  
\DlacCTp(k) +\sum_{\nu=(\ta,\tb,m) \in \tGacCT }\!\!\!\! [\ta]^\inte\dlacCT(k, [\nu]^\inte) 
=\aC(\tail(k)),  && \notag  \\
\DlacCTm(k) +\sum_{\nu=(\ta,\tb,m) \in \tGacCT }\!\!\!\! [\tb]^\inte\dlacCT(k, [\nu]^\inte) 
=  \aCT(k),  &&\notag  \\
\DlacCTp(k)+\DlacCTm(k) \leq 2|\Lambda^\inte|(1 - \dclrT(k)),
&& k\in [1, \kC],  \notag  \\
\sum_{k\in [1, \kC]}\!\! \dlacCT(k, [\nu]^\inte) =\acCT([\nu]^\inte),  
 && \nu \in  \tGacCT ,  \label{eq:AC5} 
\end{align}    

\begin{align}  
\sum_{i\in \Ilnk\cap [1, \kC]}\!\!\!\! \dlacCT(i, [\nu]^\inte)
=\acCT^\lnk([\nu]^\inte),  
&&  \nu \in \Gac^\lnk\cup \Gacl^\lnk,   \label{eq:AC6b}   
\end{align}     
 
\begin{align} 
 \aT(i)+|\Lambda^\inte|(1-\chiT(i,k)+\eT(i+1))\geq \aTC(k),   && \notag  \\  
\aTC(k)\geq \aT(i)- |\Lambda^\inte|(1-\chiT(i,k)+\eT(i+1)), 
&&  i\in [1,\tT],    \notag  \\  
\sum_{\nu=(\ta,\tb,m) \in \tGacTC }\!\!\!\! m\cdot \dlacTC(k, [\nu]^\inte) 
=\bTC(k), && \notag  \\  
\DlacTCp(k) +\sum_{\nu=(\ta,\tb,m) \in \tGacTC }\!\!\!\! [\ta]^\inte\dlacTC(k, [\nu]^\inte) 
=  \aTC(k),  &&\notag  \\
\DlacTCm(k) +\sum_{\nu=(\ta,\tb,m) \in \tGacTC }\!\!\!\! [\tb]^\inte\dlacTC(k, [\nu]^\inte) 
=\aC(\hd(k)),  &&   \notag  \\
\DlacTCp(k)+\DlacTCm(k) \leq 2|\Lambda^\inte|(1 - \dclrT(k)), && k\in [1, \kC],  \notag  \\  
\sum_{k\in [1, \kC]}\!\! \dlacTC(k, [\nu]^\inte) =\acTC([\nu]^\inte),  
 && \nu \in  \tGacTC ,  \label{eq:AC5} 
\end{align}    
 
\begin{align}  
\sum_{i\in \Ilnk\cap [1, \kC]}\!\!\!\! \dlacTC(i, [\nu]^\inte) 
=\acTC^\lnk([\nu]^\inte),  
&&  \nu \in \Gac^\lnk\cup \Gacl^\lnk,   \label{eq:AC5b}   
\end{align}     

\begin{align} 
\aF(i)+|\Lambda^\inte|(1-\chiF(i,c)+\eF(i))\geq \aCF(c),  && \notag  \\  
\aCF(c)\geq \aF(i)- |\Lambda^\inte|(1-\chiF(i,c)+\eF(i)), 
   &&   i\in [1,\tF], \notag  \\   
\sum_{\nu=(\ta,\tb,m) \in \tGacCF }\!\!\!\! m\cdot \dlacCF(c, [\nu]^\inte) 
=\bsF(c), && \notag  \\  
\DlacCFp(c) +\sum_{\nu=(\ta,\tb,m) \in \tGacCF }\!\!\!\! [\ta]^\inte\dlacCF(c, [\nu]^\inte) 
= \aC(\hd(c)),  &&\notag  \\
\DlacCFm(c) +\sum_{\nu=(\ta,\tb,m) \in \tGacCF }\!\!\!\! [\tb]^\inte\dlacCF(c, [\nu]^\inte) 
=\aCF(c) ,   &&\notag  \\
\DlacCFp(c)+\DlacCFm(c) \leq 2\max\{|\Lambda^\inte|,|\Lambda^\inte|\}(1 - \dclrF(c)), 
&& c\in [1,\widetilde{\tC}],  \notag  \\
\sum_{c\in [1,\widetilde{\tC}]}\!\! \dlacCF(c, [\nu]^\inte) =\acCF([\nu]^\inte),  
 && \nu \in  \tGacCF ,  \label{eq:AC6} 
\end{align}    

\begin{align} 
\aF(j)+|\Lambda^\inte|(1-\chiF(j,i+\widetilde{\tC})+\eF(j))\geq \aTF(i), 
   && \notag  \\  
\aTF(i)\geq \aF(j)- |\Lambda^\inte|(1-\chiF(j,i+\widetilde{\tC})+\eF(j)), 
  &&  j\in [1,\tF],  \notag  \\   
\sum_{\nu=(\ta,\tb,m) \in \tGacTF }\!\!\!\! m\cdot \dlacTF(i, [\nu]^\inte) 
=\bsF(i+\widetilde{\tC}), && \notag  \\  
\DlacTFp(i) +\sum_{\nu=(\ta,\tb,m) \in \tGacTF }\!\!\!\! [\ta]^\inte\dlacTF(i, [\nu]^\inte) 
= \aT(i),  &&\notag  \\
\DlacTFm(i) +\sum_{\nu=(\ta,\tb,m) \in \tGacTF }\!\!\!\! [\tb]^\inte\dlacTF(i, [\nu]^\inte) 
=\aTF(i) ,   
&& \notag  \\
\DlacTFp(i)+\DlacTFm(i) \leq 2\max\{|\Lambda^\inte|,|\Lambda^\inte|\}
(1 - \dclrF(i+\widetilde{\tC})), 
&& i\in [1,\tT],  \notag  \\
\sum_{i\in [1,\tT]}\!\! \dlacTF(i, [\nu]^\inte) =\acTF([\nu]^\inte),  
 && \nu \in  \tGacTF ,  \label{eq:AC5} 
\end{align}

\begin{align} 
\sum_{\mathrm{X}\in\{\mathrm{C,T,F,CT,TC,CF,TF}\}}\!\!\!\!\!\!
   (\acX([\nu]^\inte)+\acX([\overline{\nu}]^\inte)) 
 =\ac^\inte([\nu]^\inte) , && \nu\in \Gacs^\inte,  \notag \\  
\sum_{\mathrm{X}\in\{\mathrm{C,T,F,CT,TC,CF,TF}\}} \!\!\!\!\!\!
    \acX([\nu]^\inte)
 =\ac^\inte([\nu]^\inte) , && \nu\in \Gace^\inte,   
   \label{eq:AC10} 
\end{align}    
 
\begin{align} 
\sum_{\mathrm{X}\in\{\mathrm{C,T,CT,TC}\}}\!\!\!\!\!\!
   (\acX^\lnk([\nu]^\inte)+\acX^\lnk([\overline{\nu}]^\inte)) 
 =\ac^\lnk([\nu]^\inte) , && \nu\in \Gacs^\lnk,  \notag \\  
\sum_{\mathrm{X}\in\{\mathrm{C,T,CT,TC}\}} \!\!\!\!\!\!
    \acX^\lnk([\nu]^\inte)
 =\ac^\lnk([\nu]^\inte) , && \nu\in \Gace^\lnk,  
   \label{eq:AC_last} 
\end{align}     

\begin{align} 
 \sum_{\nu\in \nu\in \Gac^\lnk}\ac^\lnk([\nu]^\inte) = n_\lnk.   &&    \label{eq:link_ac_sum} 
\end{align}   

\subsection{Descriptor for the Number of Chemical Symbols}  
\label{sec:CS}

We include constraints for computing
 the frequency of each chemical symbol in $\Ldg$.
 Let $\cs(v)$ denote the chemical symbol of an interior-vertex $v$ in 
 a chemical graph $\C$ to be inferred; i.e.,
 $\cs(v)=\mu=\ta d\in \Ldg$ such that $\alpha(v)=\ta$ and
  $\deg_{\anC}(v)=\deg_H(v)-\deghyd_\C(v)=d$ in $\C=(H,\alpha,\beta)$. 

\smallskip\noindent
{\bf constants: } 
\begin{enumerate}[leftmargin=*]
\item[-] A set  $\Ldg^\inte$  of chemical symbols;
 
\item[-]
 Prepare a coding of each of the two sets 
$\Ldg^\inte$    and let $[\mu]^\inte$  denote  
the coded integer of  an element $\mu \in \Ldg^\inte$; 

\item[-]
Choose subsets  $\tLdgC, \tLdgT, \tLdgF  \subseteq \Ldg^\inte$:   
 To compute the frequency of chemical symbols exactly,  set
  $\tLdgC:= \tLdgT :=  \tLdgF :=\Ldg^\inte$;  
  
\end{enumerate}

\smallskip\noindent
{\bf variables: }
\begin{enumerate}[leftmargin=*]
\item[-] $\ns^\inte([\mu]^\inte )\in[0,\nint_\UB]$,  $\mu\in \Ldg^\inte$: 
      the number of interior-vertices $v$  with $\cs(v)=\mu$;   
\item[-] 
   $\dlnsX(i,[\mu]^\inte)\in [0,1]$, $ i\in [1,\tX],\mu\in \Ldg^\inte$, 
   $\mathrm{X}\in \{\mathrm{C,T,F}\}$;
\end{enumerate}

\smallskip\noindent
{\bf constraints: } 
\begin{align}  
   \sum_{\mu\in \tLdgX\cup\{\epsilon\} } \dlnsX(i,[\mu]^\inte)=1, ~~ 
   \sum_{\mu=\ta d\in \tLdgX }[\ta]^\inte\cdot\dlnsX(i,[\mu]^\inte)=\aX(i), 
   \notag \\
   \sum_{\mu=\ta d\in \tLdgX }d\cdot\dlnsX(i,[\mu]^\inte)
   =\degX(i),
    \hspace{2cm}\notag \\ 
  ~~   i\in [1,\tX],   
  \mathrm{X}\in \{\mathrm{C,T,F}\},  
  \label{eq:CS_first} 
\end{align}

\begin{align}  
   \sum_{i\in [1,\tC]} \dlnsC(i,[\mu]^\inte)
   +    \sum_{i\in [1,\tT]} \dlnsT(i,[\mu]^\inte) 
 + \sum_{i\in [1,\tF]} \dlnsF(i,[\mu]^\inte)=\ns^\inte([\mu]^\inte),
    && \mu\in \Ldg^\inte.
  \label{eq:CS_last} 
\end{align}

\subsection{Descriptor for the Number of Edge-configurations}  
\label{sec:EC}

We include constraints to compute the frequency of each edge-configuration
in an inferred chemical graph $\C$. 
 
\smallskip\noindent
{\bf constants: } 
\begin{enumerate}[leftmargin=*]
\item[-] A set  $\Gamma^\inte$ of edge-configurations $\gamma=(\mu,\mu',m)$
with $\mu\leq \mu'$, where we let $\overline{\gamma}$ denote $(\mu',\mu ,m)$;

\item[-]  Let $\Gamma_{<}^\inte=\{(\mu,\mu',m)\in  \Gamma^\inte\mid \mu < \mu' \}$, 
$\Gamma_{=}^\inte=\{(\mu,\mu',m)\in  \Gamma^\inte\mid \mu= \mu' \}$
and   $\Gamma_{>}^\inte=\{(\mu',\mu,m)\mid 
    (\mu,\mu',m)\in  \Gamma_{<}^\inte  \}$;
    
\item[-] 
 Prepare a coding of  the set 
$\Gamma^\inte \cup \Gamma_{>}^\inte$  and let 
$[\gamma]^\inte$   denote  
the coded integer of  an element $\gamma$ in $\Gamma^\inte \cup \Gamma_{>}^\inte$; 
  
\item[-] 
Choose subsets   $\tGecC,\tGecT,\tGecCT,\tGecTC,\tGecF, \tGecCF , \tGecTF
 \subseteq \Gamma^\inte\cup\Gamma_{>}^\inte$;  
 To compute the frequency  of edge-configurations exactly,  set 
  $\tGecC:= \tGecT :=\tGecCT:=  \tGecTC :=\tGecF:=  
  \tGecCF := \tGecTF := \Gamma^\inte\cup\Gamma_{>}^\inte$;  
  
\item[-] $\ec_\LB^\inte(\gamma),  \ec_\UB^\inte(\gamma) \in [0,2\nint_\UB ], 
\gamma=(\mu,\mu',m)\in \Gamma^\inte$: 
lower and upper bounds on the number  of interior-edges  $e=uv$ 
with $\cs(u)=\mu$,  $\cs(v)=\mu'$ and $\beta(e)=m$; 

\item[-] 
 A subset $\Gamma^\lnk\subseteq \Gamma^\inte$ 
 for edge-configurations of link-edges.
 Let $\Gamma_{<}^\lnk= \Gamma^\lnk\cap \Gamma_{<}^\inte$, 
 $\Gamma_{=}^\lnk= \Gamma^\lnk\cap \Gamma_{=}^\inte$ and  
$\Gamma_{>}^\inte=\{(\tb,\ta,m)\mid (\ta,\tb,m)\in \Gamma_{<}^\lnk\}$; 
 
\item[-]  $\ec_\LB^\lnk(\gamma),  \ec_\UB^\lnk(\gamma) \in [0,2\nint_\UB ], 
\gamma=(\mu,\mu',m)\in \Gamma^\inte$: 
lower and upper bounds on the number  of link-edges  $e=uv$ 
with $\cs(u)=\mu$,  $\cs(v)=\mu'$ and $\beta(e)=m$; 

\item[-]  
$\ns_\LB^\cnt([\mu]),\ns_\UB^\cnt([\mu])\in [0,2], \mu\in\Ldg^\inte$:
lower and upper bounds on the number of connecting-vertices $v$
with $\cs(v)=\mu$; Define \\
$\Gamma_{<}^\cnt:=
\{(\mu, \mu',1)\in \gamma\in \Gamma_{<}^\lnk \mid \mu,\mu'\in \Ldg^\inte, 
\ns_\LB^\cnt( \mu )\leq 1 \leq \ns_\UB(\mu), 
\ns_\LB^\cnt( \mu' )\leq 1 \leq \ns_\UB(\mu')\}$; \\
$\Gamma_{>}^\cnt:=
\{(\mu, \mu',1)\in \gamma\in \Gamma_{>}^\lnk \mid \mu,\mu'\in \Ldg^\inte, 
\ns_\LB^\cnt( \mu )\leq 1 \leq \ns_\UB(\mu), 
\ns_\LB^\cnt( \mu' )\leq 1 \leq \ns_\UB(\mu')\}$;\\
$\Gamma_{=}^\cnt:=
\{(\mu, \mu,1)\in \gamma\in \Gamma_{=}^\lnk \mid \mu \in \Ldg^\inte, 
 \ns_\UB^\cnt(\mu)=2\}$;
\end{enumerate}

\smallskip\noindent
{\bf variables: } 
\begin{enumerate}[leftmargin=*]
\item[-]
$\ec^\inte([\gamma]^\inte) \in [\ec_\LB^\inte(\gamma),\ec_\UB^\inte(\gamma)],
 \gamma\in \Gamma^\inte$: 
the number of interior-edges     with  edge-configuration $\gamma$;   
\item[-]
$\ecC([\gamma]^\inte)\in [0,\mC],  \gamma\in \tGecC$, 
$\ecT([\gamma]^\inte)\in [0,\tT],   \gamma\in \tGecT$, 
$\ecF([\gamma]^\inte)\in [0,\tF], \gamma\in \tGecF$:  
the number of  edges $\eC\in \EC$ (resp.,   edges $\eT\in \ET$
and  edges $\eF\in \EF$)   with  edge-configuration $\gamma$;  
\item[-]
$\ecCT([\gamma]^\inte)\in [0, \min\{\kC,\tT\} ], \gamma\in \tGecCT$,
$\ecTC([\gamma]^\inte)\in [0,\min\{\kC,\tT\} ], \gamma\in \tGecCT$, 
$\ecCF([\gamma]^\inte)\in [0,\widetilde{\tC}],  \gamma\in \tGecCF$,
$\ecTF([\gamma]^\inte)\in [0,\tT], \gamma\in \tGecTF$:
the number of  edges $\eCT\in \ECT$  
(resp.,   edges $\eTC\in \ETC$
and   edges $\eCF\in \ECF$ and $\eTF\in \ETF$)  with  edge-configuration $\gamma$;  
\item[-]
$\dlecC(i,[\gamma]^\inte)\in [0,1], 
  i\in [\widetilde{\kC}+1,\mC]=\Iw\cup \Iz\cup\Iew, \gamma\in \tGecC$,  
$\dlecT(i,[\gamma]^\inte)\in [0,1],  i\in [2,\tT],   \gamma\in \tGecT$,  
$\dlecF(i,[\gamma]^\inte)\in [0,1] , i\in [2,\tF],\gamma\in \tGecF$: 
$\dlecX(i,[\gamma]^\typ)=1$  $\Leftrightarrow$
edge  $\eX_i$ has  edge-configuration $\gamma$;  
\item[-]
$\dlecCTC(k,[\gamma]^\inte),\dlecTCC(k,[\gamma]^\inte)\in [0,1],
k\in [1, \kC]=\It\cup \Iw,  \gamma\in \tGecCT$: 
$\dlecCTC(k,[\gamma]^\inte)=1$   (resp., $\dlecTCC(k,[\gamma]^\inte)=1$)
  $\Leftrightarrow$
edge  $\eCT_{\tail(k),j}$ (resp.,  $\eTC_{\hd(k),j}$) 
   for some $j\in [1,\tT]$ has  edge-configuration $\gamma$; 
   
\item[-]
$\dlecCFC(c,[\gamma]^\inte)\in [0,1],  c\in [1,\widetilde{\tC}],\gamma\in \tGecCF$:
$\dlecCFC(c,[\gamma]^\inte)=1$    $\Leftrightarrow$
edge   $\eCF_{c,i}$  for some $i\in [1,\tF]$ has  edge-configuration $\gamma$; 

\item[-]
  $\dlecTFT(i,[\gamma]^\inte)\in [0,1],  i\in [1,\tT],  \gamma\in \tGecTF$:
    $\dlecTFT(i,[\gamma]^\inte)=1$  $\Leftrightarrow$
edge     $\eTF_{i,j}$ for some $j\in [1,\tF]$ has  edge-configuration $\gamma$; 
 
\item[-]
$\degCTT(k),\degTCT(k)\in [0, 4],   k\in [1, \kC]$: 
$\deg_{\anC}(v)$  of an end-vertex $v\in \VT$ of  
the edge $(\vC_{\mathrm{tail}(k)},v)\in \ECT$  
 (resp., $(v,\vC_{\mathrm{head}(k)})\in \ETC$)  if any; 
 
\item[-]
$\degCFF(c)\in [0, 4], c\in [1,\widetilde{\tC}]$: 
 $\deg_{\anC}(v)$  of an end-vertex $v\in \VF$ of  
 the edge $(\vC_{c},v)\in \ECF$   if any; 
 
\item[-]
$\degTFF(i)\in [0, 4], i\in [1,\tT]$: 
 $\deg_{\anC}(v)$   of an end-vertex $v\in \VF$ of  
  the edge $(\vT_{i},v)\in \ETF$   if any;  
\item[-]
$\DlecCp(i),  \DlecCm(i), \in [0,4], 
  i\in [\widetilde{\kC}+1,\mC]$,  
$\DlecTp(i),\DlecTm(i)\in [0,4],  i\in [2,\tT]$,  
$\DlecFp(i),\DlecFm(i)\in [0,4] , i\in [2,\tF]$: 
$\DlecXp(i)=\DlecXm(i)=0$ (resp., 
 $\DlecXp(i)=\deg_{\anC}(u)$
  and $\DlecXm(i)=\deg_{\anC}(v)$) $\Leftrightarrow$  
edge  $\eX_i=(u,v)\in \EX$  is used in ${\anC}$ (resp., $\eX_i\not\in E({\anC})$);

\item[-]
$\DlecCTp(k),\DlecCTm(k)\in [0,4],
k\in [1, \kC]=\It\cup \Iw$: 
$\DlecCTp(k)=\DlecCTm(k) =0$ 
(resp.,  $\DlecCTp(k)=\deg_{\anC}(u)$
 and $\DlecCTm(k)=\deg_{\anC}(v)$) 
 $\Leftrightarrow$  
edge  $\eCT_{\tail(k),j}=(u,v)\in \ECT$   
  for some $j\in [1,\tT]$ is used in ${\anC}$ (resp., otherwise); 
  
\item[-]
$\DlecTCp(k),\DlecTCm(k)\in [0,4],
k\in [1, \kC]=\It\cup \Iw$: 
Analogous with $\DlecCTp(k)$ and $\DlecCTm(k)$;

\item[-]
$\DlacCFp(c), \DlecCFm(c) \in [0,4],  c\in [1,\widetilde{\tC}]$: 
$\DlecCFp(c)=\DlecCFm(c) =0$ (resp., 
 $\DlecCFp(c)=\deg_{\anC}(u)$ 
 and $\DlecCFm(c)=\deg_{\anC}(v)$) 
 $\Leftrightarrow$  
edge  $\eCF_{c,j}=(u,v)\in \ECF$  
   for some $j\in [1,\tF]$ is used in ${\anC}$ (resp., otherwise);  
\item[-]
  $\DlecTFp(i),  \DlecTFm(i)\in [0,4],  i\in [1,\tT]$:
  Analogous with $\DlecCFp(c)$ and $\DlecCFm(c)$;
  
\item[-]
$\ec^\lnk([\gamma]^\inte)
 \in [\ec_\LB^\lnk(\gamma),  \ec_\UB^\lnk(\gamma)],
  \gamma\in \Gamma^\lnk$: 
the number of  link-edges   with  edge-configuration $\gamma$;  

\item[-]
$\ecC^\lnk([\gamma]^\inte),\ecT^\lnk([\gamma]^\inte)\in [0,\mC],  
  \gamma\in \Gamma^\lnk$:   
the number of  link-edges $\eC\in \EC$ (resp.,  edges $\eT\in \ET$)  
with  edge-configuration $\gamma$;  

\item[-]
$\ecCT^\lnk([\gamma]^\inte)\in [0, \min\{\kC,\tT\} ]$,
$\ecTC^\lnk([\gamma]^\inte)\in [0,\min\{\kC,\tT\} ],   \gamma\in \Gamma^\lnk$: 
the number of  link-edges   $\eCT\in \ECT$  
(resp.,   link-edges $\eTC\in \ETC$)  with  adjacency-configuration $\gamma$;  

\item[-] 
$\dlecTlnk(i,[\gamma]^\inte)\in [0,1],  i\in [2,\tT],   \gamma\in \Gamma^\lnk$:
$\dlecTlnk(i,[\gamma]^\inte)=1$  $\Leftrightarrow$
edge  $\eT_i \in \ET$ is a link-edge with edge-configuration $\gamma$;  

\item[-] 
$\delta^\cnt([\gamma]^\inte)\in [0,1], 
\gamma\in \Gamma_{<}^\cnt\cup \Gamma_{=}^\cnt\cup \Gamma_{>}^\cnt$:  
$\delta^\cnt([\gamma]^\inte)=1$ $\Leftrightarrow$ 
$\ec(e)=\gamma$ for the link-edge $e$ joining connecting-vertices;  
\end{enumerate}
  
\smallskip\noindent
{\bf constraints: } 

\begin{align} 
 \ecC([\gamma]^\inte) =0,  &&  \gamma \in \Gamma^\inte\setminus \tGecC , \notag \\
 \ecT([\gamma]^\inte) =0,  &&  \gamma \in \Gamma^\inte\setminus \tGecT , \notag \\
 \ecF([\gamma]^\inte) =0,  &&  \gamma \in \Gamma^\inte\setminus \tGecF , \notag \\
 %
 %
 \ecCT([\gamma]^\inte) =0,  &&  \gamma \in \Gamma^\inte\setminus \tGecCT , \notag \\
 \ecTC([\gamma]^\inte) =0,  &&  \gamma \in \Gamma^\inte\setminus \tGecTC , \notag \\
 \ecCF([\gamma]^\inte) =0,  &&  \gamma \in \Gamma^\inte\setminus \tGecCF , \notag \\
 \ecTF([\gamma]^\inte) =0,  &&  \gamma \in \Gamma^\inte\setminus \tGecTF , \notag \\
 \label{eq:EC_first} 
\end{align}

\begin{align} 
 \sum_{(\mu, \mu',m)=\gamma\in  \Gamma^\inte}\ecC([\gamma]^\inte) 
      =\sum_{i\in [\widetilde{\kC}+1,\mC]}\delbC(i,m),  &&   m\in [1,3]   , \notag \\
 \sum_{(\mu, \mu',m)=\gamma\in  \Gamma^\inte}\ecT([\gamma]^\inte) 
      =\sum_{i\in [2,\tT]}\delbT(i,m) ,  &&   m\in [1,3] , \notag \\
 \sum_{(\mu, \mu',m)=\gamma\in \Gamma^\inte}\ecF([\gamma]^\inte)
      =\sum_{i\in [2,\tF]}\delbF(i,m) ,  &&   m\in [1,3]  , \notag \\
 \sum_{(\mu, \mu',m)=\gamma\in \Gamma^\inte}\ecCT([\gamma]^\inte)
     =\sum_{k\in [1, \kC]} \delbCT(k,m),  &&   m\in [1,3]  , \notag \\
 \sum_{(\mu, \mu',m)=\gamma\in \Gamma^\inte}\ecTC([\gamma]^\inte)
    =\sum_{k\in [1, \kC]} \delbTC(k,m),  &&   m\in [1,3]  , \notag \\
 \sum_{(\mu, \mu',m)=\gamma\in \Gamma^\inte}\ecCF([\gamma]^\inte)
    =\sum_{c\in [1,\widetilde{\tC}]} \delbsF(c,m),  &&   m\in [1,3]  , \notag \\
 \sum_{(\mu, \mu',m)=\gamma\in \Gamma^\inte}\ecTF([\gamma]^\inte) 
    =\sum_{c\in [\widetilde{\tC}+1, \cF]} \delbsF(c,m),  &&   m\in [1,3]  , \notag \\
 \label{eq:EC_first2} 
\end{align}

\begin{align}  
\sum_{\gamma=(\ta d,\tb d',m) \in \tGecC }\!\!\!\! [(\ta,\tb,m)]^\inte\cdot \dlecC(i, [\gamma]^\inte) 
= \sum_{\nu \in \tGacC } [\nu]^\inte\cdot \dlacC(i, [\nu]^\inte) , && \notag  \\  
\DlecCp(i) +\sum_{\gamma=(\ta d,\mu',m) \in \tGecC }\!\!\!\! 
  d\cdot \dlecC(i, [\gamma]^\inte) 
=\degC(\tail(i)),  && \notag  \\
\DlecCm(i) +\sum_{\gamma=(\mu,\tb d,m) \in \tGecC }\!\!\!\!
  d\cdot\dlecC(i, [\gamma]^\inte) 
= \degC(\hd(i)),  &&\notag  \\
\DlecCp(i)+\DlecCm(i) \leq 8(1 - \eC(i)),
&& i\in [\widetilde{\kC}+1,\mC],  \notag  \\
\sum_{i\in [\widetilde{\kC}+1,\mC]}\!\!\!\! \dlecC(i, [\gamma]^\inte) =\ecC([\gamma]^\inte),  
&&  \gamma \in \tGecC ,   \label{eq:EC1}   
\end{align}

\begin{align}  
\sum_{i\in \Ilnk\cap [\widetilde{\kC}+1,\mC]}\!\!\!\! \dlecC(i, [\gamma]^\inte) 
=\ecC^\lnk([\gamma]^\inte),  
&&  \gamma \in \Gamma^\lnk\cup \Gamma_{>}^\lnk,   \label{eq:EC1b}   
\end{align}     

\begin{align} 
\sum_{\gamma=(\ta d,\tb d',m) \in \tGecT }\!\!\!\! [(\ta,\tb,m)]^\inte\cdot \dlecT(i, [\gamma]^\inte) 
= \sum_{\nu \in \tGacT} [\nu]^\inte\cdot \dlacT(i, [\nu]^\inte) , && \notag  \\  
\DlecTp(i) +\sum_{\gamma=(\ta d,\mu',m) \in \tGecT }\!\!\!\!
  d\cdot  \dlecT(i, [\gamma]^\inte) 
   =\degT(i-1 ),  && \notag  \\
\DlecTm(i) +\sum_{\gamma=(\mu,\tb d,m) \in \tGecT }\!\!\!\! 
 d\cdot \dlecT(i, [\gamma]^\inte) 
 =\degT(i),     &&\notag  \\
\DlecTp(i)+\DlecTm(i) \leq 8(1 - \eT(i)),
&& i\in [2,\tT],   \notag  \\
 \sum_{ i\in [2,\tT]} \!\! \dlecT(i, [\gamma]^\inte) =\ecT([\gamma]^\inte),    
&& \gamma \in \tGecT ,    \label{eq:EC2} 
\end{align}     

\begin{align}  
 \dlecT(i, [\gamma]^\inte) \!+\!  \sum_{k\in \Ilnk\cap [1,\kC]}\chiT(i,k)
 \geq  2\dlecTlnk(i, [\gamma]^\inte),
~~  i\in [2,\tT],   &&    \notag  \\
\dlecTlnk(i, [\gamma]^\inte) 
\geq \dlecT(i, [\gamma]^\inte)\!+\! \sum_{k\in \Ilnk\cap [1,\kC]}\chiT(i,k)\!-\!1, ~
~~  i\in [2,\tT],   &&    \notag  \\
\sum_{i\in [2,\tT]}\!\!\!\! \dlecTlnk(i, [\gamma]^\inte) 
=\ecT^\lnk([\gamma]^\inte),  
&&  \gamma \in  \Gamma^\lnk\cup \Gamma_{>}^\lnk, 
 \label{eq:EC1b}   
 \end{align}     

\begin{align} 
\sum_{\gamma=(\ta d,\tb d',m) \in \tGecF }\!\!\!\! 
[(\ta,\tb,m)]^\inte\cdot \dlecF(i, [\gamma]^\inte) 
= \sum_{\nu \in \tGacF } [\nu]^\inte\cdot \dlacF(i, [\nu]^\inte) , && \notag  \\  
\DlecFp(i) +\sum_{\gamma=(\ta d,\mu',m) \in \tGecF }\!\!\!\! 
  d\cdot \dlecF(i, [\gamma]^\inte) 
=\degF(i-1 ),    && \notag  \\
\DlecFm(i) +\sum_{\gamma=(\mu,\tb d, m) \in \tGecF }\!\!\!\! 
 d\cdot  \dlecF(i, [\gamma]^\inte) 
=\degF(i,0 ),    &&\notag  \\
\DlecFp(i)+\DlecFm(i) \leq 8(1 - \eF(i)),
&& i\in [2,\tF],   \notag  \\
  \sum_{ i\in [2,\tF]} \!\! \dlecF(i, [\gamma]^\inte) =\ecF([\gamma]^\inte),  
 &&  \gamma \in \tGecF ,    \label{eq:EC3} 
\end{align}    
  

\begin{align} 
\degT(i)+4(1-\chiT(i,k)+\eT(i))\geq \degCTT(k),  
 && \notag  \\  
\degCTT(k)\geq \degT(i)- 4(1-\chiT(i,k)+\eT(i)), &&  i\in [1,\tT],    \notag  \\  
\sum_{\gamma=(\ta d,\tb d',m) \in \tGecCT }\!\!\!\! 
[(\ta,\tb,m)]^\inte\cdot \dlecCTC(k, [\gamma]^\inte) 
= \sum_{\nu \in \tGacCT} [\nu]^\inte\cdot \dlacCT(k, [\nu]^\inte) , && \notag  \\  
\DlecCTp(k) +\sum_{\gamma=(\ta d,\mu',m) \in \tGecCT }\!\!\!\! 
  d\cdot \dlecCTC(k, [\gamma]^\inte) 
=\degC(\tail(k)),    && \notag  \\
\DlecCTm(k) +\sum_{\gamma=(\mu,\tb d, m) \in \tGecCT }\!\!\!\! 
 d\cdot  \dlecCTC(k, [\gamma]^\inte) 
= \degCTT(k),     &&\notag  \\
\DlecCTp(k)+\DlecCTm(k) \leq 8(1 - \dclrT(k)),
&& k\in [1, \kC],  \notag  \\
\sum_{k\in [1, \kC]}\!\! \dlecCTC(k, [\gamma]^\inte) =\ecCT([\gamma]^\inte),  
 && \gamma \in  \tGecCT ,  \label{eq:EC5} 
\end{align}

\begin{align}  
\sum_{i\in \Ilnk\cap [1, \kC]}\!\!\!\! \dlecCT(i, [\gamma]^\inte)
=\ecCT^\lnk([\gamma]^\inte),  
&&  \gamma \in \Gamma^\lnk\cup \Gamma_{>}^\lnk,   \label{eq:EC6b}   
\end{align}     

\begin{align} 
 \degT(i)+4(1-\chiT(i,k)+\eT(i+1))\geq \degTCT(k),  
   && \notag  \\  
\degTCT(k)\geq \degT(i)- 4(1-\chiT(i,k)+\eT(i+1)), 
&&  i\in [1,\tT],    \notag  \\  
\sum_{\gamma=(\ta d,\tb d',m) \in \tGecTC }\!\!\!\! 
[(\ta,\tb,m)]^\inte\cdot \dlecTCC(k, [\gamma]^\inte) 
= \sum_{\nu \in \tGacTC} [\nu]^\inte\cdot \dlacTC(k, [\nu]^\inte) , && \notag  \\  
\DlecTCp(k) +\sum_{\gamma=(\ta d,\mu',m) \in \tGecTC }\!\!\!\! 
  d\cdot \dlecTCC(k, [\gamma]^\inte) 
= \degTCT(k),     &&\notag  \\
\DlecTCm(k) +\sum_{\gamma=(\mu,\tb d, m) \in \tGecTC }\!\!\!\! 
 d\cdot  \dlecTCC(k, [\gamma]^\inte) 
=\degC(\hd(k)),    && \notag  \\
\DlecTCp(k)+\DlecTCm(k) \leq 8(1 - \dclrT(k)),
&& k\in [1, \kC],  \notag  \\
\sum_{k\in [1, \kC]}\!\! \dlecTCC(k, [\gamma]^\inte) =\ecTC([\gamma]^\inte),  
 && \gamma \in  \tGecTC ,  \label{eq:EC5} 
\end{align}    
 
\begin{align}  
\sum_{i\in \Ilnk\cap [1, \kC]}\!\!\!\! \dlecTC(i, [\gamma]^\inte) 
=\ecTC^\lnk([\gamma]^\inte),  
&&  \gamma \in \Gamma^\lnk\cup \Gamma_{>}^\lnk,   \label{eq:EC5b}   
\end{align}     

\begin{align} 
\degF(i)+4(1-\chiF(i,c)+\eF(i))\geq \degCFF(c), 
   && \notag  \\  
\degCFF(c)\geq \degF(i)- 4(1-\chiF(i,c)+\eF(i)), 
 && i\in [1,\tF],   \notag  \\   
\sum_{\gamma=(\ta d,\tb d',m) \in \tGecCF }\!\!\!\! 
[(\ta,\tb,m)]^\inte\cdot \dlecCFC(c, [\gamma]^\inte) 
= \sum_{\nu \in \tGacCF} [\nu]^\inte\cdot \dlacCF(c, [\nu]^\inte) , && \notag  \\  
\DlecCFp(c) +\sum_{\gamma=(\ta d,\mu',m) \in \tGecCF }\!\!\!\! 
  d\cdot \dlecCFC(c, [\gamma]^\inte) 
=\degC(c),    && \notag  \\
\DlecCFm(c) +\sum_{\gamma=(\mu,\tb d, m) \in \tGecCF }\!\!\!\! 
 d\cdot  \dlecCFC(c, [\gamma]^\inte) 
= \degCFF(c),     &&\notag  \\
\DlecCFp(c)+\DlecCFm(c) \leq 8(1 - \dclrF(c)),
&& c\in [1,\widetilde{\tC}],  \notag  \\
\sum_{c\in [1,\widetilde{\tC}]}\!\! \dlecCFC(c, [\gamma]^\inte) =\ecCF([\gamma]^\inte),  
 && \gamma \in  \tGecCF ,  \label{eq:EC6} 
\end{align}

\begin{align} 
\degF(j)+4(1-\chiF(j,i+\widetilde{\tC})+\eF(j))\geq \degTFF(i), 
  && \notag  \\  
\degTFF(i)\geq \degF(j)- 4(1-\chiF(j,i+\widetilde{\tC})+\eF(j)), 
 && j\in [1,\tF],   \notag  \\ 
\sum_{\gamma=(\ta d,\tb d',m) \in \tGecTF }\!\!\!\! 
[(\ta,\tb,m)]^\inte\cdot \dlecTFT(i, [\gamma]^\inte) 
= \sum_{\nu \in \tGacTF} [\nu]^\inte\cdot \dlacTF(i, [\nu]^\inte) , && \notag  \\  
\DlecTFp(i) +\sum_{\gamma=(\ta d,\mu',m) \in \tGecTF }\!\!\!\! 
  d\cdot \dlecTFT(i, [\gamma]^\inte) 
=\degT(i),    && \notag  \\
\DlecTFm(i) +\sum_{\gamma=(\mu,\tb d, m) \in \tGecTF }\!\!\!\! 
 d\cdot  \dlecTFT(i, [\gamma]^\inte) 
= \degTFF(i),     &&\notag  \\
\DlecTFp(i)+\DlecTFm(i) \leq 8(1 - \dclrF(i+\widetilde{\tC})),
&& i\in [1,\tT],  \notag  \\
\sum_{i\in [1,\tT]}\!\! \dlecTFT(i, [\gamma]^\inte) =\ecTF([\gamma]^\inte),  
 && \gamma \in  \tGecTF ,  \label{eq:EC6} 
\end{align}    

\begin{align} 
\sum_{\mathrm{X}\in\{\mathrm{C,T,F,CT,TC,CF,TF}\}}(\ecX([\gamma]^\inte)
+\ecX([\overline{\gamma}]^\inte)) 
 =\ec^\inte([\gamma]^\inte) , &&  \gamma\in \Gamma_{<}^\inte,  \notag \\  
\sum_{\mathrm{X}\in\{\mathrm{C,T,F,CT,TC,CF,TF}\}} \ecX([\gamma]^\inte) 
 =\ec^\inte([\gamma]^\inte) , && \gamma\in \Gamma_{=}^\inte, 
   \label{eq:EC10} 
\end{align}    

\begin{align} 
\sum_{\mathrm{X}\in\{\mathrm{C,T,CT,TC}\}}\!\!\!\!\!\!
   (\ecX^\lnk([\gamma]^\inte)+\ecX^\lnk([\overline{\gamma}]^\inte)) 
 =\ec^\lnk([\gamma]^\inte) , && \gamma\in \Gamma_{<}^\lnk,  \notag \\  
\sum_{\mathrm{X}\in\{\mathrm{C,T,CT,TC}\}} \!\!\!\!\!\!
    \ecX^\lnk([\gamma]^\inte)
 =\ec^\lnk([\gamma]^\inte) , && \gamma\in \Gamma_{=}^\lnk.    
   \label{eq:EC_11} 
\end{align}      

\begin{align} 
 \sum_{\gamma\in \Gamma^\lnk} \ec^\lnk([\gamma]^\inte) = n_\lnk,    &&    \label{eq:link_ec_sum} 
\end{align}

\begin{align} 
\ns_\LB^\cnt([\mu])\leq 
   \delta^\cnt(1,[\mu])+\delta^\cnt(2,[\mu])\leq \ns_\UB^\cnt([\mu]), &&
    \mu\in\Ldg^\inte,  \label{eq:EC_12} 
\end{align} 

\begin{align} 
\sum_{\gamma\in \Gamma_{<}^\cnt\cup \Gamma_{=}^\cnt\cup \Gamma_{>}^\cnt}
\delta^\cnt([\gamma]^\inte) =1, \notag \\ 
 \ec^\lnk([\gamma]^\inte)\geq  \delta^\cnt([\gamma]^\inte),  && 
 \gamma\in \Gamma_{<}^\cnt\cup \Gamma_{=}^\cnt \notag \\
 \ec^\lnk([\overline{\gamma}]^\inte)\geq  \delta^\cnt([\gamma]^\inte),  &&
 \gamma\in \Gamma_{>}^\cnt \notag \\
   \label{eq:EC_11} 
\end{align}     

%% file: Constraints_MILP_2LM_normalization.tex
\subsection{Constraints for Normalization  of Feature Vectors} 
\label{sec:NSFV}

  By introducing a tolerance $\varepsilon>0$   in the conversion 
 between integers and reals, we include the following constraints for 
normalizing of a  feature vector   $x=(x(1),x(2),\ldots,x(K))$: 
\begin{equation}\label{eq:normalization2}
\frac{(1 - \varepsilon)(x(j)  - \min(\dcp_j;D_\pi))}
{ \max(\dcp_j;D_\pi) - \min(\dcp_j;D_\pi) }
\leq \widehat{x}(j)   \leq 
\frac{(1 + \varepsilon) ( x(j) - \min(\dcp_j;D_\pi))}
{ \max(\dcp_j;D_\pi) - \min(\dcp_j;D_\pi) }, 
~ j\in [1,K].  
\end{equation} 
An example of  a tolerance is $\varepsilon=1\times 10^{-5}$.

We use the same conversion for descriptor $x_j = \overline{\mathrm{ms}}$. 
